\documentclass[lettersize,journal]{IEEEtran}
\usepackage{algorithmic}
\usepackage{array}
\usepackage{textcomp}
\usepackage{stfloats}
\usepackage{verbatim}
\def\BibTeX{{\rm B\kern-.05em{\sc i\kern-.025em b}\kern-.08em
    T\kern-.1667em\lower.7ex\hbox{E}\kern-.125emX}}
\usepackage{balance}

\usepackage{subfigure}
\usepackage{graphics}
\usepackage{graphicx}			
\usepackage{epsfig} 			
\usepackage{mathptmx} 	
\usepackage{times}
\usepackage{amsmath,amsthm}		
\usepackage{amsfonts}
\usepackage{amssymb}
\usepackage{epstopdf}
\usepackage[english]{babel}
\usepackage{pgfplots}
\usepackage{url}
\newtheorem{theorem}{Theorem}
\newtheorem{lemma}{Lemma}
\usepackage{mathtools}

\usepackage{graphicx}			
\usepackage{graphbox}				
\usepackage{times} 						
\usepackage{amsfonts}
\usepackage{amstext}
\usepackage{gensymb}

\usepackage{subfigure}
\usepackage{color}
\usepackage{epstopdf}

\begin{document}

\title{Analytical Model and Experimental Testing of the SoftFoot: an Adaptive Robot Foot for Walking over\\ Obstacles and Irregular Terrains}

\author{Cristina Piazza$^{1}$, Cosimo Della Santina$^{2,3}$, Giorgio Grioli$^{4,5}$, Antonio Bicchi$^{4,5}$, Manuel G. Catalano$^{4}$ %
\thanks{This work has received funding from the European Union's Horizon 2020 ERC programme under the Grant Agreement No 810346 (Natural Bionics)}%
\thanks{$^{1}$Department of Computer Engineering, School of Computation Information and Technology, and Munich Institute of Robotics and Machine Intelligence (MIRMI), Technical University of Munich (TUM), Munich, Germany} 
\thanks{$^{2}$Department of Cognitive Robotics, Delft University of Technology (TU Delft), Delft, The Netherlands} 
\thanks{$^{3}$Institute of Robotics and Mechatronics, German Aerospace Center (DLR), Oberpfaffenhofen, Germany} 
\thanks{$^{4}$Istituto Italiano di Tecnologia, via Morego, 30, 16163 Genova, Italy}%
\thanks{$^{5}$Centro “E. Piaggio” and Dipartimento di Ingegneria dell’Informazione, U\-ni\-ver\-si\-ty of Pisa, Largo Lucio Lazzarino 1, Pisa, Italy} 

\thanks{{\tt\footnotesize cristina.piazza@tum.de}}%
}

\markboth{Journal of \LaTeX\ Class Files,~Vol.~18, No.~9, September~2020}%
{How to Use the IEEEtran \LaTeX \ Templates}

\maketitle

\begin{abstract}
Robot feet are crucial for maintaining dynamic stability and propelling the body during walking, especially on uneven terrains. Traditionally, robot feet were mostly designed as flat and stiff pieces of metal, which meets its limitations when the robot is required to step on irregular grounds, e.g. stones.
While one could think that adding compliance under such feet would solve the problem, this is not the case. To address this problem, we introduced the SoftFoot, an adaptive foot design that can enhance walking performance over irregular grounds. The proposed design is completely passive and varies its shape and stiffness based on the exerted forces, through a system of pulley, tendons, and springs opportunely placed in the structure.
This paper outlines the motivation behind the SoftFoot and describes the theoretical model which led to its final design. The proposed system has been experimentally tested and compared with two analogous conventional feet, a rigid one and a compliant one, with similar footprints and soles. The experimental validation focuses on the analysis of the standing performance, measured in terms of the equivalent support surface extension and the compensatory ankle angle, and the rejection of impulsive forces, which is important in events such as stepping on unforeseen obstacles.
Results show that the SoftFoot has the largest equivalent support surface when standing on obstacles, and absorbs impulsive loads in a way almost as good as a compliant foot.
\end{abstract}


\section{INTRODUCTION}\label{sec:introduction}
\IEEEPARstart{T}{he} complex mechanical architecture of the human feet plays a fundamental role in the stability and support of the whole body under several circumstances, e.g. during controlled limb loading, locomotion, and running.
This functional importance is guaranteed by the combination of multiple subsystems, which adapt and contribute in different ways to support the anatomical arch of the foot while standing or walking, even over complex terrains. 
While a pyramid-like structure of bones provides primarily rigidity to the human foot, a complex architecture of muscles, tendons, and ligaments that runs along the entire structure provides elasticity and enables intricate movements required for motion and balance, such as standing on the toes \cite{putz2006sobotta}. 
Three arches in the foot serve as natural shock absorbers: the \textit{Medial arch}, the \textit{Longitudinal arch}, and the \textit{Tranverse arch}. These structures allow to distribute the impact of each step and reduce the amount of stress on the bones and joints. These flexible arches make walking more efficient, help to absorb impacts, and allow adaptation to uneven surfaces. Moreover, recent research in biomechanics emphasizes the significant role of the transverse tarsal arch morphology in providing over 40\% of the foot's longitudinal stiffness, which contributed to the evolutionary development of human bipedalism \cite{venkadesan2020stiffness}.

\begin{figure}[t!]%
\centering
\subfigure[]{\includegraphics[width=0.49\columnwidth]{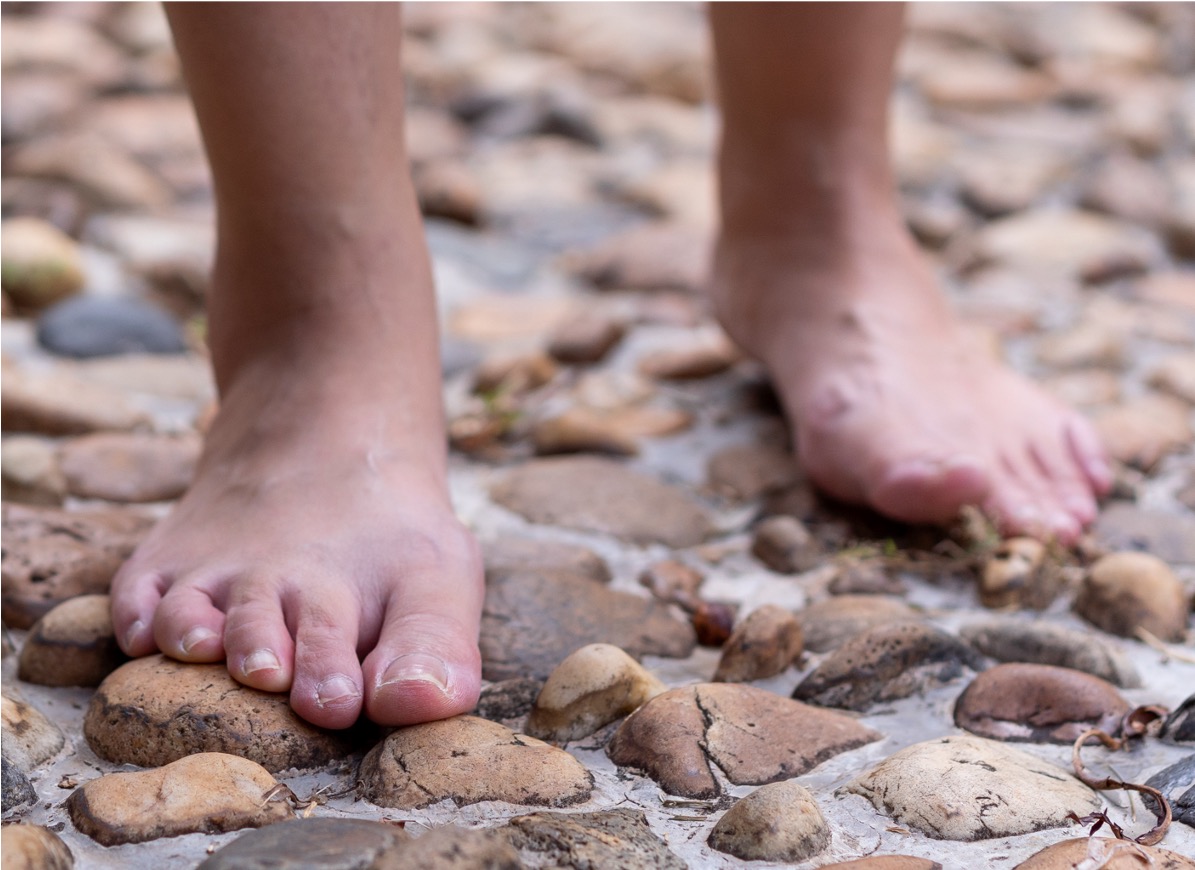}}
\subfigure[]{\includegraphics[width=0.49\columnwidth]{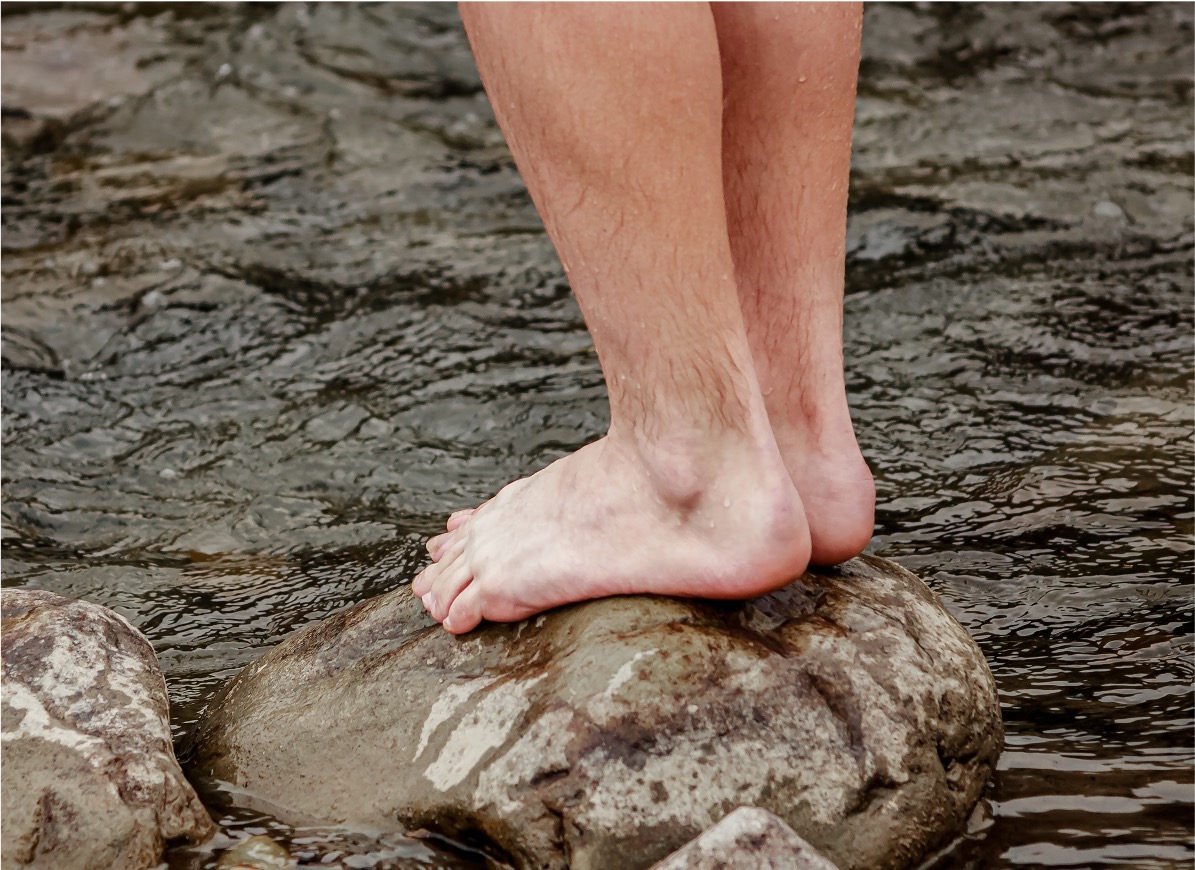}}
\subfigure[]{\includegraphics[width=0.49\columnwidth]{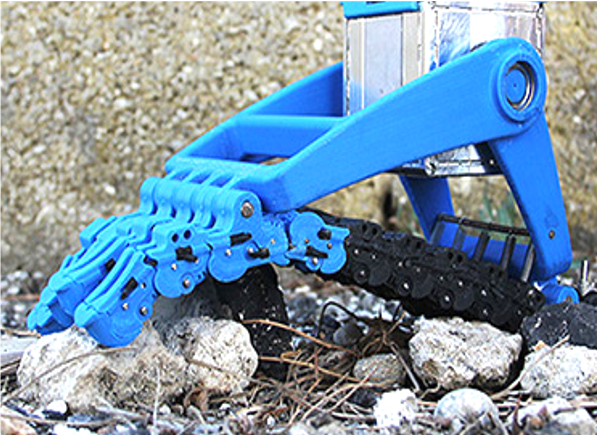}}
\subfigure[]{\includegraphics[width=0.49\columnwidth]{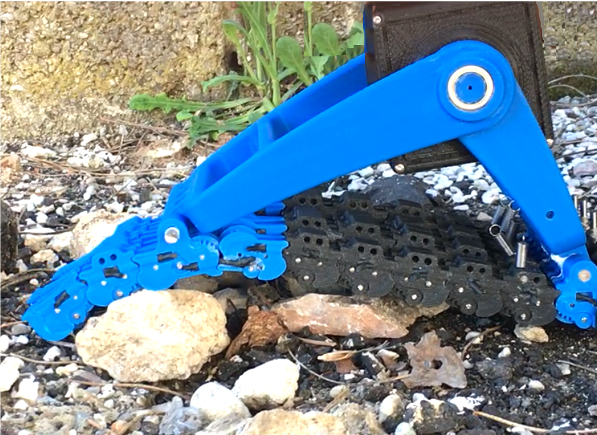}}
\caption{Pictures show a comparison between a human foot (a-b) and the SoftFoot (c-d), while walking on irregular ground. The SoftFoot is designed with a compliant structure that allows it to adapt to uncertain environments and increase the device robustness towards impact. Pictures a-b adapted from Shutterstock.}%
\label{fig:human_soft_foot}%
\end{figure}

However, to recreate the architecture, mechanical proprieties, and functionalities of the human foot in an artificial counterpart is not trivial  \cite{torricelli2016human}. For this reason, the main approach adopted in robotics is oriented towards minimalistic design solutions and is based on simplified models of human foot anatomy \cite{frizza2022humanoids, jaeger2023cybathlon}. 
Quadrupedal robots adopt design with cylindrical or spherical shape \cite{hutter2017anymal, sprowitz2013towards}, while most of the humanoid robot feet are designed as passive flat feet \cite{BostonDyn, park2005mechanical, kaneko2008humanoid, gouaillier2009mechatronic, negrello2016walkman}. This approach can offer great stability and control, with generally good performance in case of flat ground.
However, the need for more complex control algorithms to achieve effective coordination with the rest of the robot is crucial in real-world unstructured scenarios. Addressing these challenges requires innovative approaches, to estimate the stability margin without increasing the computation time \cite{bretl2008testing}.

An alternative consists of more adaptive human-like design approaches.
Some architectures enhance the simple flat design with the introduction of layers of compliant material in the feet sole \cite{li2008flexible, najmuddin2012experimental, tsagarakis2011design}, or actuation mechanism \cite{davis2010design, kang2010realization, kuehn2012active} to be more adaptive to different environments. However, all these solutions lead to a significant increase in complexity in terms of robot control and motion planning of walking in environments with uncertainty. More complex designs take inspiration from the natural biomechanics of the human foot and include multiple segments and a high level of articulation \cite{narioka2012humanlike, seo2009modeling, yoon2006novel}. While they offer increased adaptability to uncertainties, they may require additional sensors and actuators to control the increased number of degrees of freedom.

In recent years, soft robotics technologies have been explored as a potential design approach for robotic systems designs \cite{frizza2022humanoids},\cite{piazza2019century}.
The adoption of soft and flexible materials allows to create compliant structures that can deform and adapt to external variations, while increasing the overall robustness. In the last century, the same approach was extensively investigated in the design of robotic hands \cite{piazza2019century} for different applications (e.g. prosthetics and rehabilitation, human-robot interaction, etc) and has the potential to be extended also to lower limb.
The introduction of flexible elements allows to get an adaptive design which makes the system more efficient and stable to unknown and uneven terrains. Several works are exploring this approach, proposing different design solutions \cite{ davis2010design, kang2010realization}, \cite{paez2019adaptive}, \cite{asano2016human}, \cite{kaslin2018towards} which in most cases try to mimic the phalanges and metatarsal joints of the human foot. An alternative approach explores the compliant universal gripper as an innovative foot design \cite{hauser2018compliant}, aiming to reproduce the stiffness-varying function of human tarsal bones. Its granular jamming mechanism allows the foot to dynamically switch between a soft state, ideal for impact damping and adaptation, and a hard state, optimized for efficient propulsion. 

The idea proposed in this work consists of the introduction of a biomimetic shape-morphing robotic foot, that can passively adapt its shape to the ground on which it walks. This design is the result of a scientifically principled biomimetic combination of flexible and rigid structures inspired by the human architecture and realized by exploiting soft robotics technologies. The proposed prototype, called SoftFoot and presented in Fig. \ref{fig:human_soft_foot}, is a completely passive system able to change its shape according to different obstacles or soil irregularities commonly encountered in a real-world scenario, without the need for additional sensors or actuators. 
A preliminary investigation of this concept is presented in \cite{piazza2016toward}, that introduces the first prototype and propose a first validation. Encouraging results led to explorative testing of the SoftFoot concept in real-world scenarios. Specifically, \cite{mura2019exploiting} proposes a method for reconstructing the distribution of contact forces on the sole of the foot, while in  \cite{catalano2020hrp} the SoftFoot was interfaced with the biped humanoid robot HRP-4  to test balancing, stepping, and walking. A similar concept was also tested on the quadrupedal robot ANYmal \cite{catalano2021adaptive, bednarek2020cnn}, but employing a simplified version of the SoftFoot with significantly reduced compliance in the joint design.
However, none of the existing studies have analyzed the mathematical framework underlying this approach or conducted a thorough experimental validation to assess the system's robustness and measure its ability to absorb impacts. Given the novelty of the approach compared to literature, conducting an in-deep investigation could enable design improvements to meet specific requirements and provide essential insights into the potential application of the SoftFoot concept in other fields.

This work provides a comprehensive investigation of the SoftFoot concept and discusses its unique adaptive capabilities through a detailed mathematical framework. In contrast to existing studies, this work includes a thorough analysis of the deformation dynamics of the SoftFoot and presents an extensive experimental validation of the proposed prototype. Extending the preliminary analysis of \cite{piazza2016toward}, two different sets of experiments are conducted, to evaluate its standing performance and ability to reject impulsive forces during impacts. The SoftFoot performance is compared to those obtained with the most commonly used design solutions in the robotic feet state-of-art, such as rigid and compliant feet with a similar footprint and sole. Results highlight the potential of this novel design and its possible application to different research fields.

The rest of the paper is organized as follows: Section \ref{sec:pro_def} introduces the motivations of using an adaptive design in locomotion and discuss a framework for its design. Section \ref{sec:design} presents the SoftFoot mechanical design, while the experimental validation and the results are presented and discussed in Section \ref{sec:experimental_results}. Finally, Section \ref{sec:concl} draws the conclusions of our work.

\section{PROBLEM DEFINITION}\label{sec:pro_def}
\begin{figure}[t!]
	\centering
	\subfigure[two surfaces in contact]{\includegraphics[width=0.5\columnwidth]{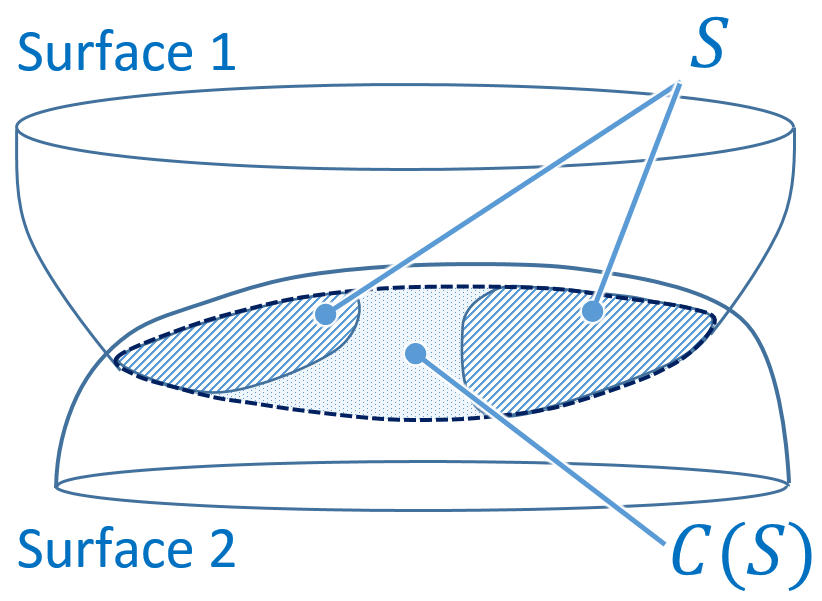}}
	\subfigure[traction for the contact point $r$]{\includegraphics[width=0.45\columnwidth]{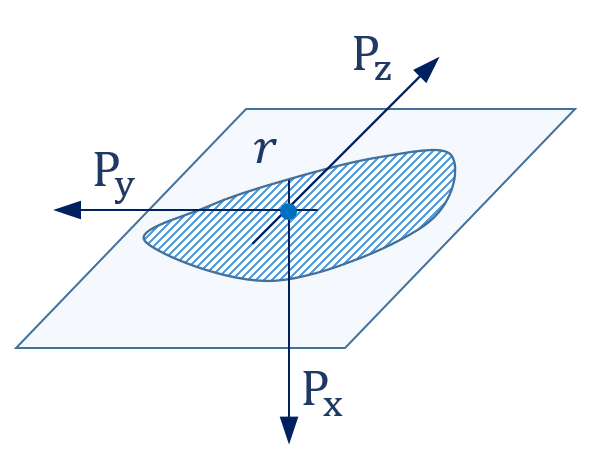}}
	\caption{Two surfaces in contact (a): $S$ is the eventually non\--convex contact area, $C(S)$ is the convex hull of $S$. In (b) we report the traction $P(r)$ exerted in a generic contact point $r \in S$. \label{fig:sdr}}
\end{figure}	

\subsection{Theoretical Background}\label{sec:theoretical}

Consider a humanoid robot during the single-support phase, supported by its foot laying on the ground. The effect of gravity, interactions of the upper body with the environment, body dynamics, create a set of forces acting on the robot. These forces can be characterized by the total resulting force acting on the robot, called $F_{\mathrm A}$ , and the total momentum with respect to its center of mass, called $M_{\mathrm A}$. For the robot to maintain its balance and standing configuration, $F_{\mathrm A}$ and $M_{\mathrm A}$ need to be compensated by the ground-foot interaction.

If we assume that there is no sliding at the ground-foot contact, static friction compensates for the component of $F_{\mathrm A}$ parallel to the ground, and for the component of $M_{\mathrm A}$ orthogonal to the ground. Assume that the ground compensates for the component of $F_{\mathrm A}$ orthogonal to the ground itself, the components of $M_{\mathrm A}$ tangent to the ground are the only actions left to compensate. In the literature the point on the ground surface with respect to which such components are null is called the Zero Moment Point (ZMP). 

\begin{figure*}[h]
	\centering
	\subfigure[static cart table model]{\includegraphics[width = 0.24\columnwidth]{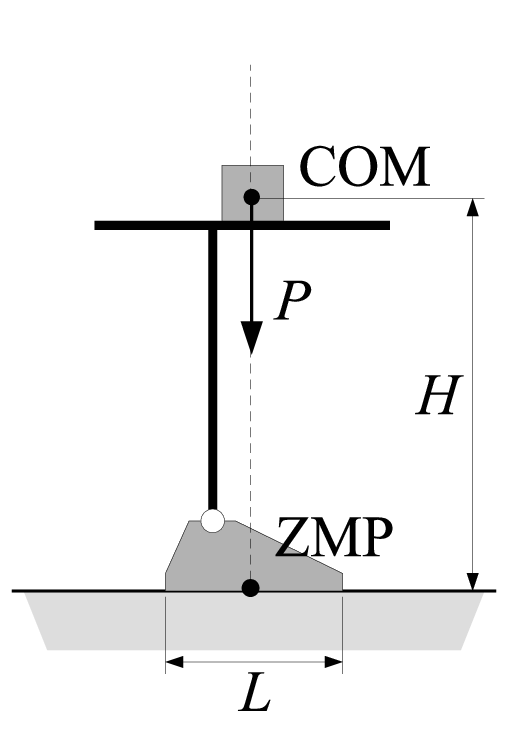}}
	\subfigure[rigid flat foot on obstacle]{\includegraphics[width = 0.24\columnwidth]{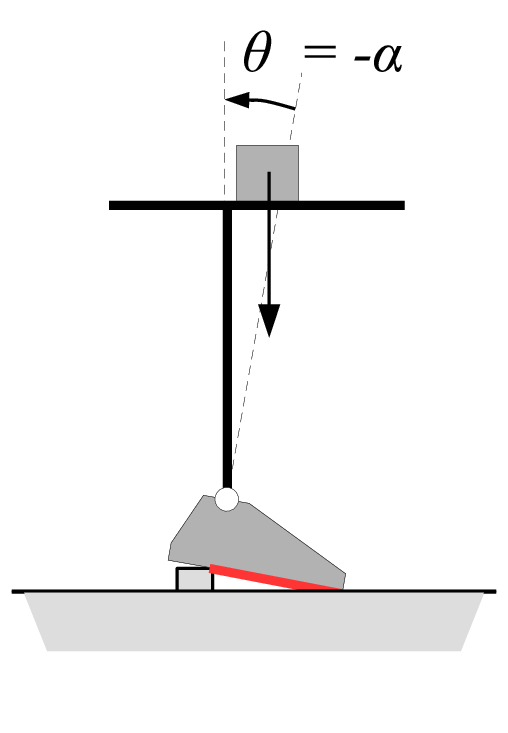}}
	\subfigure[compliant flat foot]{\includegraphics[width = 0.24\columnwidth]{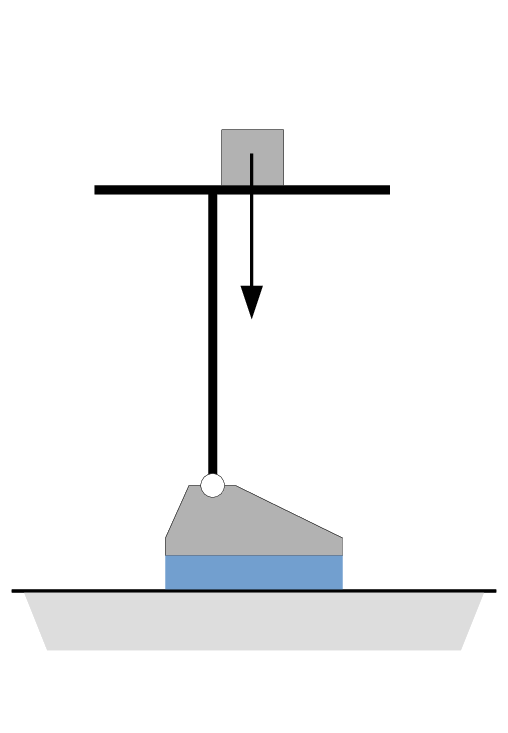}}
	\subfigure[compliant flat foot on obstacle]{\includegraphics[width = 0.24\columnwidth]{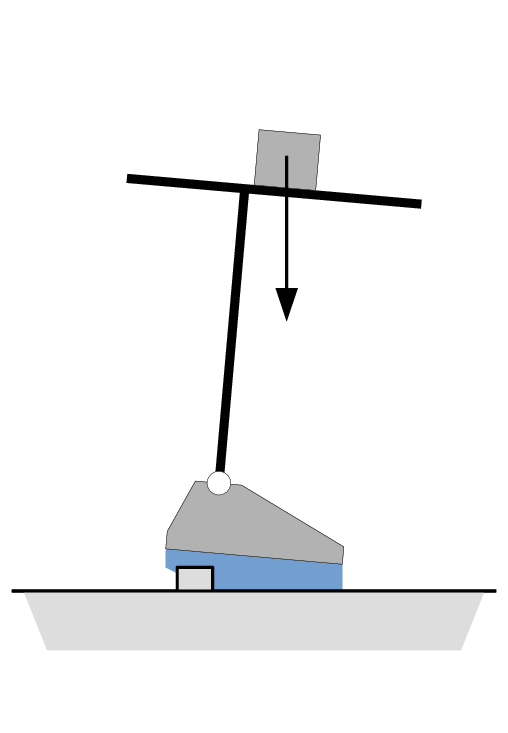}}
	\subfigure[compliant flat foot lumped model]{\includegraphics[width = 0.24\columnwidth]{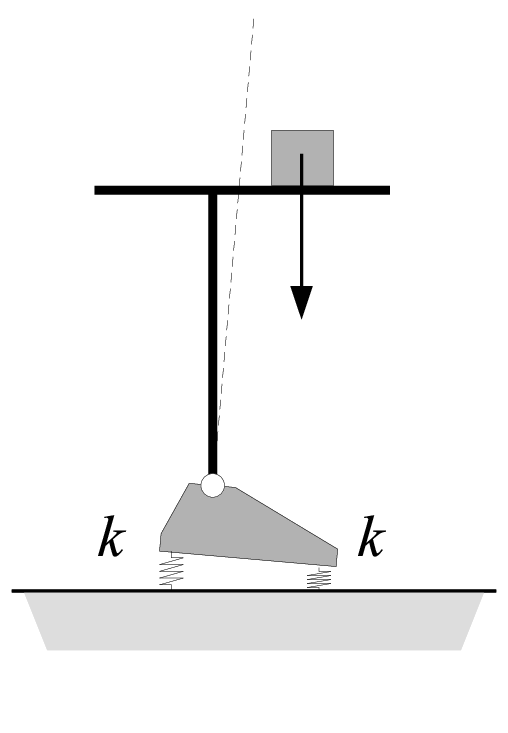}}
	\subfigure[adaptive foot]{\includegraphics[width = 0.24\columnwidth]{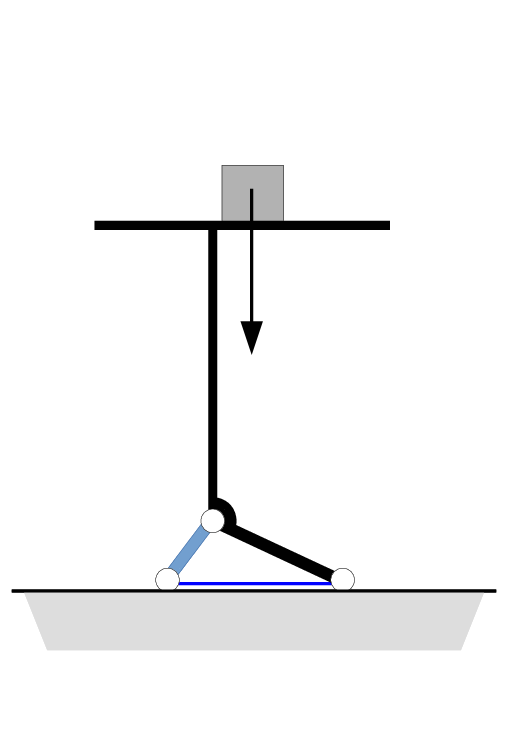}}
	\subfigure[adaptive foot on obstacle]{\includegraphics[width = 0.24\columnwidth]{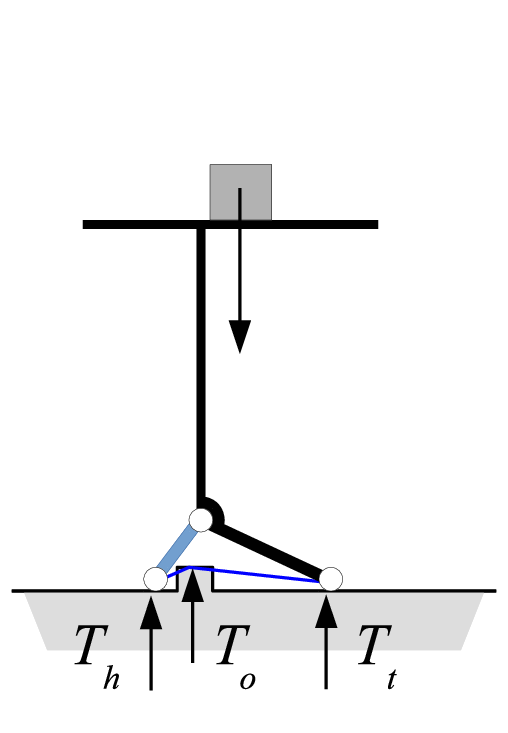}}
	\caption{Schematic representation of different feet models described in Section \ref{sec:comparison} and \ref{sec:adaptive}: rigid flat foot (a-b), compliant flat foot (c-e), and adaptive foot (f-g). It is possible to observe different behaviors according to the foot design characteristics and the shape of the terrain encountered. COM = center of mass; ZMP = Zero Moment Point;  P = traction force; H = leg height; L = contact surface; $\theta$ = ankle motion angle; $\alpha$ = rotation of the foot ankle due to the obstacle; $k$ = sole stiffness; $T_{\mathrm{h}}$ = contact force at the heel of the foot; $T_{\mathrm{o}}$ = contact force at the obstacle; $T_{\mathrm{t}}$ = contact force at the tip of the foot.} \label{fig:piedi}
\end{figure*}

Considering the case of flat foot lying on flat ground, the ZMP is always well defined whenever the component of $F_{\mathrm A}$ orthogonal to the ground is greater than zero. 
It is a well-known in literature that a robot supported by a flat foot in contact with flat ground can maintain balance if the ZMP is contained within the set of points that establish the contact area of the foot. In the case of multiple coplanar flat contacts this result extends to the convex hull of all the contact regions \cite{popovic2005ground}.
However, extending the ZMP stability test to uneven terrain requires more complex analysis, and several works have been devoted to this topic. Important examples are \cite{sardain2004forces,sato2009stability} and \cite{caron2015zmp}. The characterization of force distribution on non-planar surfaces, which generalizes the concept of Center of Pressure, on the other hand, is discussed and generalized in \cite{bicchi1993contact}.

Considering two general surfaces, we define $S$ their contact region, i.e. the set of areas and points where they are in contact (see Fig. \ref{fig:sdr}). The contact convex hull, called $C(S)$, is the smallest convex portion of the two surfaces that encloses every contact point and/or area. We assume in the following that there exists a plane $P$ such that $C(S) \subset P$. 
If the bodies are compliant, finite portions of the surface may come into contact, and if friction is present, torques may also be exerted. 

It results in a transmission of a distribution of contact tractions over $S$. At each point $r$ of the contact area $S$, bodies mutually exert traction. Be $P(r) = [P_{\mathrm x}(r), \, P_{\mathrm y}(r), \, P_{\mathrm z}(r)]^T$ the traction for every point $r \in S$ expressed in a frame tangent to the contact, such that the component $P_{\mathrm x}$ is that normal to the tanget plane. The $P_{\mathrm x}$ component is usually referred to as pressure, while $P_{\mathrm y}$ and $P_{\mathrm z}$ are the friction components. The tractions are assumed to be compressive, meaning that $P_{\mathrm x}(r) \geq 0 \; \forall r \in S$. Please note that this model disregards adhesive forces between bodies.

The overall resulting contact force that is exerted by the ground is $ F_c \triangleq \int_S P(r) \, \mathrm{d}r $. 

A contact centroid $c$ \cite{bicchi1993contact} for $S$ and $P$ is a point on the contact such that a set of forces equivalent to $P$ exists, having the following characteristics: i) it is comprised of only a force and a torque; ii) the force is applied at that point, and is directed into S; iii) the moment is parallel to the surface normal at that point.

\begin{theorem}\label{th:bicchi90}
	Be $P_{\mathrm x}(r) \geq 0 \; \forall r \in S$, i.e. the contact pressure can only be directed towards the inside of the bodies. Thus $c \in C(s)$, i.e. contact centroid lies in the contact surface convex hull.
\end{theorem}

If we consider the robot foot sole and the ground as the two contacting surfaces described before, the next corollary derives directly:

\begin{lemma}
	If a system of forces acting on a robot admits ZMP $z$ on a generic ground surface and $z \in C(S)$, then a distribution of contact forces exists such that it balances the system of forces, and the robot is in equilibrium.
\end{lemma}

This generalization of ZMP balancing condition leads to the design guideline that feet architectures that maximize the convex hull $C(S)$ should be considered to improve the robot stability. Given a flat solid ground, it is clear that a flat rigid foot is the best option to obtain this result. On the other hand, in the presence of a generic uneven terrain, different designs can be more effective and obtain better results. 

The following sections discuss and evaluate the use of soft and adaptive designs to increase the contact surface area $S$, and hence $C(S)$ on uneven terrain.

\subsection{Rigid vs Soft foot}\label{sec:comparison}
To conduct a simplified analysis, the problem can be reduced by considering it two-dimensional and static.
As shown in Fig.~\ref{fig:piedi} (a), a robot standing on a flat foot can be modelled as a static table-cart system \cite{kajita2003biped}. Referring to the figure, the sole force acting on the robot (excluding the ground) is gravity, thus the ZMP corresponds to the vertical projection of the center of mass (COM) on the ground. It is easy to show that all the set of points, in which the COM can be moved so that the ZMP falls within the contact surface (which is the foot sole), admit a balancing reaction force distribution.

If the foot steps over a small obstacle, as shown in the schematics of Fig.~\ref{fig:piedi} (b), two negative effects arise. First, the obstacle induces a rotation $\alpha$ on the foot, which tends to displace the COM by $\alpha_{\mathrm H}$. Unless this displacement is very small, it needs to be compensated by a rotation of the ankle of an angle $\theta -\alpha$, which we assume. Second, the convex hull of the contact surface shrinks (to the colored segment, which is sensibly shorter of the feet length).

As presented in Fig.~\ref{fig:piedi} (c), a possible alternative to improve the rigid flat foot would be to make a compliant flat foot. This foot has an additional layer of soft material between the rigid parts of the foot and the ground. The heuristic behind this choice is that a soft foot could absorb the obstacle, as in \ref{fig:piedi} (d), thus avoiding the need for ankle compensation and, moreover, keeping a larger contact surface. 

One may think that the softer is the foot sole the better. However lower limits exist for the sole stiffness. To understand why, consider the simplified compliant flat foot model of Fig.~\ref{fig:piedi} (e), where all the compliance is concentrated in two springs supporting the tip and the heel of the foot, standing on a flat ground. 
Moving the robot COM forward by $x$ from the vertical line above the foot mid point yields a rotation of the foot of an angle $\alpha \approx  x 2P/kL^2 $. Physical limits on the range for the ankle motion $|\theta| <\theta_{max}$, yields a lower bound on 
\begin{equation}
\begin{split}
k > k_{min_s} = P/L\theta_{max} \;,
\end{split}
\end{equation}
below which the ankle can not compensate for the foot rotation over all the possible contact surface and keep the robot vertical (which means that if $k < k_{min}$ feasible support length shrinks with respect to the full contact surface). A second practical limit to the sole stiffness comes from the necessity for the system equilibrium to be stable. It is in fact well known that an elastic inverted pendulum is stable on the topmost equilibrium if the torsional stiffness $k_\theta > mgH$, that in our simplified model translates as 
\begin{equation} \label{eq:n2}
\begin{split}
k > k_{\mathrm{min}_g} = 2mgL^2/H \;.
\end{split}
\end{equation}
If this second condition is not met, it would be in practice impossible to keep the robot standing passively, and equilibrium would require a possibly expensive active control action. 

\subsection{Adaptive Foot} \label{sec:conti}
A different solution, which constitutes the basic idea of the system we propose in this paper, is that of an adaptive mechanism as that of Fig.~\ref{fig:piedi} (f). This foot is composed of a frontal arch, which connects to the robot through the ankle and lays on the ground on the foot tip, and a backward heel arch, idle on the ankle, which supports the back side of the foot thanks to a flexible traction beam which holds the two arches together. The equilibrium of this foot on a solid flat ground is the same of that of a rigid foot, thus avoiding the tilting problems of compliant flat feet, but lets the foot adapt to obstacles as in the example of Fig.~\ref{fig:piedi}(g). Some calculations can be used to show that the contact force balance so as to obtain\footnote{The angles $\alpha_1$ and $\alpha_2$ are the angles the traction beam forms with the flat ground on the tip and heel, respectively, while $\alpha_H$ is the angle that the backward heel arch forms with respect to the vertical direction.}:
\begin{eqnarray}
T_{\mathrm{h}} &=& P\frac{(L - x_{\mathrm{com}})}{L} (1 - \tan\alpha_1 \tan\alpha_H)\\
T_{\mathrm{o}} &=& P\frac{(L - x_{\mathrm{com}})}{L} (\tan\alpha_1 + \tan\alpha_2) \tan\alpha_H\\
T_{\mathrm{t}} &=& P\frac{(x_{\mathrm{com}} - L)( \tan\alpha_2 \tan\alpha_H ) + x_{\mathrm{com}}}{L}  \;, 
\label{eq:}
\end{eqnarray}
which are all always positive (thus admittable) as long as the x coordinate of the COM $x_{\mathrm{com}} < L$ (in analogy with the rigid foot) and 
\begin{equation} 
\begin{split}
x_{\mathrm{com}} > L \left( \frac{1 - \tan \alpha_H \tan \alpha_2}{\tan \alpha_H \tan \alpha_2} \right)\;.
\end{split}
\end{equation}
It is worthwhile noting that although it is adaptive, the foot of Fig.~\ref{fig:piedi}(f), displays infinite stiffness once the contact with the ground is acquired. This property which is desirable in terms of stability of the support, performs poorly in terms of step shock absorption, where a compliant flat foot as that if Fig.~\ref{fig:piedi} (c) would probably offer better performance.

\section{ADAPTIVE FOOT MODEL}\label{sec:adaptive}
\begin{figure}
	\centering
	\subfigure[Human foot]{\includegraphics[height = 0.66\columnwidth]{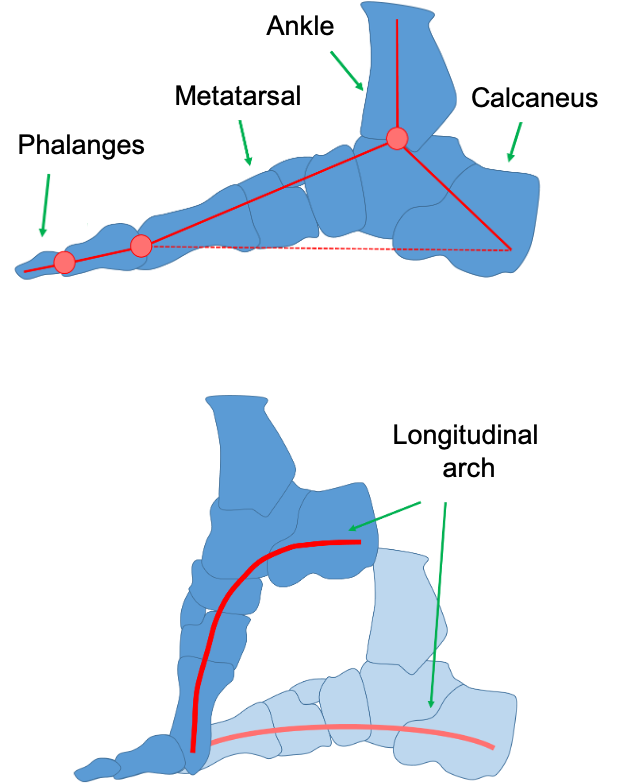}}
	\subfigure[SoftFoot]{\includegraphics[height = 0.66 \columnwidth]{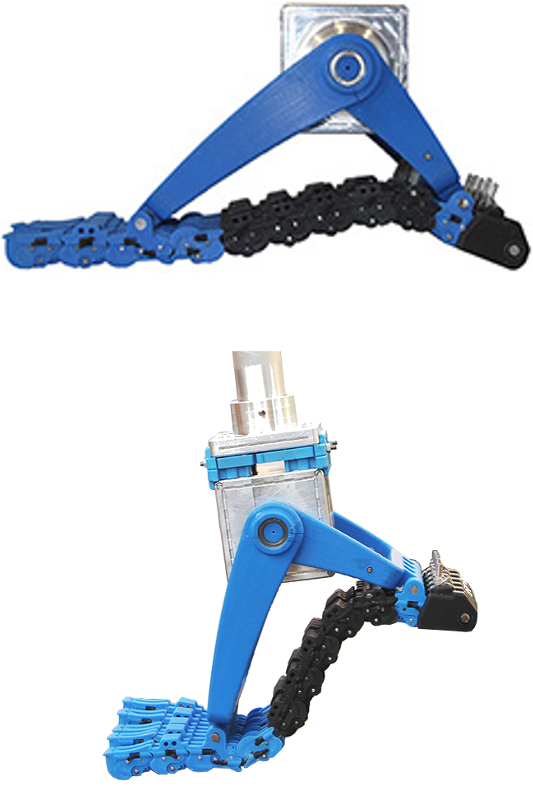}}
	\caption{Schematic architecture of the human foot, bones, phalanges and representation of the \textit{longitudinal arch} and windlass mechanism $(a)$. Prototype of the robotic foot, with components adopted for the implementation of the artificial \textit{longitudinal arch} and windlass mechanism $(b)$.}
	\label{fig:human_foot}
\end{figure}

The intricate yet functional biomechanical structure of the human foot allows to withstand the weight of the human body while remaining flexible and elastic \cite{venkadesan2020stiffness}.  The foot's bones are organized into two primary arch structures: the longitudinal arch and the transverse arch (refer to \cite{putz2006sobotta}). The longitudinal arch, illustrated in Fig.~\ref{fig:human_foot} $(a)$, forms a triangular geometry involving the Calcaneus, a set of metatarsal bones, phalanges, and the Planar Fascia at the foot base. This geometric distribution, in conjunction with a system of tendons and muscles, creates so-called Foot Windlass Mechanism, initially studied by Hicks in 1954 \cite{hicks1954mechanics}. In engineering (e.g. in yachting) windlass mechanisms are adopted to move heavy loads, in foot mechanics the windlass mechanism describes the tightening action of the long plantar fascia of the foot to maintain arch stability when the heel comes off the ground (late stance phase of the gait). This architecture contributes to key features of the foot system, including energy storage, impact absorption, adaptability to terrain irregularities, and stabilization during weight-bearing on the ankle. The foot prototype presented in this paper aims to emulate some of these features in a robotic system, implementing a mechanical architecture that translates the behavior of a human foot into feasible engineered complexity. Fig. ~\ref{fig:human_foot} $(b)$ illustrates the proposed prototype. The following section presents the mathematical framework that gives the foundations and motivates our design choices.

\subsection{Mathematical Model} \label{sec:adaptive}

\begin{figure}
	\centering
	\includegraphics[width=0.9\columnwidth]{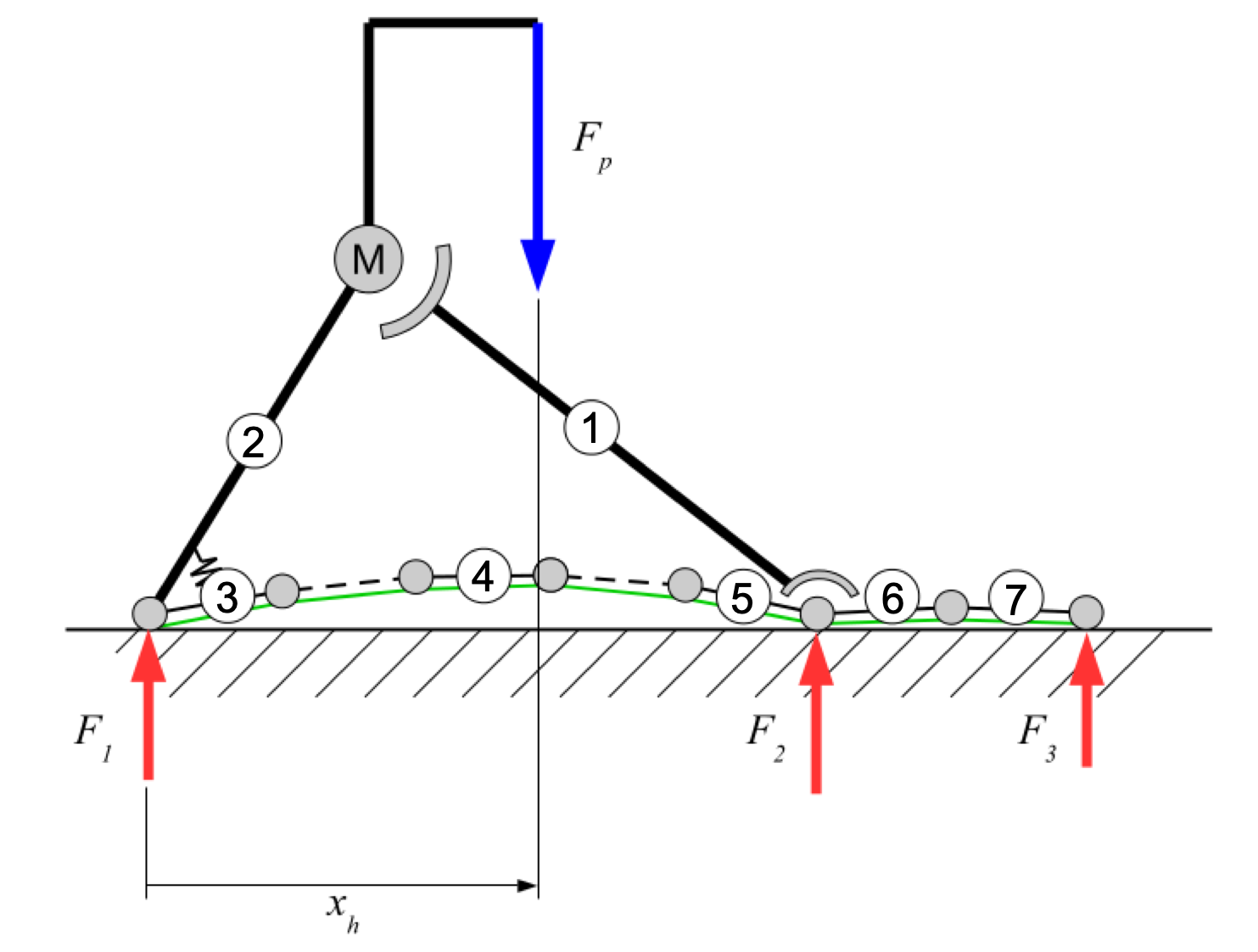}
	\caption{Architecture of the SoftFoot, simplified kinematic with the main parts underlined. $F_1,F_2,F_3$ are the are the three considered contact forces, $F_P$ is the load applied by the robot, which is connected to body (2) through the ankle. (6-7) represent the phalanxes. The plantar fascia is implemented by the set of links (3-4-5) and the tendon (green in figure) which is connected from the calcaneous to the tip of the toe. Bodies (3-4-5-6-7) are connected each other through a spring of stiffness $e$. Bodies (2-3) are also connected through a spring of stiffness $e_0$.} 
	\label{statics}
\end{figure}

In \cite{piazza2016toward} we introduced a foot prototype based on a refinement of the previous idea, a scheme of which is shown in Fig. \ref{statics}. In the following we will refer to the variables, the reference frames and the rigid bodies defined in Fig. \ref{statics}, to deduce a mathematical model that can be used to study its behaviour when statically balancing.
The variable $q \, = \begin{bmatrix} q_0 & \dots & q_{n+2} \end{bmatrix} \in \mathbb{R}^{n+3}$ collects the configuration of the foot sole, and of the phalanges.

\begin{figure*}[h!]%
	\centering
	\subfigure[body 1]{\includegraphics[trim = {100 20 100 0}, clip, width=0.5\columnwidth]{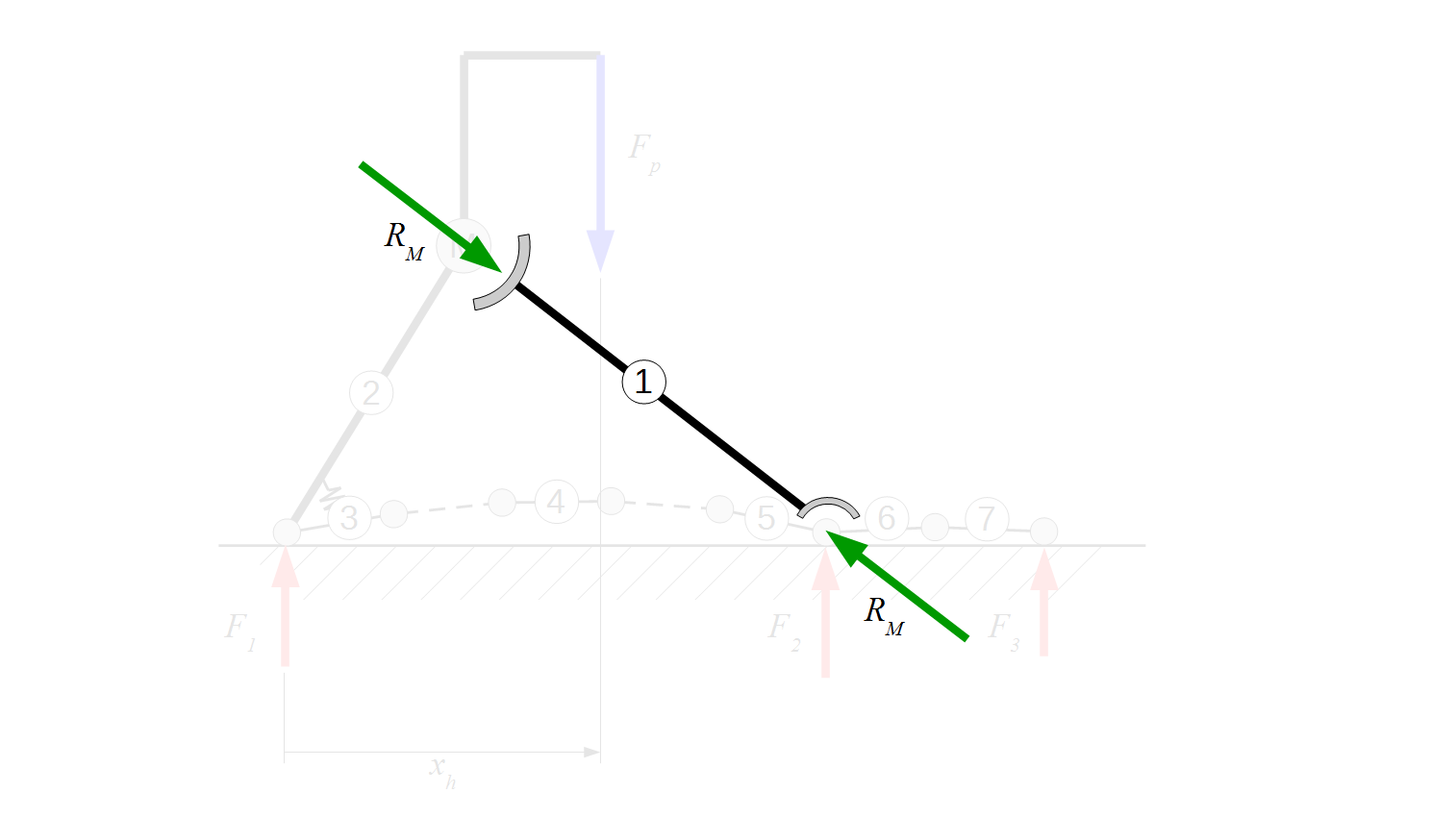}}
	\subfigure[body 2]{\includegraphics[trim = {100 20 100 0}, clip, width=0.5\columnwidth]{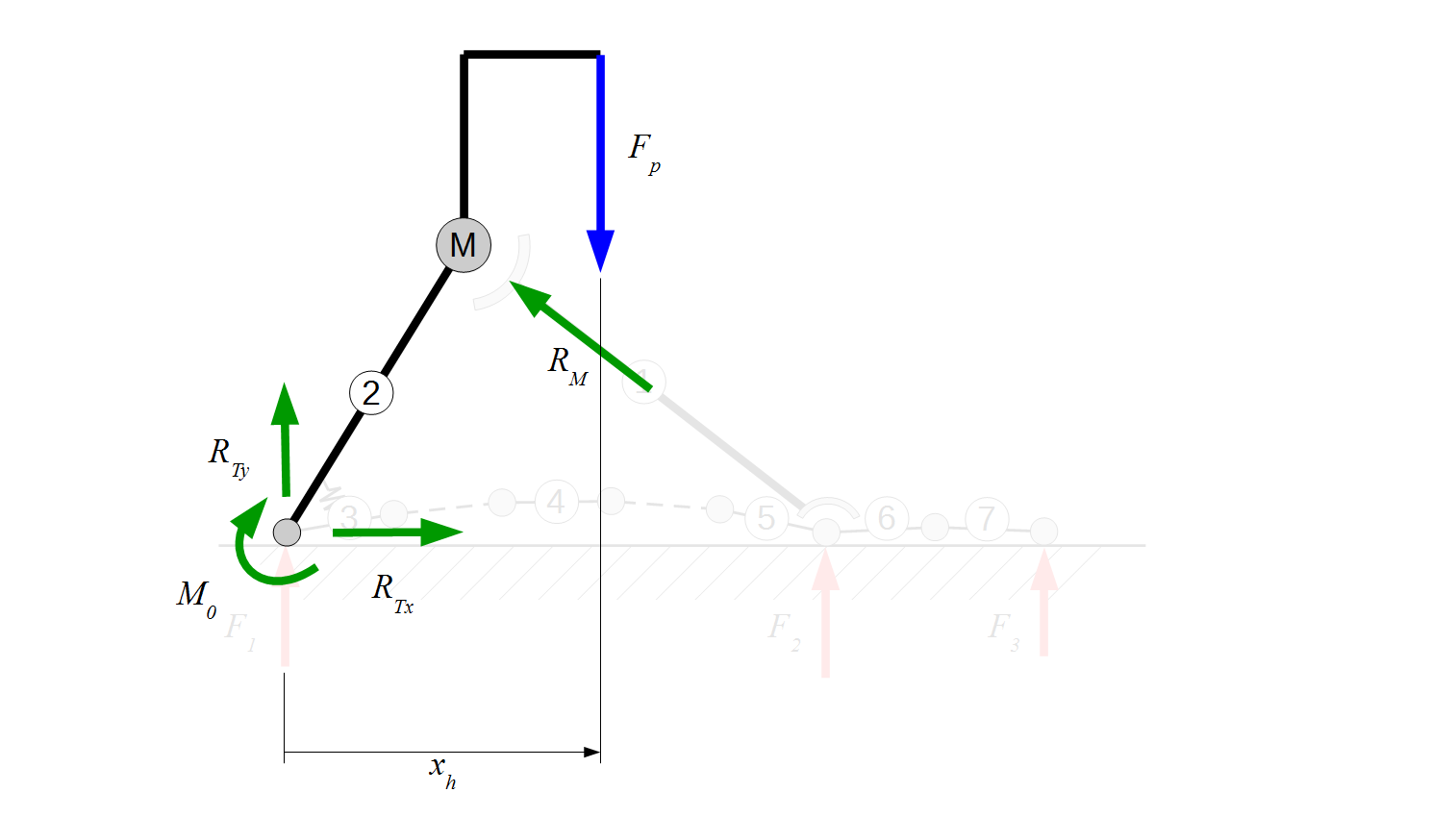}}
	\subfigure[body 3]{\includegraphics[trim = {100 20 100 0}, clip, width=0.5\columnwidth]{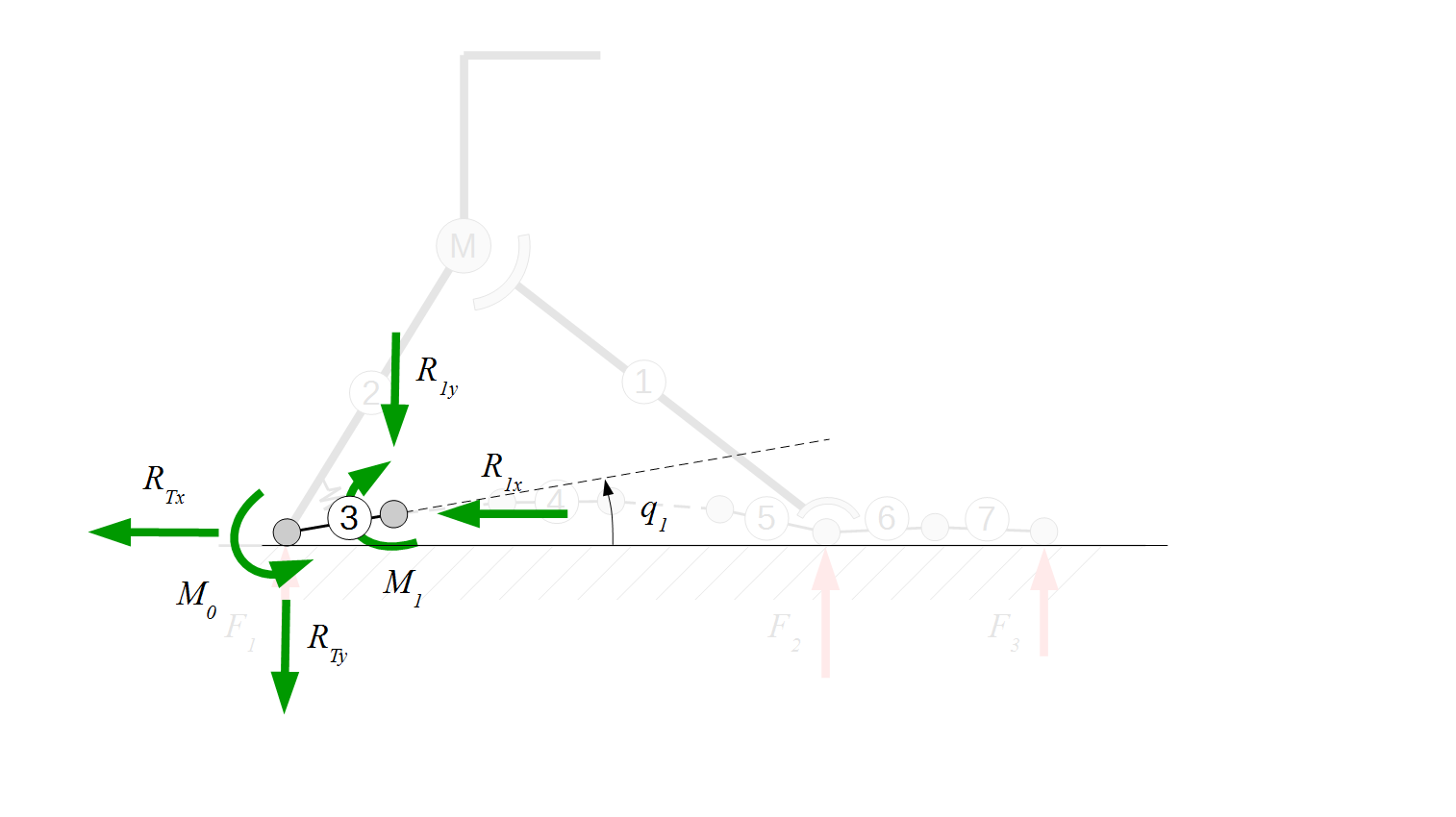}}
	\subfigure[body 4]{\includegraphics[trim = {100 20 100 0}, clip, width=0.5\columnwidth]{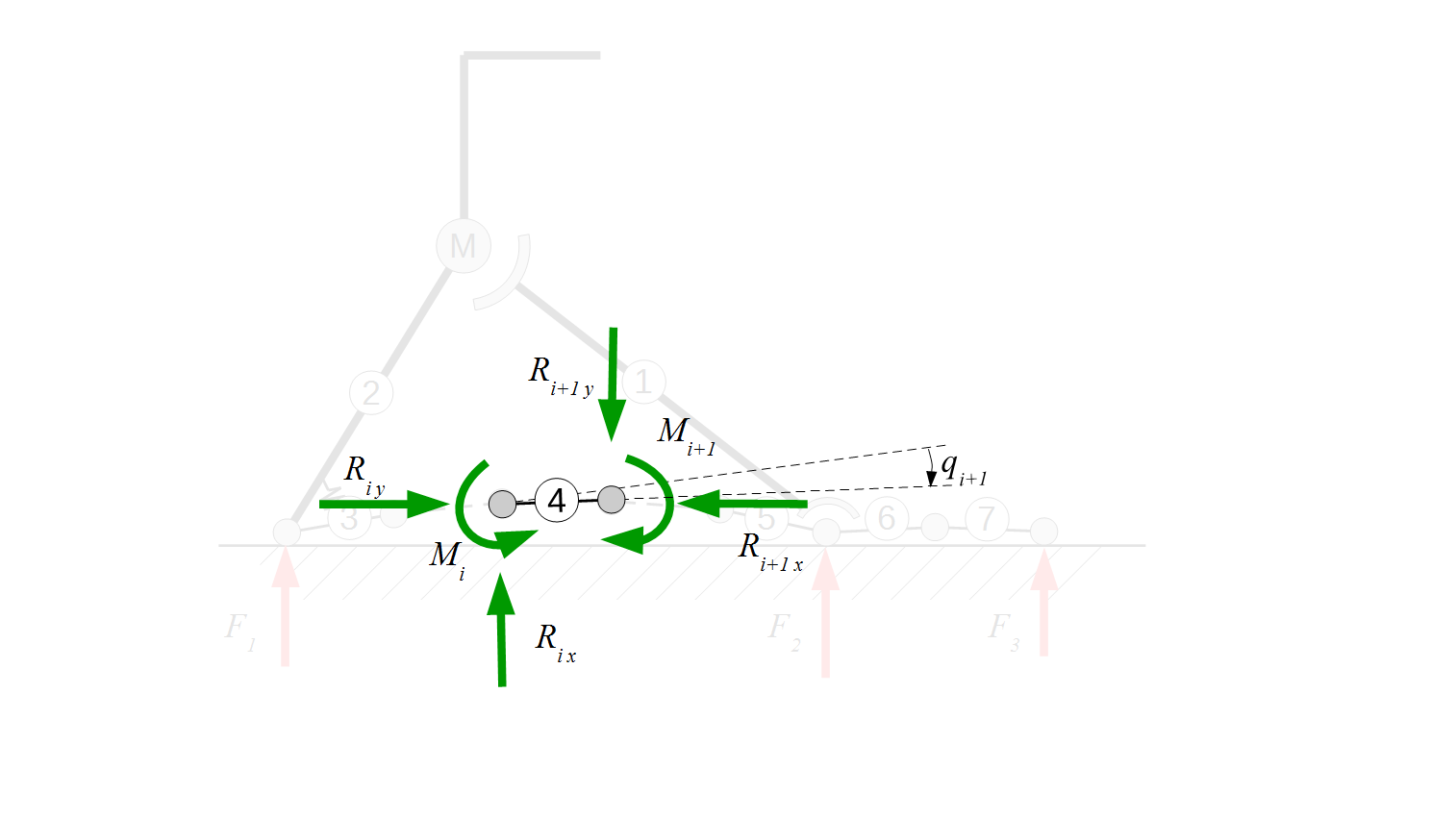}}
	\subfigure[body 5]{\includegraphics[trim = {100 20 100 0}, clip, width=0.5\columnwidth]{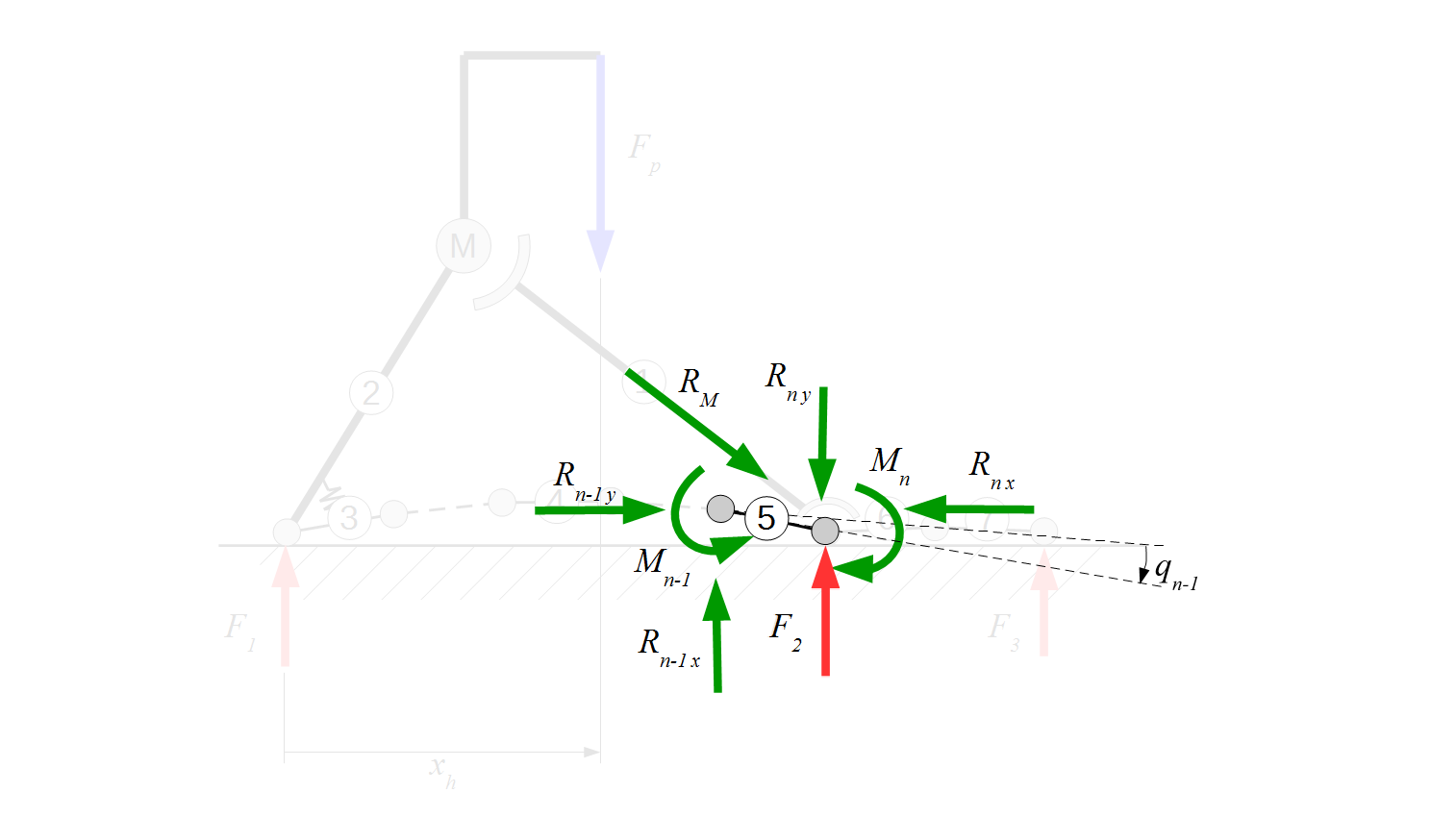}}\hspace{0.1\columnwidth}
	\subfigure[body 6]{\includegraphics[trim = {100 20 100 0}, clip, width=0.5\columnwidth]{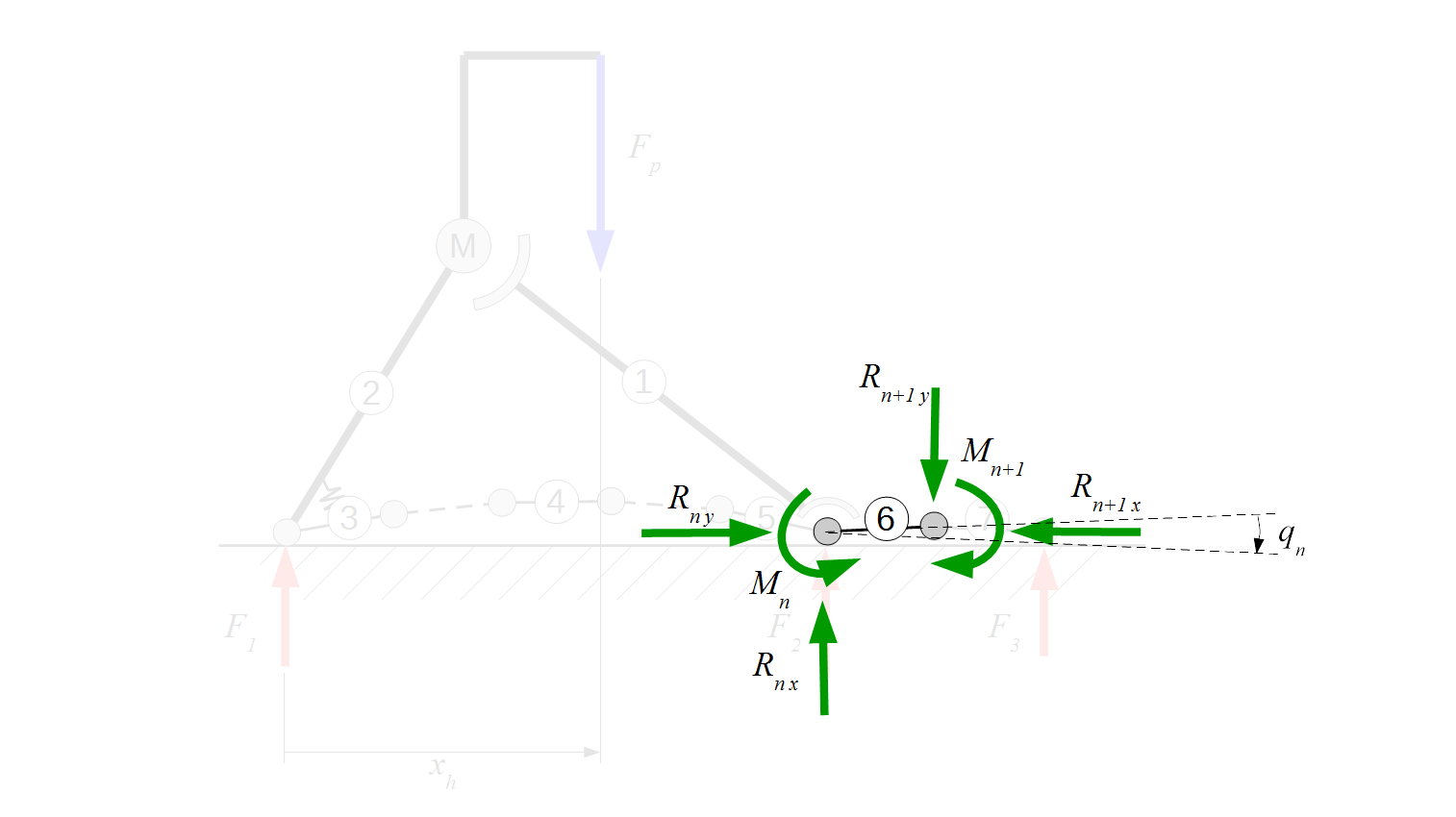}}\hspace{0.1\columnwidth}
	\subfigure[body 7]{\includegraphics[trim = {100 20 100 0}, clip, width=0.5\columnwidth]{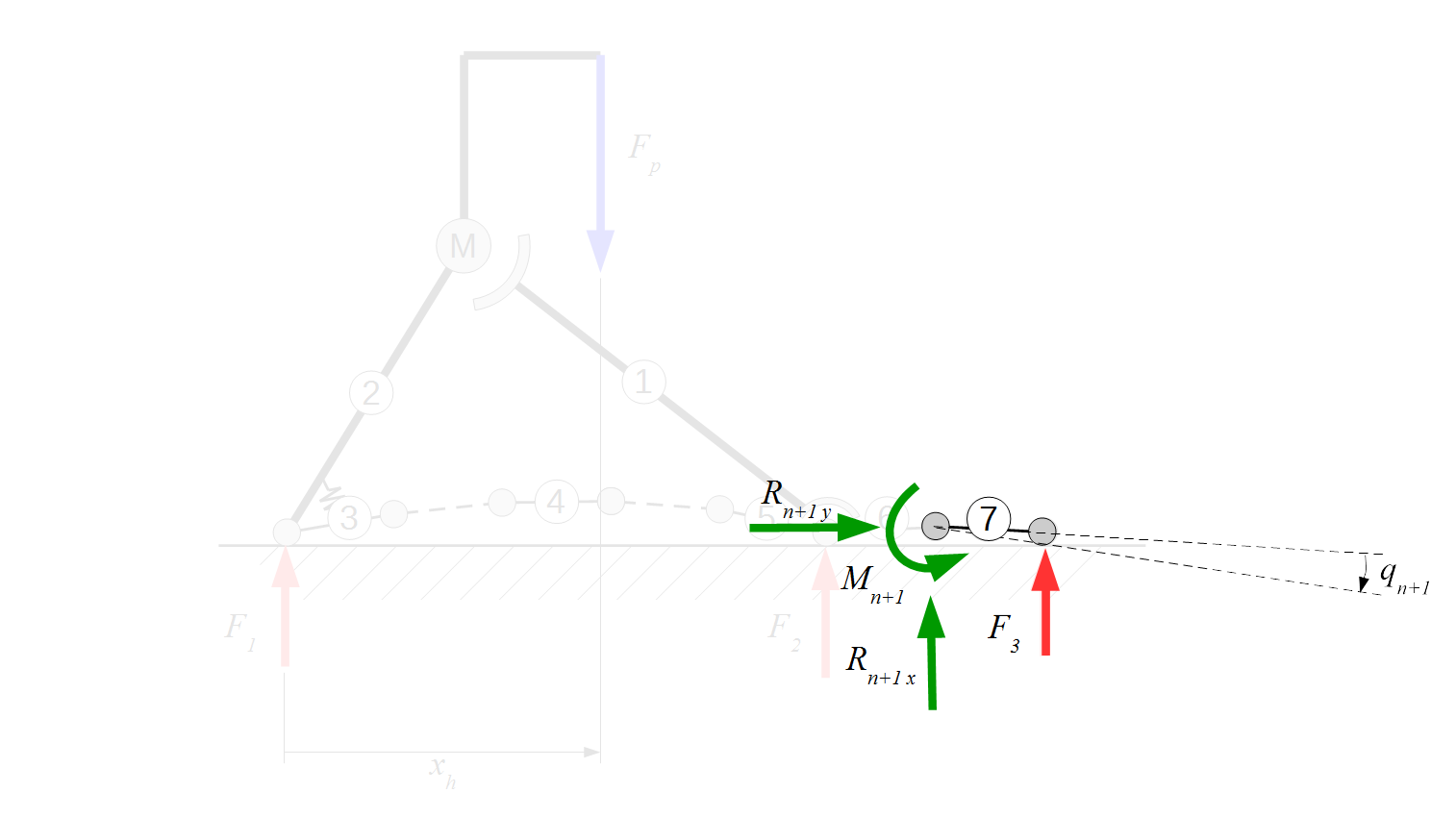}}
	\caption{Schematics of the free bodies composing a SoftFoot within the proposed mathematical model that captures the system behaviour when statically loaded. Each panel presents the detailed definition of variables for each rigid body presented in Fig. \ref{statics}. Note that body $3$ refers to a generic element of the sole that is not connected to bodies 1 and 2.  $F_1$, $F_2$, $F_3$ are the three ground reaction forces; $F_P$ is the force applied by the robot on the foot; $x_{\mathrm{H}}$ is the projection of the application point.}
	\label{fig:statics_foot}
\end{figure*}

Imposing the force and torque equilibria of the foot as a whole yields
\begin{equation} \label{eq:tot_foot}
\begin{split}
F_{\mathrm{P}} &= F_1 + F_2 + F_3, \\
F_{\mathrm{P}} &= \frac{b \, \mathrm{C}_\beta + a \, \mathrm{C}_\alpha}{x_{\mathrm{H}}} \, F_2 + \frac{b \, \mathrm{C}_\beta + a \, \mathrm{C}_\alpha + L \, (\mathrm{C}_n + \mathrm{C}_{n+1})}{x_{\mathrm{H}}} \, F_3 \; ,
\end{split}
\end{equation}
where $F_1$, $F_2$, $F_3$ are the three ground reaction forces, $F_P$ is the force applied by the robot on the foot, $x_{\mathrm{H}}$ is the projection of the application point, $a$, $b$, $\alpha$, and $\beta$ are geometric constants (see Fig. \ref{fig:statics_foot}). We also used the following notation for the sake of compactness $\mathrm{C}_\alpha = \cos{(\alpha)}$, $\mathrm{C}_\beta = \cos{(\beta)}$.

Through the force balance along $x$ of the rigid bodies composing the foot we obtain $\forall \; i \in \{1,\dots,n-1\}$
\begin{equation}\label{eq:x_foot}
	\begin{split}
		R_M \, \mathrm{C}_{\alpha} &= R_{\mathrm{T}}^\mathrm{x}, \\
		R_{\mathrm{T}}^\mathrm{x} &= R_{1}^\mathrm{x}, \\
		R_{i}^\mathrm{x} &= R_{i+1}^\mathrm{x}, \\
		R_{\mathrm{M}} \, \mathrm{C}_{\alpha} + R_n^\mathrm{x} &= R_{n + 1}^\mathrm{x}, \\
		R_{n + 1}^\mathrm{x} &= R_{n + 2}^\mathrm{x}, \\
		R_{n + 2}^\mathrm{x} &= 0,
	\end{split}
\end{equation}
where $R_{i}^\mathrm{x}$ is the horizontal reaction force exerted on the $i$\--th link by the $i-1$\--th link. $R_{\mathrm{T}}^\mathrm{x}$ and $R_M \, \mathrm{C}_{\alpha}$ are the horizontal reaction forces due to the interaction with the structure. Straightforward algebraic manipulations allow to transform \eqref{eq:x_foot} in
\begin{equation}\label{eq:x_s_foot}
\begin{split}
R_{\mathrm{T}}^\mathrm{x} &= + R_{\mathrm{M}} \, \mathrm{C}_{\alpha}, \\
R_{i}^\mathrm{x} &= -R_{\mathrm{M}} \, \mathrm{C}_{\alpha}, \\
R_{n + 1}^\mathrm{x} = R_{n + 2}^\mathrm{x} &= 0.
\end{split}
\end{equation}
Following the same steps, the balance along $y$ is 
\begin{equation}\label{eq:y_foot}
\begin{split}
R_M \, \mathrm{S}_{\alpha} + R_{\mathrm{T}}^\mathrm{y} &= F_{\mathrm{P}},  \\
R_{1}^\mathrm{y} &= F_1 - R_{\mathrm{T}}^\mathrm{y}, \\
R_{i}^\mathrm{y} &= R_{i+1}^\mathrm{y} \quad \forall \; i \in \{1,\dots,n-1\}, \\
R_n^\mathrm{y} - R_{\mathrm{M}} \, \mathrm{S}_{\alpha} &= R_{n + 1}^\mathrm{y} - F_2, \\
R_{n + 1}^\mathrm{y} &= R_{n + 2}^\mathrm{y}, \\
R_{n + 2}^\mathrm{y} &= -F_3,
\end{split}
\end{equation}
which again holds true $\forall \; i \in \{1,\dots,n\}$, and where $R_{i}^\mathrm{y}$ is the vertical reaction force exerted on the $i$\--th link by the $i-1$\--th link. $R_{\mathrm{T}}^\mathrm{y}$ and $R_M \, \mathrm{S}_{\alpha}$ are the vertical reaction forces due to the interaction with the structure. $F_1$, $F_2$, $F_3$ are the three ground reaction forces, $F_P$ is the force applied by the robot on the foot. Solving \eqref{eq:y_foot} yields
\begin{equation}\label{eq:y_s_foot}
\begin{split}
R_{\mathrm{T}}^\mathrm{y} &= - R_{\mathrm{M}} \, \mathrm{S}_{\alpha} + F_1 + F_2 + F_3, \\
R_{i}^\mathrm{y} &= + R_{\mathrm{M}} \, \mathrm{S}_{\alpha} - F_2 - F_3, \\ 
R_{n + 1}^\mathrm{y} = R_{n + 2}^\mathrm{y} &= -F_3 \, .
\end{split}
\end{equation}
Finally, we evaluate the torque balance for the bodies $3$, $4$, $5$, $6$, $7$
\begin{equation}\label{eq:m_s_foot}
	\begin{split}
		m_i - m_{i+1} + R_{i+1}^\mathrm{x} \, L \, \mathrm{S}_i - R_{i+1}^\mathrm{y} \, L \, \mathrm{C}_i &= 0, \\
		m_n - m_{n+1} + R_{n}^\mathrm{x} \, L \, \mathrm{S}_n - R_{n}^\mathrm{y} \, L \, \mathrm{C}_n &= 0, \\
		m_{n+1} - m_{n+2} + R_{n+2}^\mathrm{x} \, L \, \mathrm{S}_{n+1} - R_{n+2}^\mathrm{y} \, L \, \mathrm{C}_{n+1} &= 0, \\
		m_{n + 2} &= -F_3 \, L \, \mathrm{C}_{n+2} ,
	\end{split}
\end{equation}
where $m_{i}$ is the torque exerted on the $i$\--th link. $R_{i}^\mathrm{x}$ and $R_{i}^\mathrm{y}$ are the reaction forces exerted on the $i$\--th link by the $i-1$\--th link. $R_{\mathrm{T}}^\mathrm{y}$ and $R_M \, \mathrm{S}_{\alpha}$ are the vertical reaction forces due to the interaction with the structure. $F_1$,$F_2$,$F_3$ are the three ground reaction forces, $F_P$ is the force applied by the robot on the foot. We also use the abbreviations $\mathrm{S}_i = \sin{(\sum_{1}^{i} \, q_j)}$, $\mathrm{C}_i = \cos{(\sum_{1}^{i} \, q_j)}$.

\begin{figure*}[h]%
	\subfigure[0 Kg load]{\includegraphics[width=0.9\columnwidth]{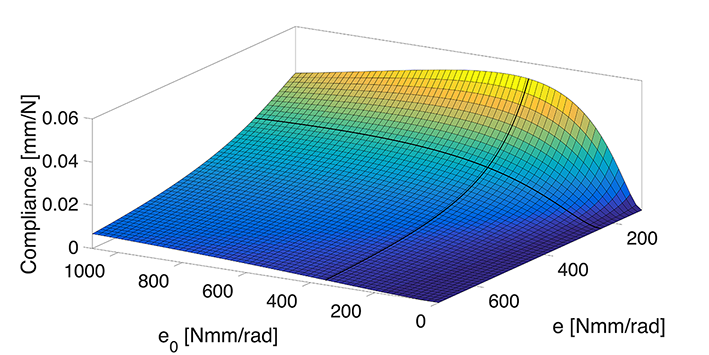}}
	\subfigure[1.5 Kg load]{\includegraphics[width=0.9\columnwidth]{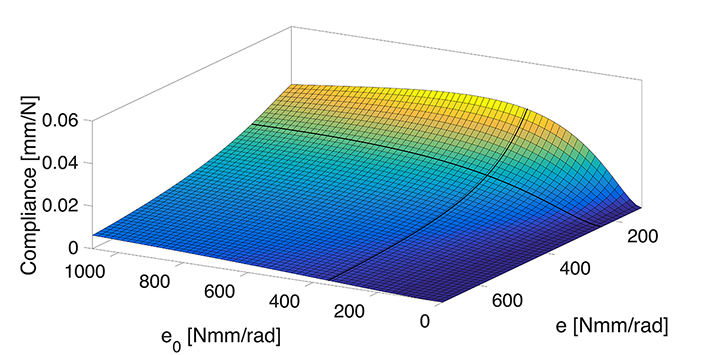}}
	\centering
	\subfigure[Configuration]{\includegraphics[width=\columnwidth]{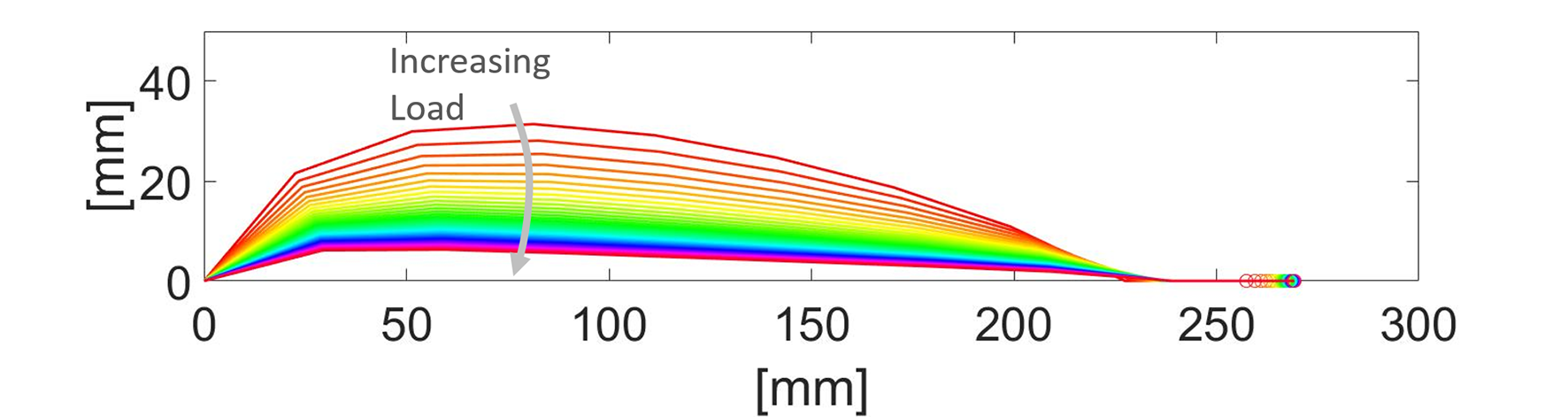}}
	\caption{Static equilibrium of the SoftFoot for various loads. Panel (a) shows foot compliance varying stiffness $\bar e$ and $e_0$, for a null load. Panel (b) presents the same for a 1.5Kg load, which is the one used in the experimental setup. In both figures the nominal stiffness values used in the prototype are underlined in black. Panel (c) shows the resulting configuration of the foot for a load from $0$Kg to $60$Kg.}%
	\label{fig:compliance_foot}%
\end{figure*}

By substituting \eqref{eq:x_s_foot} and \eqref{eq:y_s_foot} in \eqref{eq:m_s_foot}, we get $\forall \; i \in \{0,\dots,n\}$
\begin{equation}
	\begin{split}
		m_i - m_{i+1} &= R_{\mathrm{M}} \, \mathrm{C}_{\alpha} \, L \, \mathrm{S}_i + (R_{\mathrm{M}} \, \mathrm{S}_{\alpha} - F_2 - F_3) \, L \, \mathrm{C}_i, \\
		m_{n + 1} - m_{n + 2} &= -F_3 \, L \, \mathrm{C}_{n+1}, \\
		m_{n + 2} &= -F_3 \, L \, \mathrm{C}_{n+2} \; .
	\end{split}
\end{equation}
This can in turn be expressed in matrix form as
\begin{equation}\label{eq:eq_moment_3_7_matrix_foot}
M_{n+3}\, m + \mathbb{L}^{\mathrm{x}} + \mathbb{L}^{\mathrm{y}} = 0,
\end{equation}
where $m \in \mathbb{R}^{n+3}$ collects the terms $m_i$,
\begin{equation} \small
	M_m = 
	\begin{bmatrix}
		1 &-1 & 0 &\cdots &0 \\
		0 & 1 &-1 &\ddots &\vdots \\
		\vdots &\ddots &\ddots &\ddots &0 \\
		0 &\cdots & 0 &1 &-1 \\
		0 &\cdots & 0 &0 & 1 
	\end{bmatrix} \in \mathbb{R}^{m \times m} \; ,
\end{equation}
$\mathbb{L}^{\mathrm{x}} \in \mathbb{R}^{n+3}$ collects the terms $R_{\mathrm{M}} \, \mathrm{C}_{\alpha} \, L \, \mathrm{S}_i$, and $\mathbb{L}^{\mathrm{y}} \in \mathbb{R}^{n+3}$ collects the terms $ (R_{\mathrm{M}} \, \mathrm{S}_{\alpha} - F_2 - F_3) \, L \, \mathrm{C}_i$ and $-F_3 \, L \, \mathrm{C}_{n+2}$.

The torque $m$ is due to elastic effects and the tendon coupling, and can be explicitly evaluated to be $\forall \; i \in \{1,\dots,n + 2\}$
\begin{equation} \label{eq:m_e_foot}
\begin{split}
m_0 = - e_0 \, (q_0 - \beta) + r_0 \, T, \;\; m_i = - \, e_i \; q_i + r_i \, T ,
\end{split}
\end{equation}
where $m_{i}$ is the torque exerted on the $i$\--th link, $e_i$ is the elastic constant of the spring, $T$ is the tendon tension, $r_i$ is the transmission ratio (i.e. the pulley radius) on the $i$\--th joint, $\beta$ is the pretension of the first spring. In analogy to \cite{della2018toward}, we express \eqref{eq:m_e_foot} in matrix form as
\begin{equation}\label{eq:torque_el_matrix_foot}
	m = -E \, q - e_{\mathrm{E}} \, \beta + \mathfrak{R}^{\mathrm T}\,T
\end{equation}
where $e_E = \begin{bmatrix} e_0 & 0 & \cdots & 0 \end{bmatrix}^T$ maps the effect of the spring connected at the structure, $E \in \mathbb{R}^{n+3 \times n+3}$ collects all the elastic terms $e_i$, $\mathfrak{R} \in \mathbb{R}^{1 \times n+3}$ collects the transmission ratio $r_i$.

Finally we express the reaction force $R_M$ from the torque balance in the body 2
\begin{equation} \label{eq:RM_foot}
\begin{split}
&R_M \, \mathrm{C}_\alpha \, b \, \mathrm{S}_\beta + R_M \, \mathrm{S}_\alpha \, b \, \mathrm{C}_{\beta} - F_\mathrm{P} \, x_{\mathrm{H}} -  e_0 \, (q_0 - \beta) = 0 \; \Rightarrow\\
&R_M = F_\mathrm{P} \, \frac{x_{\mathrm{H}}}{b} \, \frac{1}{\mathrm{S}_{\alpha + \beta}}  -  \frac{ e_0 \, (q_0 - \beta)}{b} \, \frac{1}{\mathrm{S}_{\alpha + \beta}}
\end{split}
\end{equation}
where $F_P$ is the force exerted by the robot on the foot, $x_{\mathrm{H}}$ is the projection of the application point, $e_0$, $b$ $\alpha$ $\beta$ are geometric values and, $\mathrm{S}_{\alpha + \beta} = \sin{(\alpha + \beta)}$.

We also include the constrains imposed by the ground structure $L \sum_{0}^{n+2} \mathrm{S}_i = \delta $, and by the tendon $\mathfrak{R}\, q = \sigma$. Where $L$ is a phalanx length, $h$ is the terrain height, $\mathfrak{R}$ is the vector collecting the pulley radii, $\sigma$ is the tendon length.

Collecting Eq. \eqref{eq:tot_foot}, \eqref{eq:eq_moment_3_7_matrix_foot} and \eqref{eq:torque_el_matrix_foot} yields
\begin{equation}\label{eq:nl_balance_foot}
\begin{split}
	0 &= M\, (-E \, q - e_{\mathbf{E}} \, \beta + \mathfrak{R}\,T) + \mathbb{L}^{\mathrm{x}} + \mathbb{L}^{\mathrm{y}}, \\
	F_{\mathrm{P}} &= F_1 + F_2 + F_3, \\
	F_{\mathrm{P}} &= \frac{b \, \mathrm{C}_\beta + a \, \mathrm{C}_\alpha}{x_{\mathrm{H}}} \, F_2 \\
	&+ \frac{b \, \mathrm{C}_\beta + a \, \mathrm{C}_\alpha + L \, (\cos(q_n) + \cos(q_n + q_{n+1}))}{x_{\mathrm{H}}} \, F_3, \\
	\sum_{0}^{n+2} \mathrm{S}_i &= \frac{\delta}{L}, \\
	\mathfrak{R}^T \, q  &= \sigma.
\end{split}
\end{equation}
This is a set of $n+7$ nonlinear equations, in the $n + 7$ unknown quantities $q \in \mathbb{R}^{n+3}$, $F_1,F_2,F_3 \in \mathbb{R}$ and $T \in \mathbb{R}$. Although this paper will not follow this path, such a system can be solved numerically \cite{kelley2018numerical}, provide accurate description of the foot static behavior. We will investigate this possibility in future work.

However, we are interested here in deriving a tool for model based design of the considered class of systems. In other words, we want a closed form solution of \eqref{eq:nl_balance_foot}, analytically connecting all physical quantities\footnote{Namely, elastic terms $E$, geometry $\alpha,\beta,a,b$, and pulley radii $\mathfrak{R}$.} to how the foot configuration $q$ will respond to external stimuli. To achieve this goal, we accept to introduce some approximations. More precisely, we introduce the following small angles hypothesis 
\begin{equation}
	q_i \simeq 0, \quad \forall \; i \in \{0,\dots,n + 2\},
\end{equation}
which in turn implies
\begin{equation}
	\begin{split}
	\mathrm{S}_i \simeq \sum_{1}^{i} q_j , \;\; \mathrm{C}_i \simeq 1, \;\; \alpha \simeq \bar \alpha, \;\; \beta \simeq \bar \beta, \; T \simeq 0.
	\end{split} 
\end{equation}
Applying these simplifying assumptions to \eqref{eq:nl_balance_foot} yields
\begin{equation}
{
\begin{split}
    \mathbb{L}^{\mathrm{x}} &\simeq  { -{
    		\begin{bmatrix}
    		M_{n+2}^{-\mathrm T} &\bar{0}, \\
    		\bar{0}^{\mathrm T}  & 0
    		\end{bmatrix}} \, \frac{(F_{\mathrm{P}} \, X_{\mathrm{H}} + e_0 \, \beta) \, \mathrm{C}_{\bar \alpha} \, L}{b \, \mathrm{S}_{\bar \alpha + \bar \beta}}} \, q, \\
	\mathbb{L}^{\mathrm{y}} &\simeq -e_{\mathrm R} \, R_{\mathrm{M}} + e_{\mathrm F} \, F, \\
	\sum_{0}^{n+2} \mathrm{S}_i &\simeq c \, q, 
\end{split}}
\end{equation}
where
\begin{equation}
	c^T = 
	\begin{bmatrix}
		n+3 \\ \vdots \\ 1 
	\end{bmatrix}, \quad
	e_{\mathrm{R}} = L \, \mathrm{S}_\alpha \,
	\begin{bmatrix}
	1      \\ 
	\vdots \\
	1      \\
	0      \\
	0      
	\end{bmatrix}, \quad
	e_{\mathrm{F}} = L \,
	\begin{bmatrix}
	0      &1      &1 \\ 
	\vdots &\vdots &\vdots \\
	0      &1      &1 \\
	0      &0      &1 \\
	0      &0      &1
	\end{bmatrix}.
\end{equation}
To obtain a more compact structure of \eqref{eq:nl_balance_foot}, we reduce the set of unknown variables. We explicit the contact force from
		$F_{\mathrm{P}} = \frac{\sqrt{n^2 \, L^2 + \delta^2}}{x_{\mathrm{H}}} \, F_2 + \frac{\sqrt{(n + 2)^2 \, L^2 + \delta^2}}{x_{\mathrm{H}}} \, F_3$ and 
		$F_{\mathrm{P}} = F_1 + F_2 + F_3 $ ,
obtaining $F_2 = \frac{F_{\mathrm{P}} \, x_{\mathrm{H}} - F_3 \, \sqrt{(n + 2)^2 \, L^2 + \delta^2}}{\sqrt{n^2 \, L^2 + \delta^2}}$.
This allow to re\--write \eqref{eq:nl_balance_foot} in the linear form 
\begin{equation}\label{eq:block_stati_bil_foot}
	\begin{bmatrix}
		-\mathbb{E} & [\mathfrak{R}^T \; d^T] \\
		\begin{bmatrix} \mathfrak{R} \\ c \end{bmatrix} & \emptyset
	\end{bmatrix} \begin{bmatrix} q \\ 
	\begin{bmatrix}
		T \\
		L\,F_3
	\end{bmatrix}
\end{bmatrix}= \begin{bmatrix}
m_{\mathrm{E}} \\
\begin{bmatrix}
	\sigma \\
	\frac{\delta}{L}
\end{bmatrix}
\end{bmatrix}.
\end{equation}
where we defined
\begin{equation}\label{eq:me_and_other}
\begin{split}
d^T &= \frac{1}{L}M_{n+3}^{-1}\,e_{\mathrm{F}} \, \begin{bmatrix}
0 \\ -\,\frac{\sqrt{(n + 2)^2 \, L^2 + \delta^2}}{\sqrt{n^2 \, L^2 + \delta^2}} \\ 1
\end{bmatrix}\\
\mathbb{E} &= E \, + \, M^{-1}_{n+3} \, \begin{bmatrix}
M_{n+2}^{-\mathrm T} &\bar{0},\\
\bar{0}^{\mathrm T}  & 0
\end{bmatrix} \, \frac{(F_{\mathrm{P}} \, X_{\mathrm{H}} + e_0 \, \beta) \, \mathrm{C}_{\bar \alpha} \, L}{b \, \mathrm{S}_{\bar \alpha + \bar \beta}},\\
m_{\mathrm{E}} &= M^{-1}_{n+3} \, e_{\mathrm{R}} \, (\frac{F_\mathrm{P} \, x_{\mathrm{H}} \, + \, e_0 \, \beta}{b \, \mathrm{S}_{\bar \alpha + \bar \beta}} + \frac{-F_{\mathrm{P}} \, x_{\mathrm{H}}}{\sqrt{n^2 \, L^2 + h^2} \, \mathrm{S}_{\alpha}}) + e_{\mathbf{E}} \, \beta \, .
\end{split}
\end{equation}
Through block inversion, \eqref{eq:block_stati_bil_foot} leads to
\begin{equation}\label{eq:two_sigma_foot}
\begin{split}
	q = -\, (I &- \mathbb{E}^{-1} \, [\mathfrak{R}^T \; d^T](\begin{bmatrix}
	\mathfrak{R} \\
	c
	\end{bmatrix}\mathbb{E}^{-1} \, [\mathfrak{R}^T \; d^T])^{-1}
		\begin{bmatrix}
			\mathfrak{R} \\
			c
		\end{bmatrix}) \, \mathbb{E}^{-1} \, m_{\mathrm{E}}\\
	  &+ \mathbb{E}^{-1} \, [\mathfrak{R}^T \; d^T] \, (\begin{bmatrix}
	  \mathfrak{R} \\
	  c
	  \end{bmatrix}\mathbb{E}^{-1} \, [\mathfrak{R}^T \; d^T])^{-1}	\begin{bmatrix}
	  	\sigma \\
	  	\frac{\delta}{L}
	  \end{bmatrix}.
\end{split}%
\end{equation}
This equation expresses in closed form the relation that we were looking for, connecting foot configuration $q$, to external forces applied on the robot {\footnote{Note that $m_{\mathrm{E}}$ is defined in \eqref{eq:me_and_other}.}, and constructive parameters. 

\subsection{Example of application: study of the foot compliance to compression}\label{sec:example}
We consider as example of use of \eqref{eq:two_sigma_foot} the study of the foot compliance to compression $\Sigma_{\mathrm{H}}^{-1}$ w.r.t. force $F_{\mathrm P}$, in the case of no tendon pretension (i.e. $\sigma = 0$) and flat ground (i.e. $h = 0$).
By definition
\begin{equation}
		\Sigma_{\mathrm{H}}^{-1} = \frac{\partial \, h(q)}{\partial \, F_{\mathrm{H}}} = J(q) \, \frac{\partial \, q}{\partial \, F_{\mathrm{H}}}, \,
\end{equation}
where $J(q) = L \, \mathrm{C}_{\bar \alpha}^2 \, \begin{bmatrix}	\mathrm{S}_0 & \mathrm{S}_1 & \dots & \mathrm{S}_{n+2} \end{bmatrix}$. Thus for $q_i = 0 \; \forall \, i$ the foot has that the compliance is null, i.e. the foot behaves as a rigid foot, correctly supporting the weight of the robot. Furthermore, through the opportune choice of the spring terms and the pulley radii we can design the foot stiffness in the other configurations. Taking for example $\bar\beta = \pi/3$, $\bar\alpha = \pi/6$, ($\mathrm{S}_{\bar \alpha + \bar \beta} = 1$), $n = 6$, $E = \bar e \, I$, $x_{\mathrm{H}} = b$, we obtain the compliance in Fig. \ref{fig:compliance_foot}. The same figure also shows some examples of steady state configurations as predicted by the model.

\begin{figure*}[th!]
	\centering
	\subfigure[]{\includegraphics[width = 0.95\columnwidth]{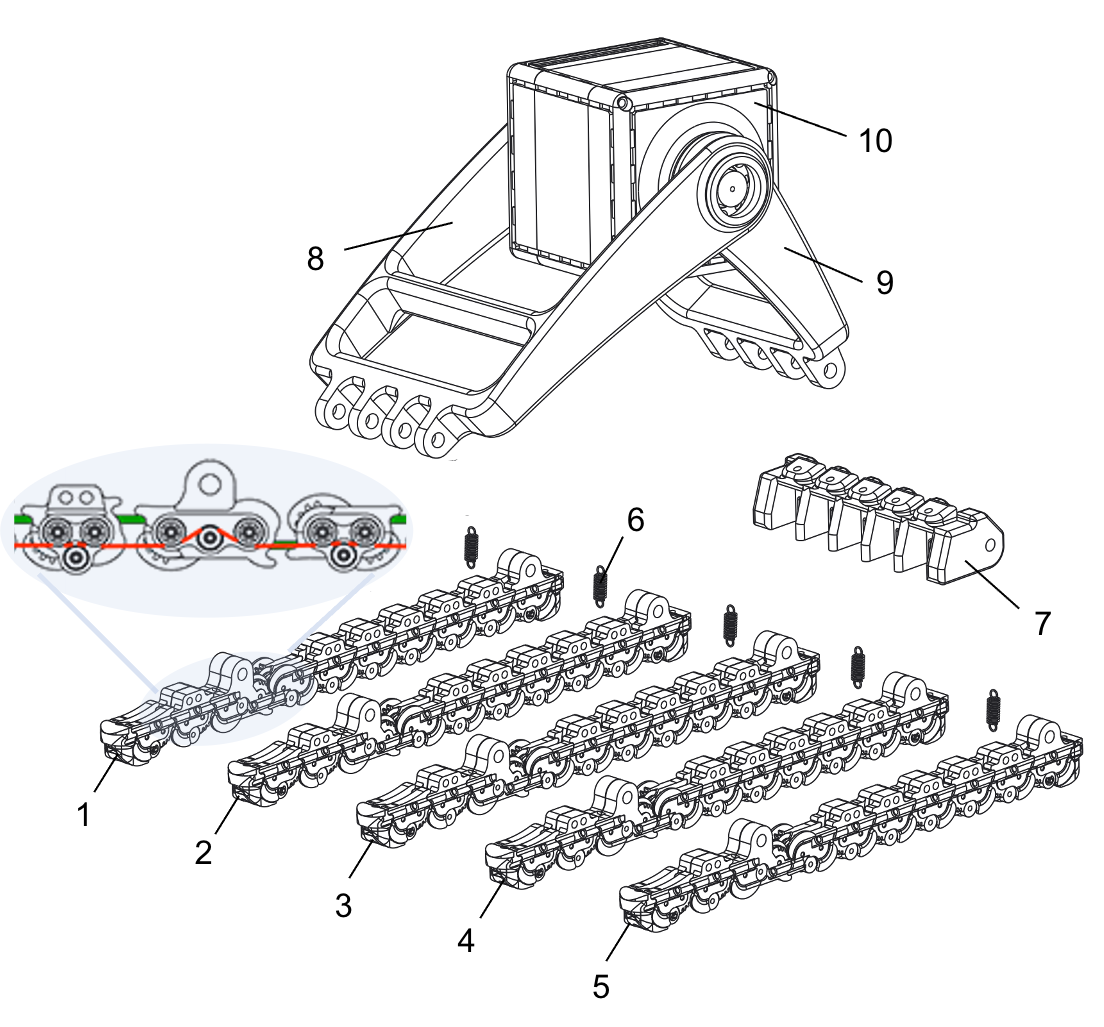}}\label{fig:fig_a}
		\subfigure[]{\includegraphics[width = 0.95\columnwidth]{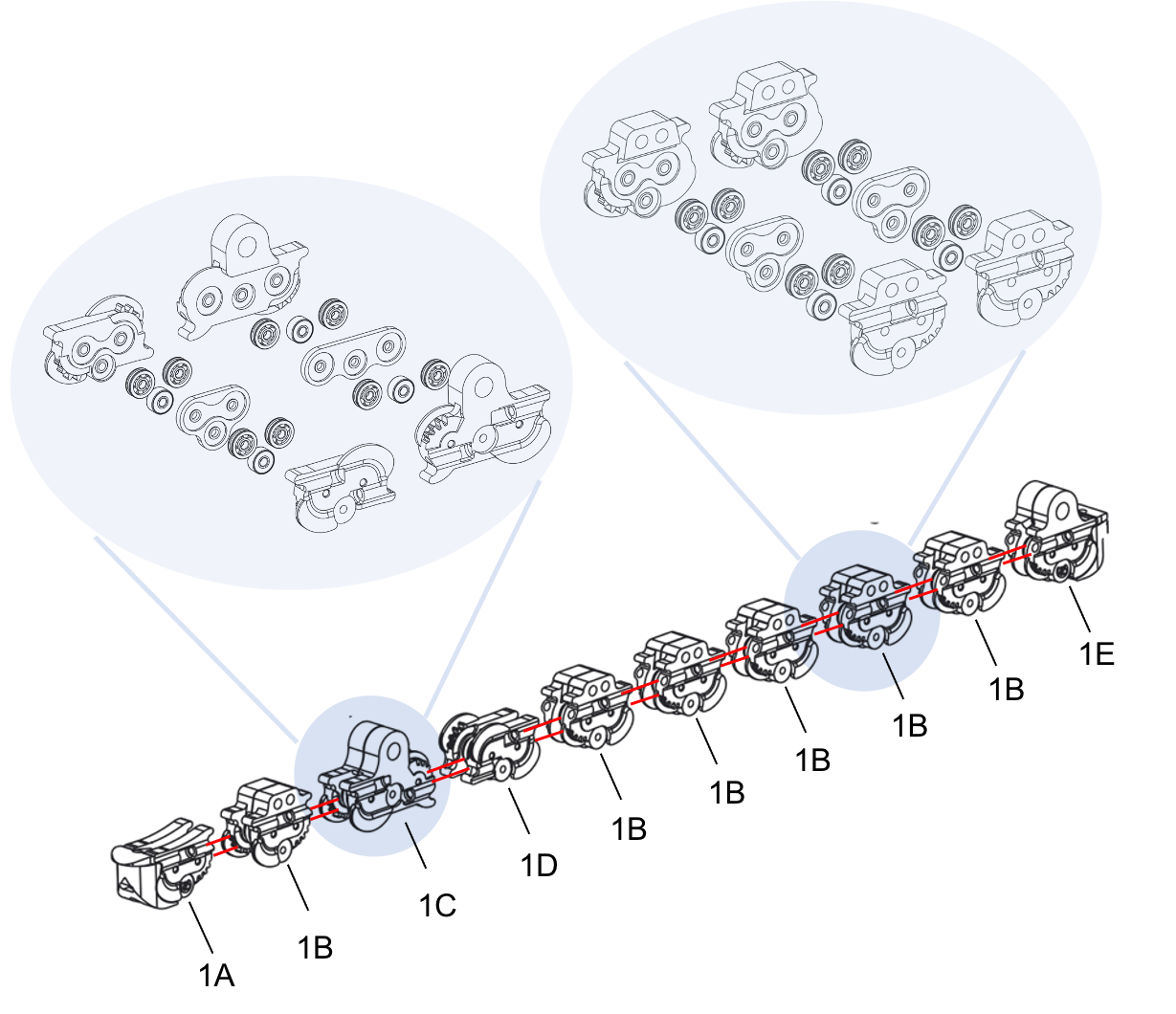}}\label{fig:cad}
	\caption{Exploded view of 3D CAD model of the SoftFoot prototype $(a)$ and of detail of one flexible structure (i.e. 1) which forms the sole of the SoftFoot $(b)$. The elastic bands are highlighted in green, while the cable used for the routing in red. The exploded view of the modules design (i.e. 1C or 1B) is presented in the light blue circle.}
	\label{fig:fing_routing}
\end{figure*}


\section{SOFTFOOT DESIGN}\label{sec:design}
\begin{table*}[h]
\centering
\caption{Specifics of the SoftFoot and other robotic feet design from the state of the art}
\label{tab:feet_comp}
\setlength{\tabcolsep}{3pt}
\begin{tabular}{l|c|c|c|c|c|c}
& Foot Structure & Foot DOF & Active/Passive & Sensors & Adaptability & Capability to Absorb Impact\\
\hline \hline
\textbf{SoftFoot} & Soft Articulated  &  47   & Passive  & NO & YES	& YES \\
\hline
Humanoid Foot \cite{davis2010design} &	Articulated & 20 & Active & YES & YES & YES \\
WABIAN-2R Foot \cite{kang2010realization} & Unique segment & 0 & Passive & YES & NO & NO \\
Salamander Foot \cite{paez2019adaptive} & Soft Articulated &  9 & Active & NO & YES & YES \\
Kengoro 3 Foot \cite{asano2016human} & Rigid joint and 2 spring elements & 16 & Active & YES & YES & NO \\
Adaptive Planar Foot  \cite{kaslin2018towards} & Unique segment with rubber sole &  0 & Passive & NO & YES & YES\\
Jamming Foot   \cite{hauser2018compliant} & Soft Continuous & $\infty$ & Active & YES & YES & YES  \\
\hline
\end{tabular}
\end{table*}

\subsection{Mechanical Design}

The SoftFoot design takes inspiration from the architecture and main features of the human foot, incorporating a reliable and robust shape-morphing design. 
As presented in the exploded view of Fig.~\ref{fig:fing_routing}, the SoftFoot is composed of five modular and flexible plantar structures (1-5) connected by three rigid elements (7, 8, and 9).
The two central rigid components (8 and 9) recreate the shape of the anatomical longitudinal arch and are interconnected with a revolute joint. This design allows SoftFoot to provide proper support and stability to the robot during locomotion. The actuator (10) replaces the ankle joint of the foot and is rigidly linked to part (9) through the output shaft.  The five modular structures (1-5) are designed to create a foot sole that can withstand the weight of the robot, while also being compliant and adaptive to different terrain shapes. In the back side of the foot, a group of five springs (6) connects each modular structure to the central rigid component (9), which can store and release energy during the push-off.  This design presents a stiffening-by-compression behaviour, being compliant when the load is low (i.e. when contact occurs) and rigid when the load grows, to support the weight of the robot. Each structure consists of a series of modules made by rapid prototyping techniques and connected by a pair of customized elastic bands (highlighted in green in the detail of Fig.~\ref{fig:fing_routing} (a)), which gives intrinsic compliance to the whole sole. Additionally, this enhances the robustness of the foot and resilience to impact with the environment. The design of these modules takes inspiration from those used in the Pisa/IIT SoftHand \cite{catalano2014adaptive}. They are an evolution of Hillberry's joint \cite{hillberry1976rolling}, designed to be very robust and easy to combine.  A series of bearings with customized design (Fig.~\ref{fig:fing_routing} (b)) is placed inside each module and hosts a tendon (highlighted in red) that runs across the whole flexible structure with a specific route to allow proper force distribution. This design ensures to distribute the force in a way that minimizes damage to the robot. The five modular structures are held together by a rigid part (7), placed on the back of the foot, which resembles the Calcaneus of the human foot.  The design is easy to scale and customizable, enabling the foot to be tailored to specific applications or environments. 
We used \eqref{eq:two_sigma_foot} to dimension the physical parameters of our system. We select them as in Sec. \ref{sec:example}, with pulleys of $1.5 mm$ radius. We varied $\Bar{e}$ across the ones we could achieve with elements we could realize with our injection-mold process. We ended up selecting a stiffness of $1.2 mm$ as a trade-off between firmness and softness, so that the foot finds itself halfway its compression width when loaded with $25$Kg on flat ground as we estimated the weight of a small scale robot to be around $50$Kg.

\subsection{Comparative analysis}

Table~\ref{tab:feet_comp} presents a comparative overview of the main technical characteristics of the the SoftFoot and a selection of robotic feet from the state of the art. This selection includes several types of designs (from unique segment feet to articulated or compliant solutions) that try to mimic the role of phalanges and metatarsal joints of the human foot. 
The absence of motors or active sensors at the foot level, combined with the foot flexibility, allows to increase its capability to absorb impact and interact safely with different terrain types. 
While compliant solutions are already present in the state of the art, the proposed design is the only one that incorporates a soft articulated foot sole with a passive mechanism. 
Moreover, besides soft continuous solutions, the SoftFoot features the highest number of DOFs in the foot sole.

\section{EXPERIMENTAL VALIDATION}\label{sec:exp}

\begin{figure*}[t]
\centering
	\includegraphics[width =0.8\textwidth]{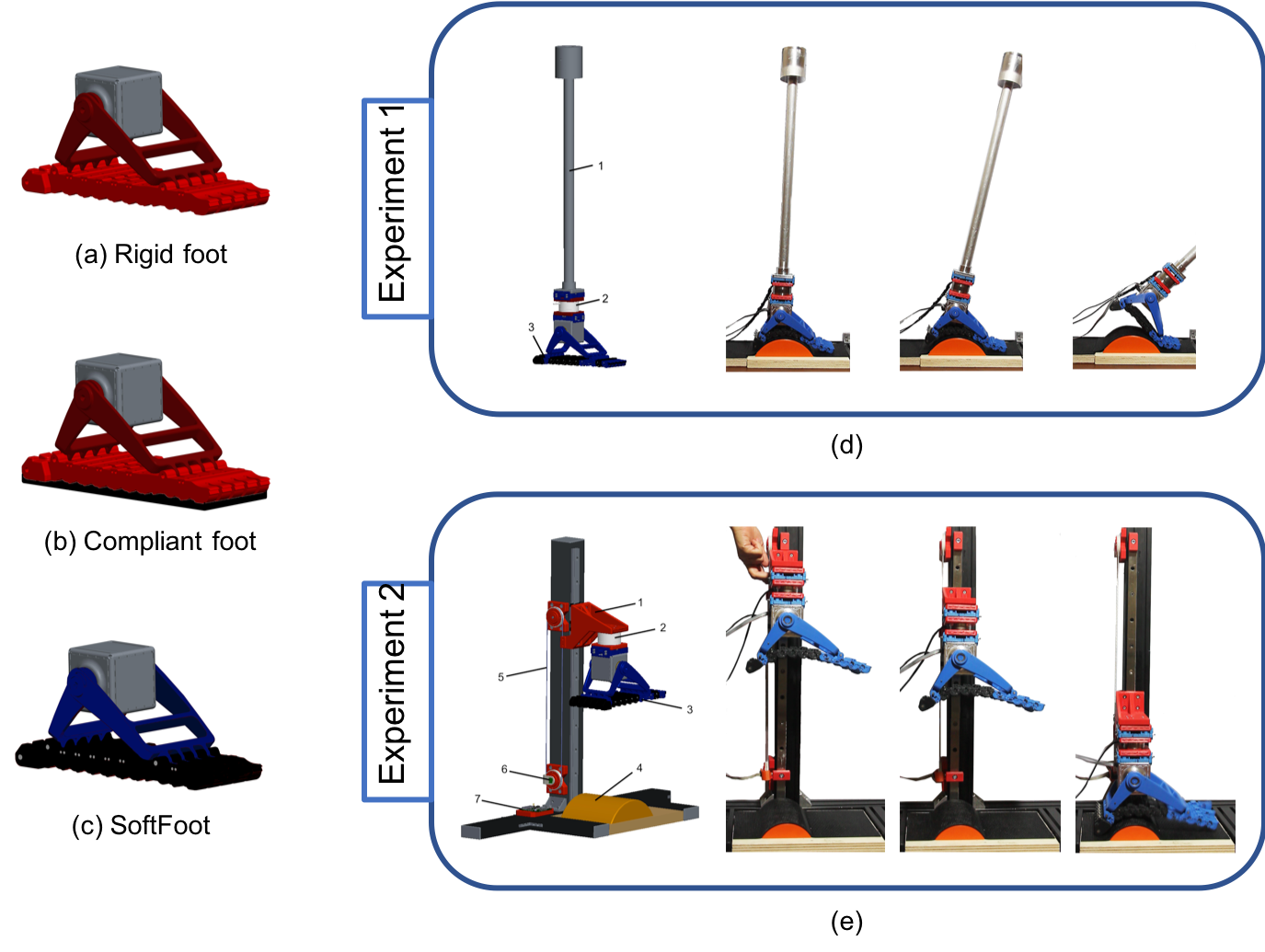}
	\caption{The experimental validation was conducted with three different foot design, with the same weight and size: a rigid foot (a), a compliant foot (b) and the SoftFoot (c). The design robustness and performance validation included two main aspects: push-off phase (d) and impact test (e).}
	\label{fig:iexp}
\end{figure*}

To evaluate the performance and ability of the SoftFoot in mechanically managing the complexity introduced by uneven terrains, we considered two experiments:
\begin{itemize}

\item \emph{Experiment 1:} The first experiment was designed to analyze the foot stability through the empiric evaluation of the support polygon. The experimental setup used in this test is shown in Fig.~\ref{fig:iexp} (d). It consists of a rigid aluminum bar (1) connected with the ankle joint of the foot (3). Between the foot and the bar, there is an ATI Mini45 Fore/Torque sensor (2). A mass is placed on the top of the bar. The total weight on top of the sensor is equivalent to $15$N. The center of gravity is fixed at $250$mm from the ankle axis.

While the foot was placed in a predefined position on an obstacle, the motor at the ankle was controlled to move the bar either in the backward or forward directions very slowly until falling occurred. An example of the test is shown in the photo sequence of Fig.~\ref{fig:iexp} (d), where the SoftFoot is tested on a round obstacle. 
The bar movement was very slow, i.e. $0.1\,\frac{\text{deg}}{\text{s}}$, in order to avoid dynamic effects. To measure the extension of the support polygon readings of the F/T sensors and of the ankle joint encoder were combined to project the forces on vertical and frontal directions of the sagittal plane, obtaining Fx and Fz, respectively, and then using the definition of CoP with respect to the underlying flat base.\\
\begin{table*}
\centering
\begin{tabular}{c|p{4cm}|p{4cm}|p{4cm}|}
&
Rigid
\includegraphics[align=c, width = 0.20\columnwidth]{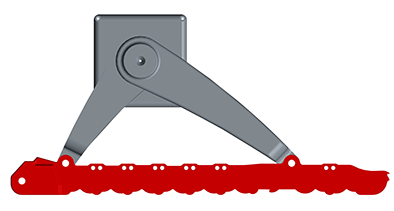}
&
Compliant
\includegraphics[align=c, width = 0.20\columnwidth]{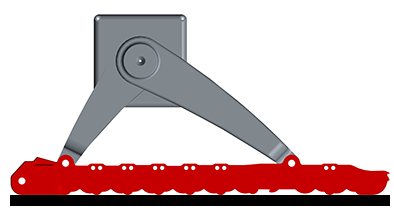}
& 
SoftFoot
\includegraphics[align=c, width = 0.20\columnwidth]{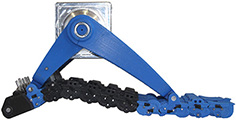}\\
%
\hline 1
\includegraphics[align=c,width = 0.20\columnwidth]{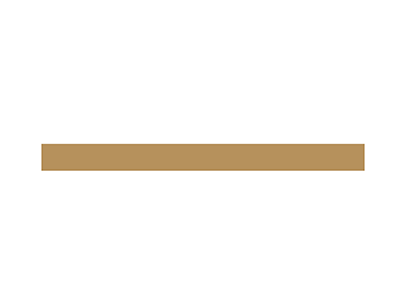}
&
\parbox{0.2\columnwidth}{
\centering
\includegraphics[align=c, width = 0.20\columnwidth]{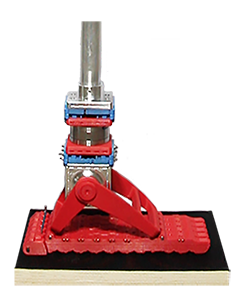}\\
219 mm
0 \degree
\vspace{0.03 cm}}
& 
\parbox{0.2\columnwidth}{
\centering
\includegraphics[align=c, width = 0.20\columnwidth]{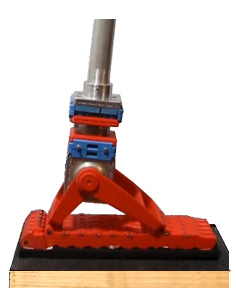}\\
116 mm
0 \degree
\vspace{0.03 cm}}
&
\parbox{0.2\columnwidth}{
\centering
\includegraphics[align=c, width = 0.20\columnwidth]{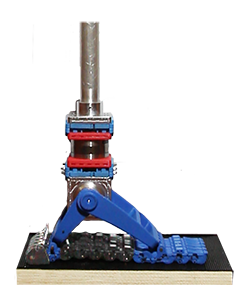}\\
134 mm
0 \degree
\vspace{0.03 cm}} 
\\
\hline 2
%
\includegraphics[align=c, width = 0.20\columnwidth]{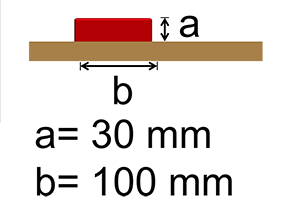}
&
\parbox{0.2\columnwidth}{
\centering
\includegraphics[align=c, width = 0.20\columnwidth]{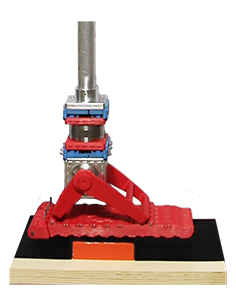}\\
67 mm
3 \degree
\vspace{0.03 cm}}
&
\parbox{0.2\columnwidth}{
\centering
\includegraphics[align=c, width = 0.20\columnwidth]{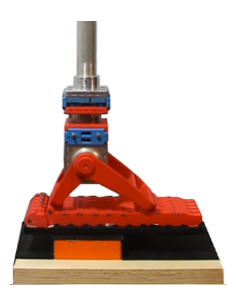}\\
45 mm
1 \degree
\vspace{0.03 cm}}
&
\parbox{0.2\columnwidth}{
\centering
\includegraphics[align=c, width = 0.20\columnwidth]{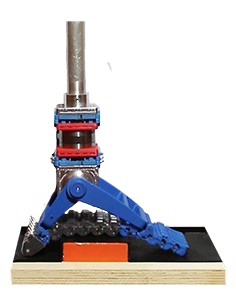}\\
94 mm
4 \degree
\vspace{0.03 cm}}  
\\
\hline 3
%
\includegraphics[align=c, width = 0.20\columnwidth]{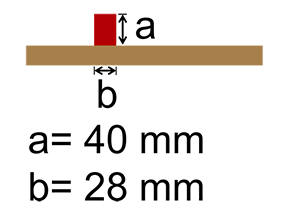}
&
\parbox{0.2\columnwidth}{
\centering
\includegraphics[align=c, width = 0.20\columnwidth]{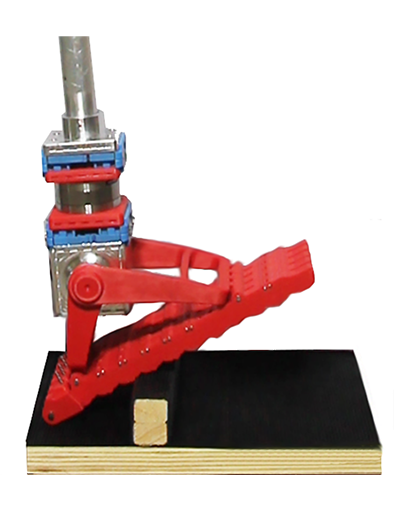}\\
40 mm
-26 \degree
\vspace{0.03 cm}}
\parbox{0.2\columnwidth}{
\centering
\includegraphics[align=c, width = 0.20\columnwidth]{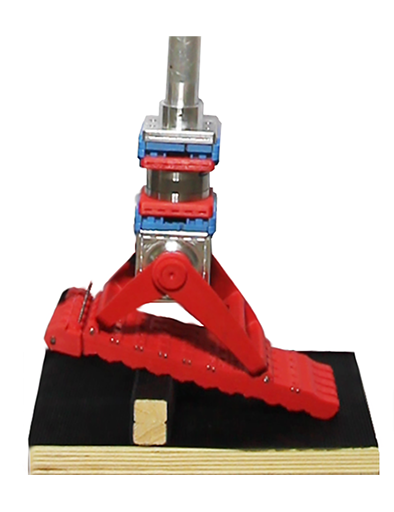}\\
79 mm
23 \degree
\vspace{0.03 cm}}
&
\parbox{0.2\columnwidth}{
\centering
\includegraphics[align=c, width = 0.20\columnwidth]{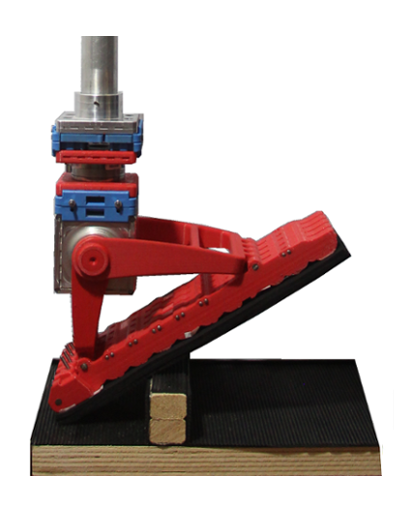}\\
63 mm
-22 \degree
\vspace{0.03 cm}}
\parbox{0.2\columnwidth}{
\centering
\includegraphics[align=c, width = 0.20\columnwidth]{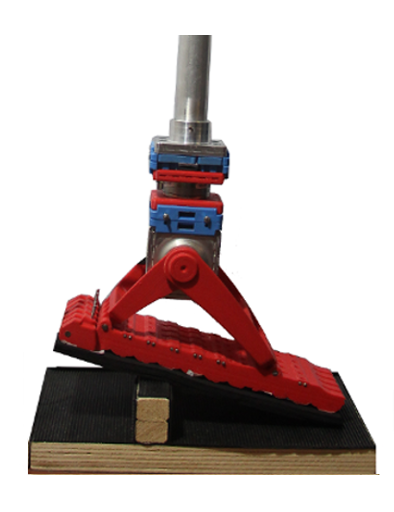}\\
101 mm
13 \degree
\vspace{0.03 cm}}&
\parbox{0.2\columnwidth}{
\centering
\includegraphics[align=c, width = 0.20\columnwidth]{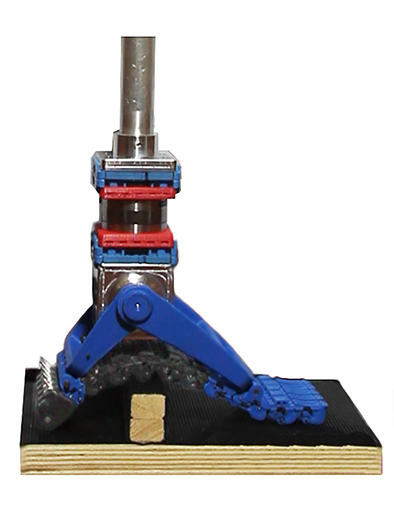}\\
47 mm
3 \degree
\vspace{0.03 cm}}  \\
\hline 4
%
\includegraphics[align=c, width = 0.20\columnwidth]{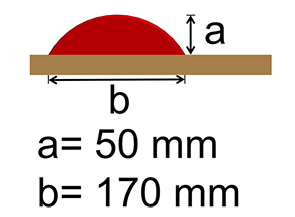}
&
\parbox{0.2\columnwidth}{
\centering
\includegraphics[align=c, width = 0.20\columnwidth]{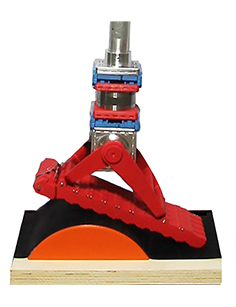}\\
17 mm
25 \degree
\vspace{0.03 cm}}
&
\parbox{0.2\columnwidth}{
\centering
\includegraphics[align=c, width = 0.20\columnwidth]{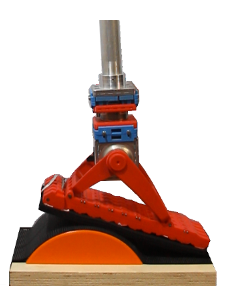}\\
45 mm
-27 \degree
\vspace{0.03 cm}}
\parbox{0.2\columnwidth}{
\centering
\includegraphics[align=c, width = 0.20\columnwidth]{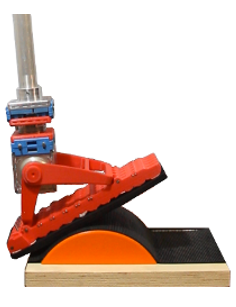}\\
86 mm
17 \degree
\vspace{0.03 cm}}
& 
\parbox{0.2\columnwidth}{
\centering
\includegraphics[align=c, width = 0.20\columnwidth]{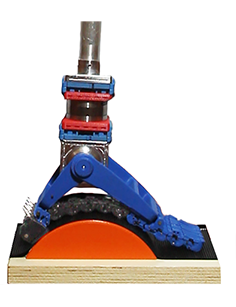}\\
53 mm
10 \degree
\vspace{0.03 cm}} \\
%
\hline 5
\includegraphics[align=c, width = 0.20\columnwidth]{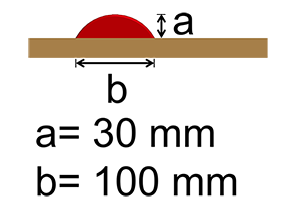}
&
\parbox{0.2\columnwidth}{
\centering
\includegraphics[align=c, width = 0.20\columnwidth]{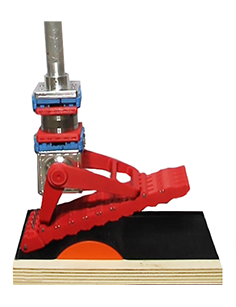}\\
61 mm
-22 \degree
\vspace{0.03 cm}}
\parbox{0.2\columnwidth}{
\centering
\includegraphics[align=c, width = 0.20\columnwidth]{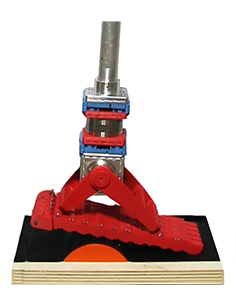}\\
112 mm
13 \degree
\vspace{0.03 cm}}
&
\parbox{0.2\columnwidth}{
\centering
\includegraphics[align=c, width = 0.20\columnwidth]{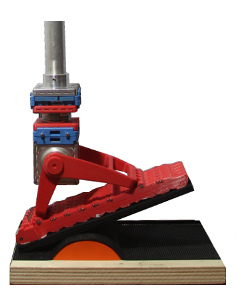}\\
53 mm
-16 \degree
\vspace{0.03 cm}}
\parbox{0.2\columnwidth}{
\centering
\includegraphics[align=c, width = 0.20\columnwidth]{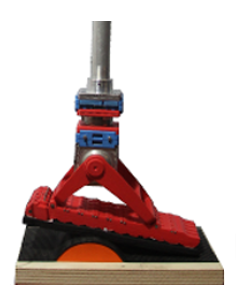}\\
52 mm
10 \degree
\vspace{0.03 cm}}&
\parbox{0.2\columnwidth}{
\centering
\includegraphics[align=c, width = 0.20\columnwidth]{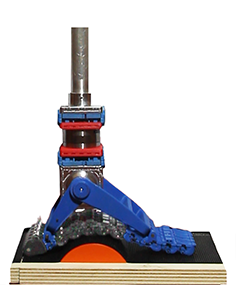}\\
103 mm
4 \degree
\vspace{0.03 cm}}  \\
\hline 6
%
\includegraphics[align=c, width = 0.20\columnwidth]{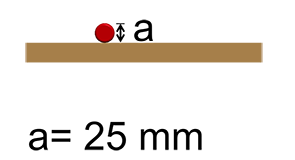}&
\parbox{0.2\columnwidth}{
\centering
\includegraphics[align=c, width = 0.20\columnwidth]{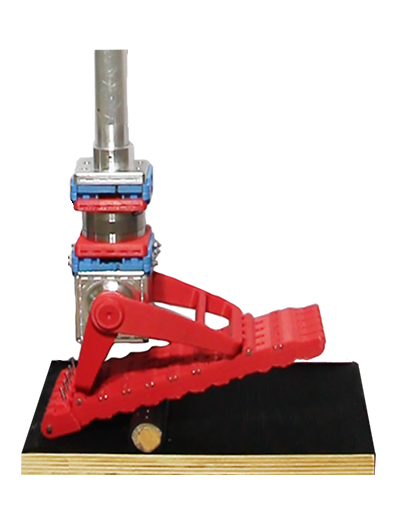}\\
46 mm
-11 \degree
\vspace{0.03 cm}}
\parbox{0.2\columnwidth}{
\centering
\includegraphics[align=c, width = 0.20\columnwidth]{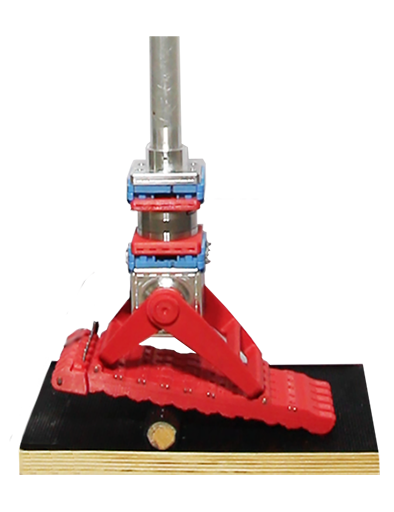}\\
78 mm
17 \degree
\vspace{0.03 cm}}
&
\parbox{0.2\columnwidth}{
\centering
\includegraphics[align=c, width = 0.20\columnwidth]{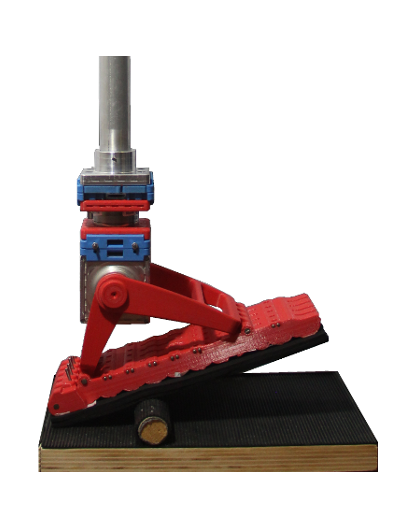}\\
47 mm
-16 \degree
\vspace{0.03 cm}}
\parbox{0.2\columnwidth}{
\centering
\includegraphics[align=c, width = 0.20\columnwidth]{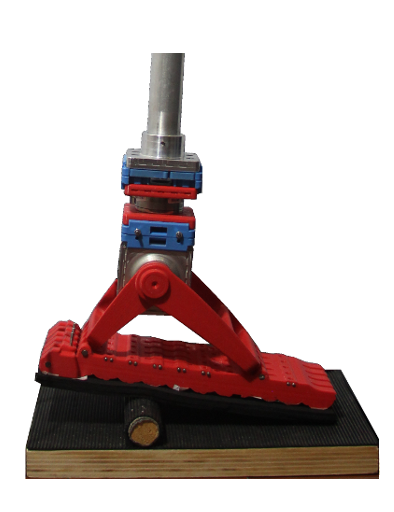}\\
92 mm
9 \degree
\vspace{0.03 cm}}
&
\parbox{0.2\columnwidth}{
\centering
\includegraphics[align=c, width = 0.20\columnwidth]{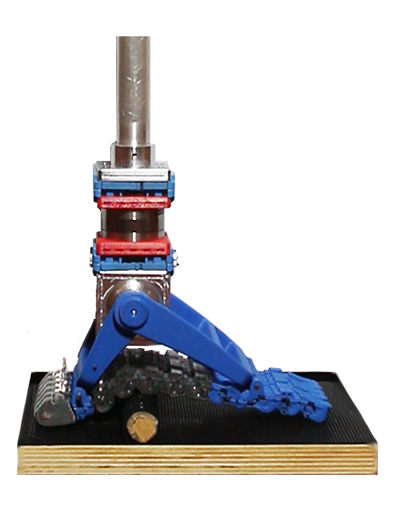}\\
107 mm
2 \degree
\vspace{0.03 cm}}  \\
\hline 7
%
\includegraphics[align=c, width = 0.20\columnwidth]{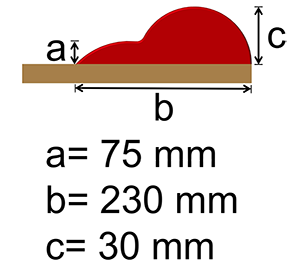}
&
\parbox{0.2\columnwidth}{
\centering
\includegraphics[align=c, width = 0.20\columnwidth]{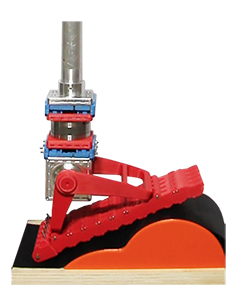}\\
103 mm
-19 \degree
\vspace{0.03 cm}}
&
\parbox{0.2\columnwidth}{
\centering
\includegraphics[align=c, width = 0.20\columnwidth]{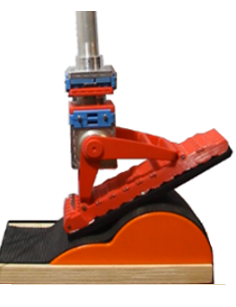}\\
45 mm
-28 \degree
\vspace{0.03 cm}}
&
\parbox{0.2\columnwidth}{
\centering
\includegraphics[align=c, width = 0.20\columnwidth]{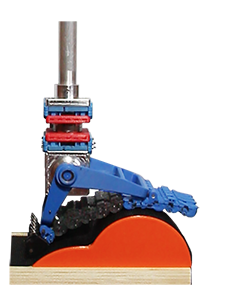}\\
77 mm
-17 \degree
\vspace{0.03 cm}}  \\
\hline 8
\includegraphics[align=c, width = 0.20\columnwidth]{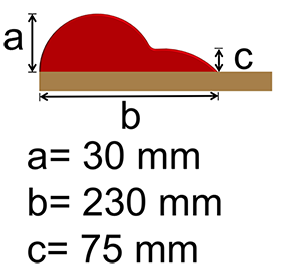}
&
\parbox{0.2\columnwidth}{
\centering
\includegraphics[align=c, width = 0.20\columnwidth]{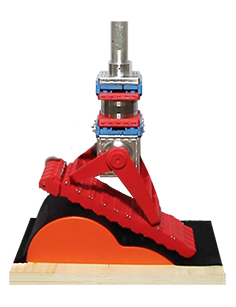}\\
41 mm
32 \degree
\vspace{0.03 cm}}
&
\parbox{0.2\columnwidth}{
\centering
\includegraphics[align=c, width = 0.20\columnwidth]{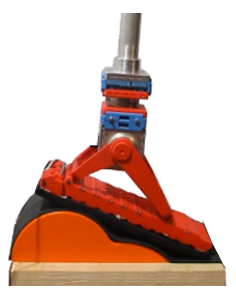}\\
98 mm
23 \degree
\vspace{0.03 cm}}
&
\parbox{0.2\columnwidth}{
\centering
\includegraphics[align=c, width = 0.20\columnwidth]{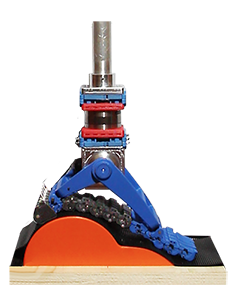}\\
102 mm
17 \degree
\vspace{0.03 cm}} \\
\hline
\end{tabular}
\caption{Results of Experiment 1. The SoftFoot design is compared to the rigid flat foot and to the compliant foot in terms of linear extension of the support polygon (Support length) and compensatory ankle pitch angle to keep the leg vertical (Ankle pitch).}
\label{tab:humble}
\end{table*}

\begin{figure*}[h!]
	\centering
	\subfigure[]{\includegraphics[width = 0.74\columnwidth]{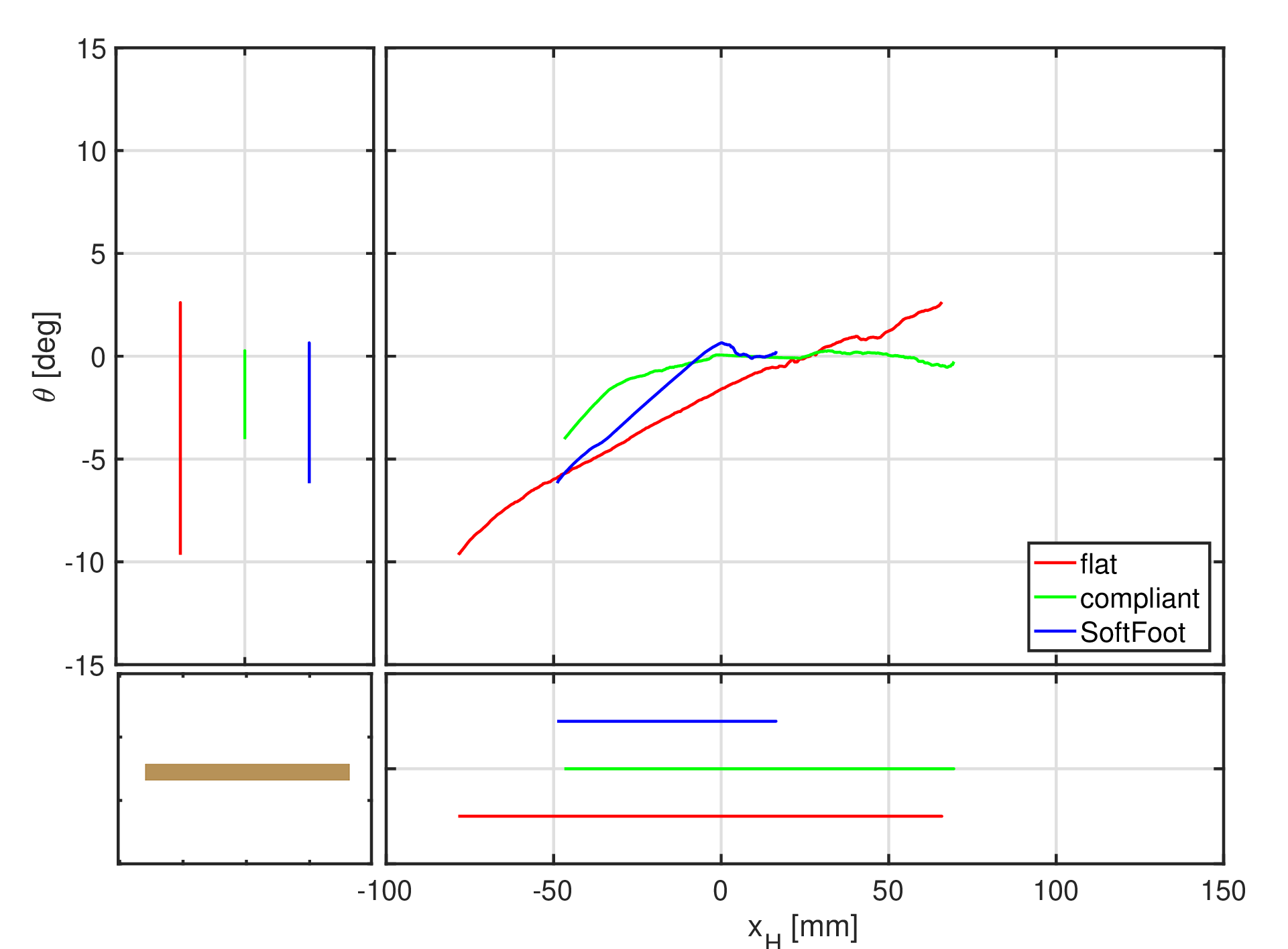}}
	\subfigure[]{\includegraphics[width = 0.74\columnwidth]{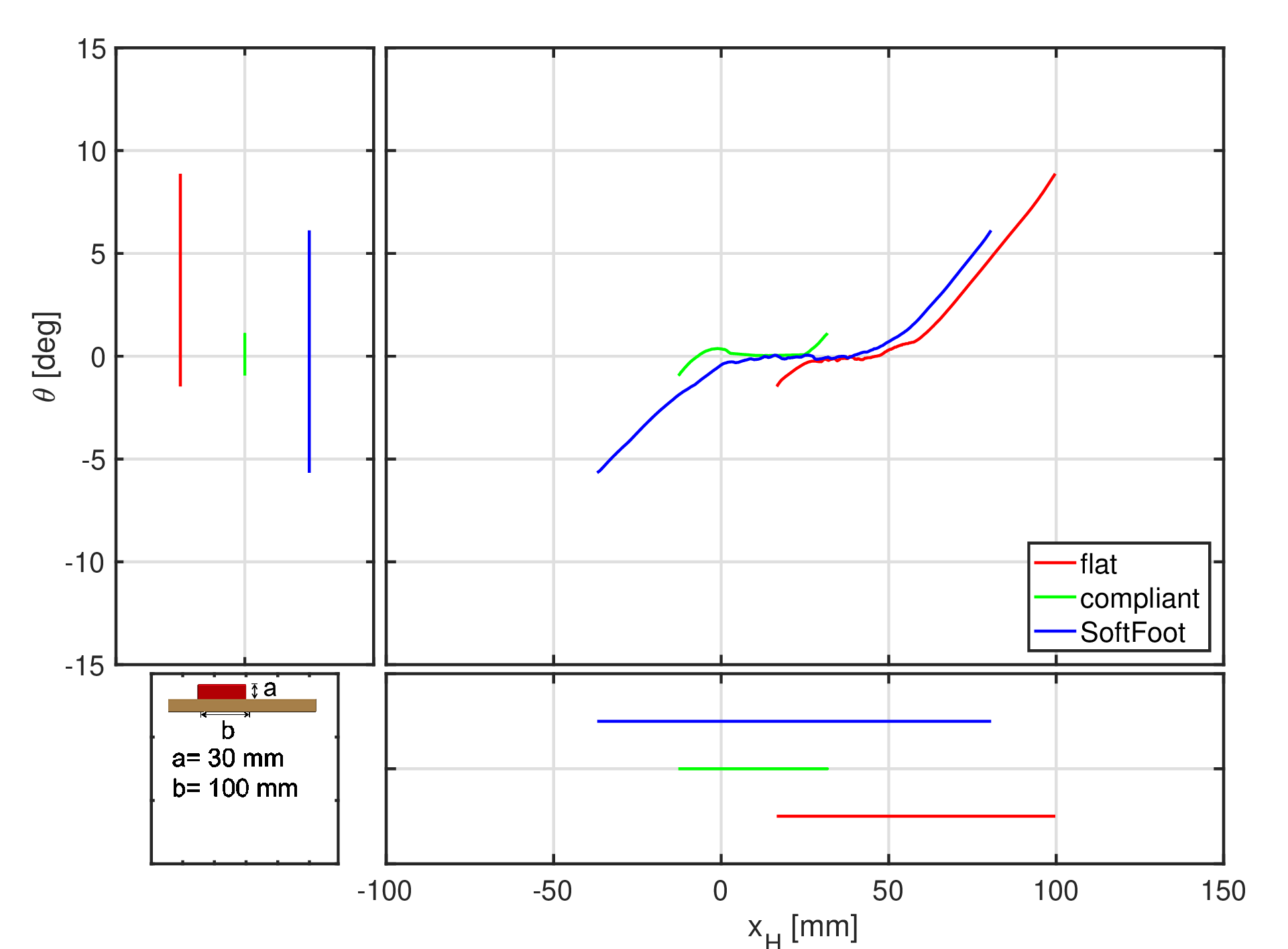}}
	\subfigure[]{\includegraphics[width = 0.74\columnwidth]{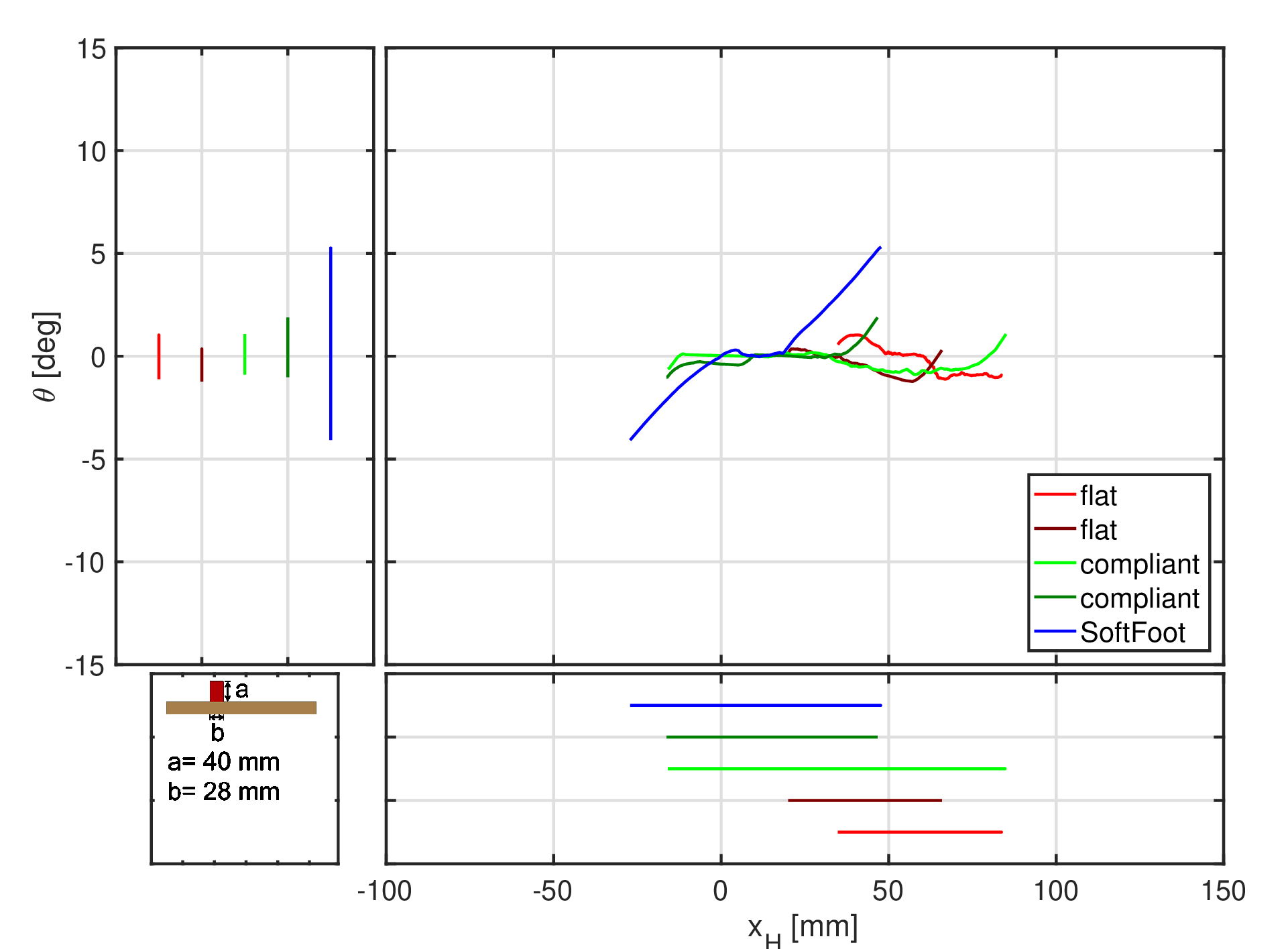}}
	\subfigure[]{\includegraphics[width = 0.74\columnwidth]{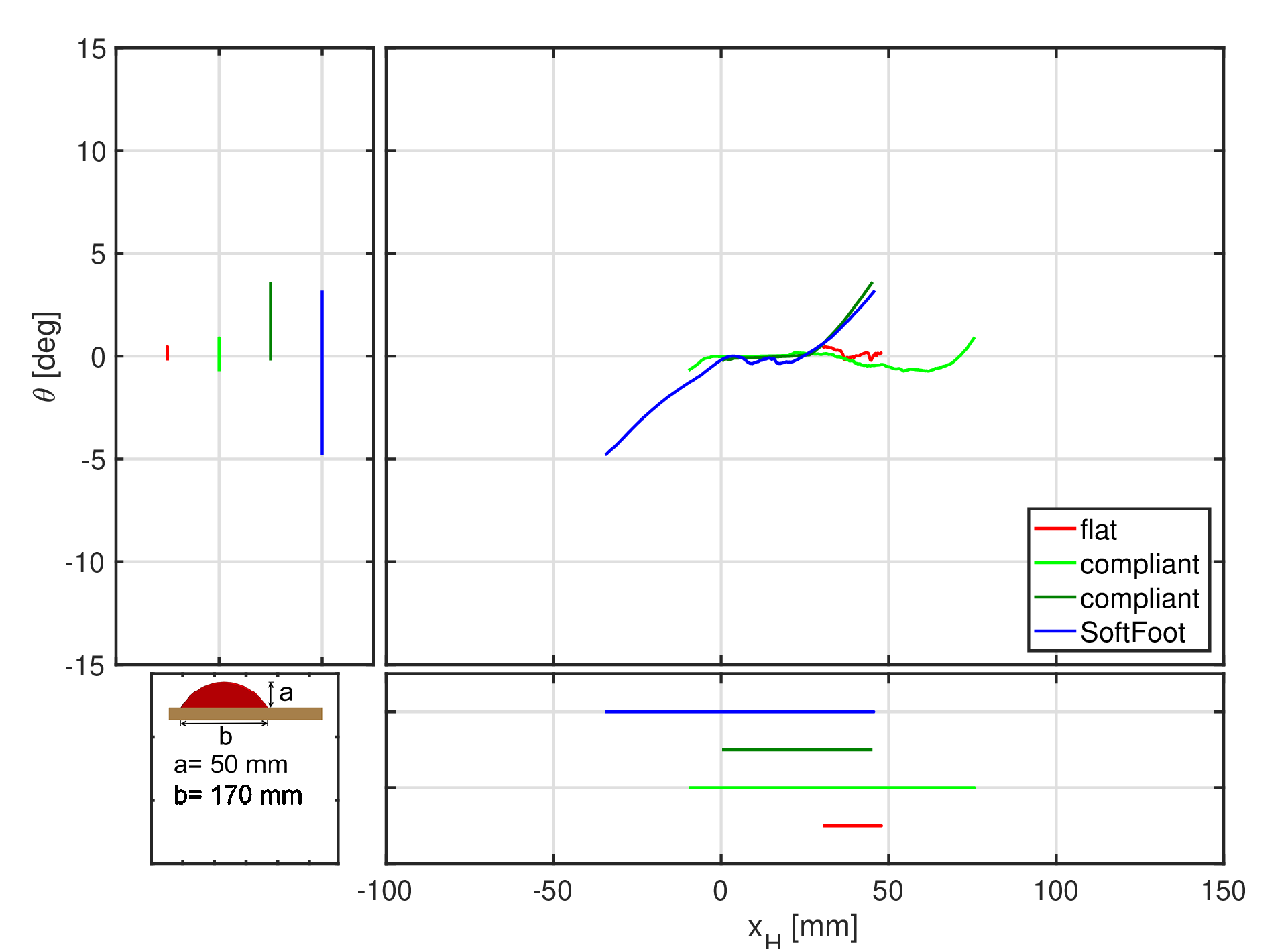}}
	\subfigure[]{\includegraphics[width = 0.74\columnwidth]{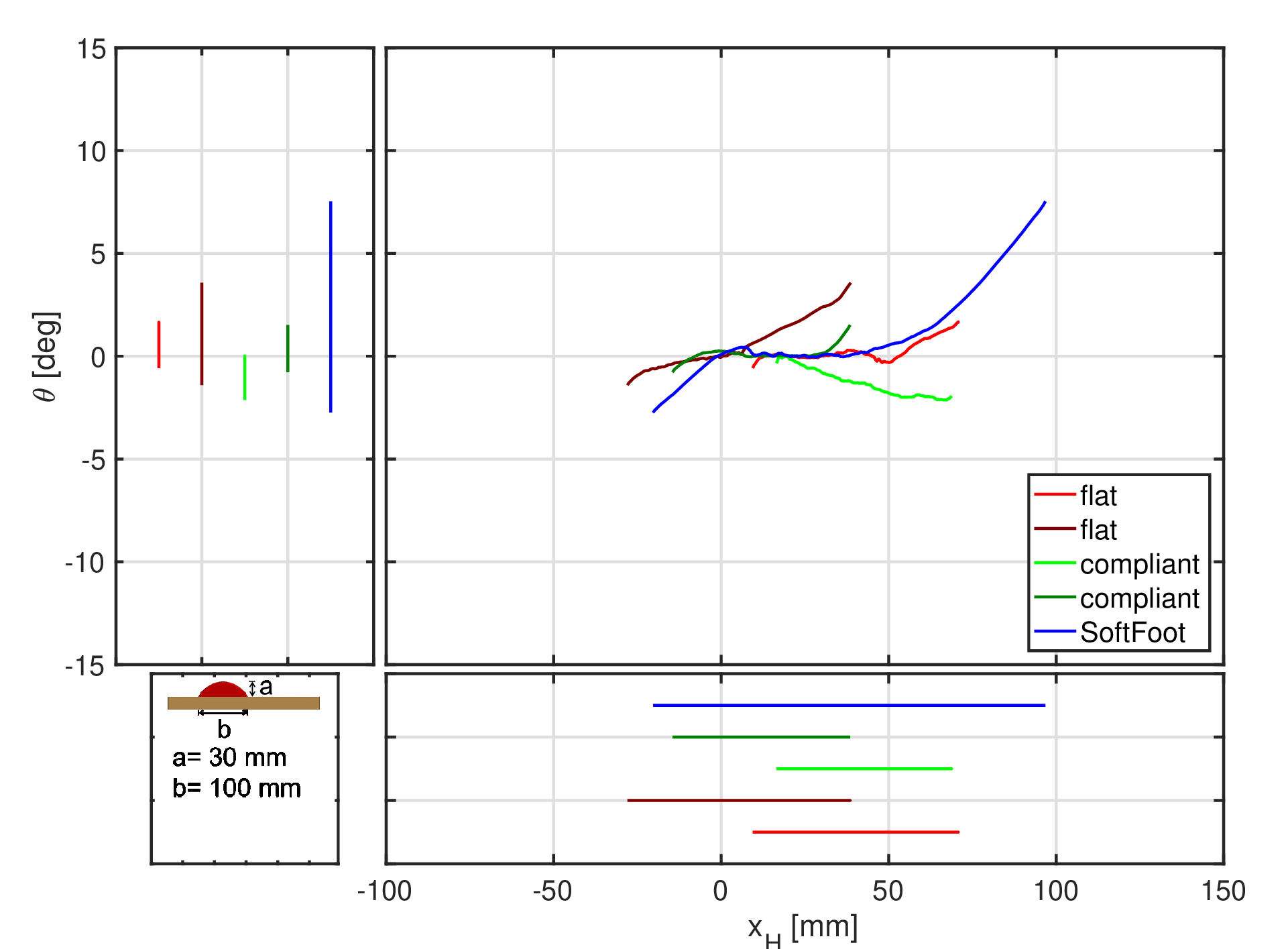}}
	\subfigure[]{\includegraphics[width = 0.74\columnwidth]{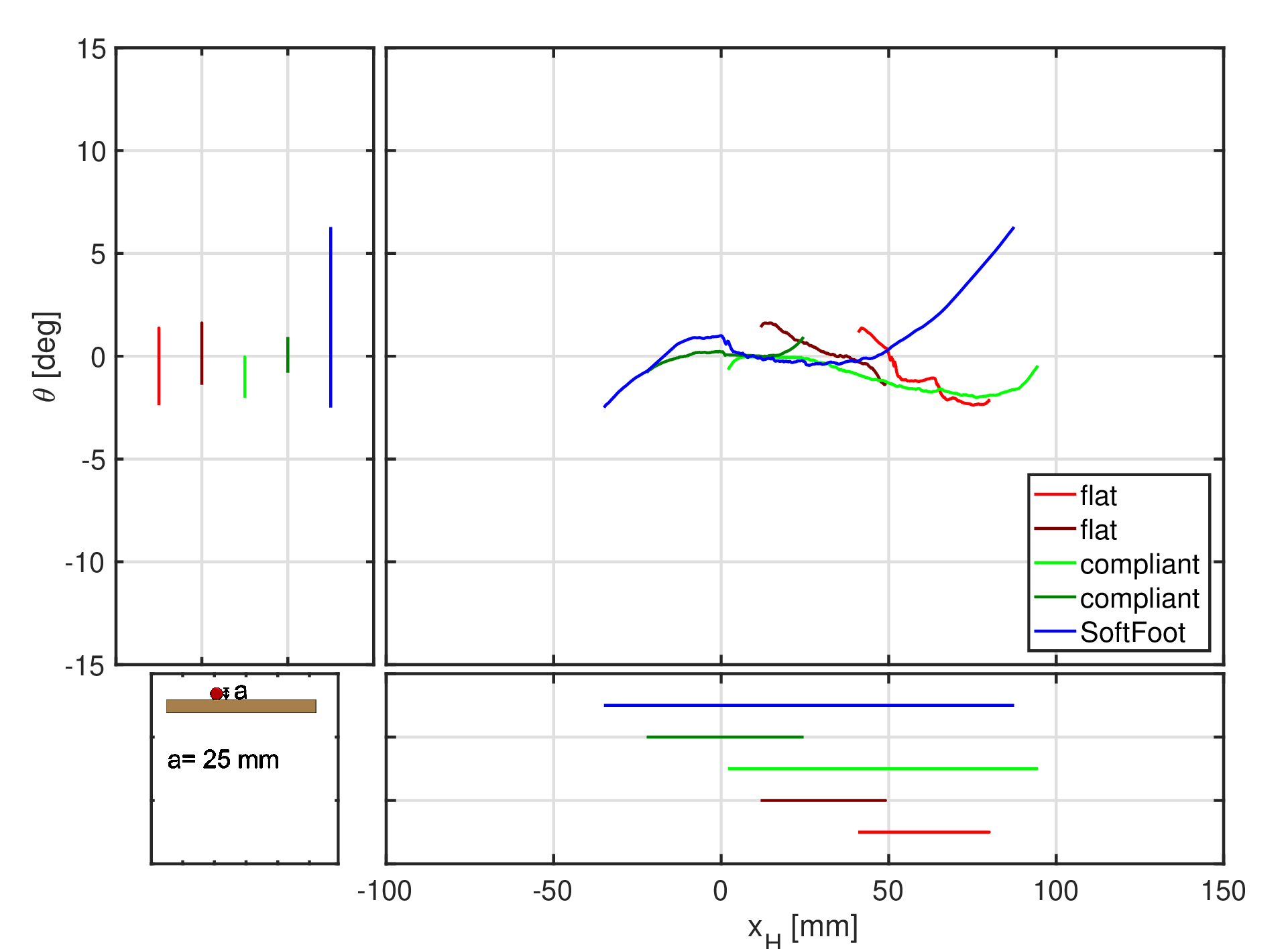}}
	\subfigure[]{\includegraphics[width = 0.74\columnwidth]{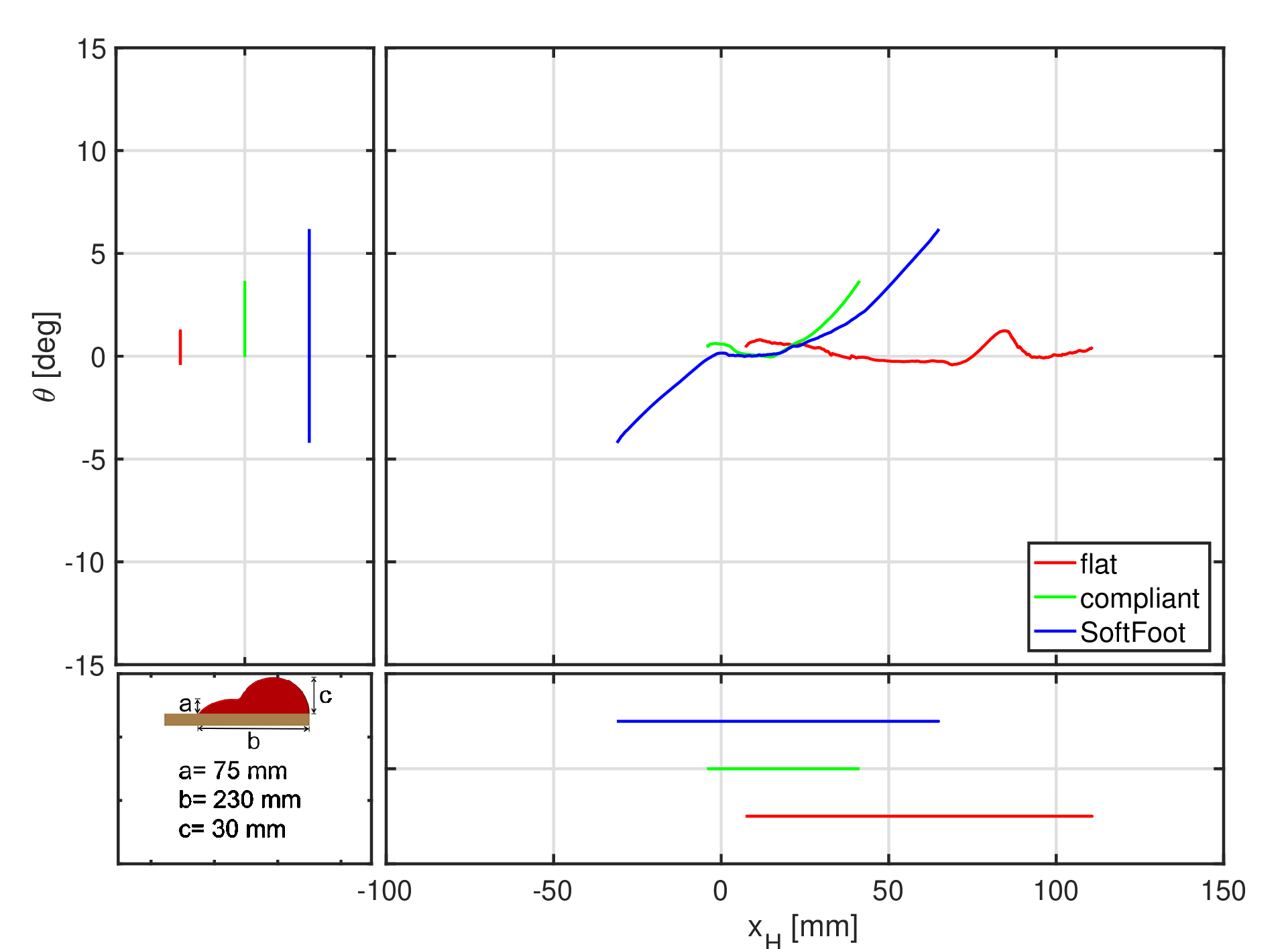}}
	\subfigure[]{\includegraphics[width = 0.74\columnwidth]{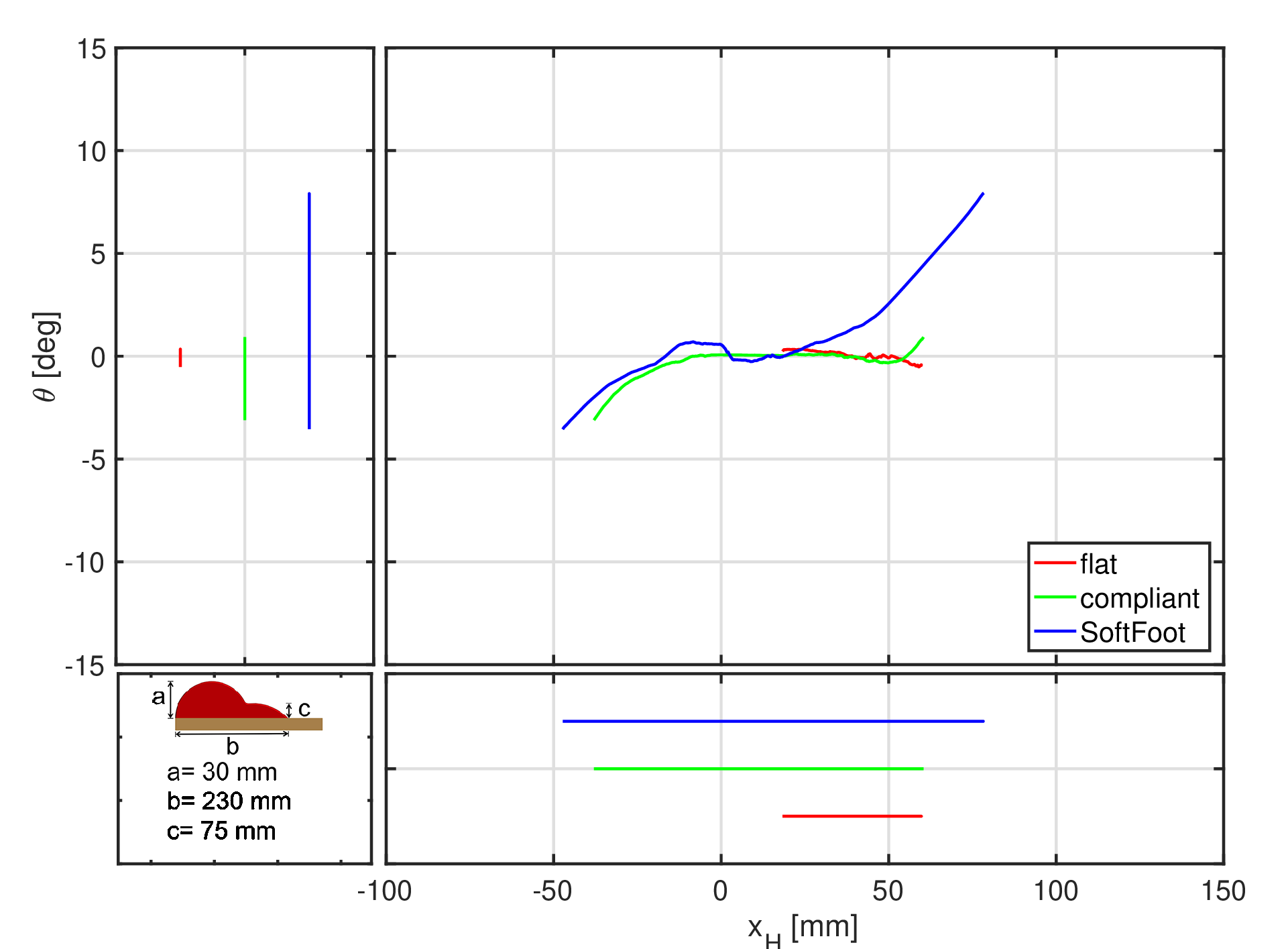}}
	\caption{Experimental Feet Stability Range Comparison: Compensatory Ankle Rotation as a function of Center of Pressure Position, for the three feet and the eight supports. For each loading condition, we report the experimental Compensatory Ankle Rotation as a function of Center of Pressure Position. The range of stable CoP and the range of necessary compensation angle are reported on the bottom and left side of each plot for clearer reading. Red color is for the rigid flat foot, Green for the compliant flat foot and Blue for the SoftFoot. When the two flat feet admit two different equilibria, both results are shown with a lighter and darker shades of the same color. For clearer comparison, compensatory angles are relative to the starting position of each experiment. For an Evaluation of the Static Compensatory angle see Table \ref{tab:humble}. Note that some compensatory angle can be appreciated also when the rigid flat foot is tested on the flat ground, as an effect of the ground compliance.}
	\label{fig:compensations}
\end{figure*}

\item \emph{Experiment 2:} The effectiveness of SoftFoot design was also evaluated testing the impact forces on different obstacles in a second experiment. This test is based on the experimental platform presented in Fig.~\ref{fig:iexp} (e). In this setup, the foot (3) can be moved along the vertical direction thanks to a rail (1). Different shapes and sizes of obstacles are exchanged in the lower part of the platform (4), in order to simulate and test different uneven terrain shapes. A wire (5) is locked on the lateral side of the carriage of the rail and course through a system with two pulleys. The vertical position of the foot is registered using a magnetic encoder (6) placed on the lower pulley and an electronic board (7). The force measures are revealed by an ATI Mini45 Force/Torque sensor (2) placed on top of the foot.
A mechanical stop was included in the vertical rail, to limit the maximum height of the foot and ensure the same starting position for each test. A human operator was moving the foot to the predefined initial position, as presented in the photo sequence of Fig.~\ref{fig:iexp} (e). After reaching the position, the foot was dropped against the obstacle. The experiment was repeated 3 times for each obstacle. 

\end{itemize}

Both tests were performed comparing the SoftFoot with a rigid and a compliant foot with the same weight and size, presented in the schematics of Fig.~\ref{fig:iexp} (a-c). The rigid foot was a 3-D printed fully rigid replica of the SoftFoot, while the compliant foot had a rigid structure but with an additional soft sole of 1 cm made of styrofoam. The thickness of the deformable sole was chosen to get close to the passive stability limit due to stiffness as defined in equation \ref{eq:n2}.
Note that while we study the foot behaviour on the sagittal plane, softer soles would have made the prototype leg unstable on the frontal plane. This would have rendered the experiment unfeasible without adding a mechanical constraint system that would have severely impacted results due to the unavoidable friction.
The three different feet designs were tested with eight artificial terrains to simulate different conditions like obstacles or roughness of the ground. The artificial terrains are made of plastic and fixed on a wooden support. The dimensions and shapes of each artificial terrain are presented in the first column of Table \ref{tab:humble}.
The results of both experimental evaluations conducted on the three feet design are presented and discussed in the following section.

\section{RESULTS AND DISCUSSION}\label{sec:experimental_results}
The outcomes of the initial experiment are presented in Table~\ref{tab:humble}. The first column shows the dimensions and shapes of the eight different artificial terrains adopted, while subsequent columns present the results for the three feet designs tested: a rigid flat foot, a compliant foot, and the SoftFoot. Each cell provides the linear extension of the projection of the support polygon (Support length) and the compensatory ankle pitch angle required to maintain the leg vertical (Ankle pitch). The images show the starting condition of the experiments and the level of adaptability of the foot to the corresponding obstacle.
The SoftFoot consistently demonstrated a smaller compensatory ankle pitch angle compared to the other two design solutions across all obstacles simulating uneven terrains (e.g., from obstacles 3 to 8). Similar results among the three feet are observed in more flat ground scenarios (e.g. obstacles 1 and 2). Additionally, it is worth noting that in cases of experiments with the SoftFoot, the support polygon was wider for six out of eight terrains compared to the rigid foot, and for seven out of eight terrains compared to the compliant one. The reduced support length in the flat terrain, in contrast to the rigid foot, can be attributed to the presence of the toe articulation in the SoftFoot.
\begin{figure}[t!]
\centering
	\includegraphics[width =\columnwidth]{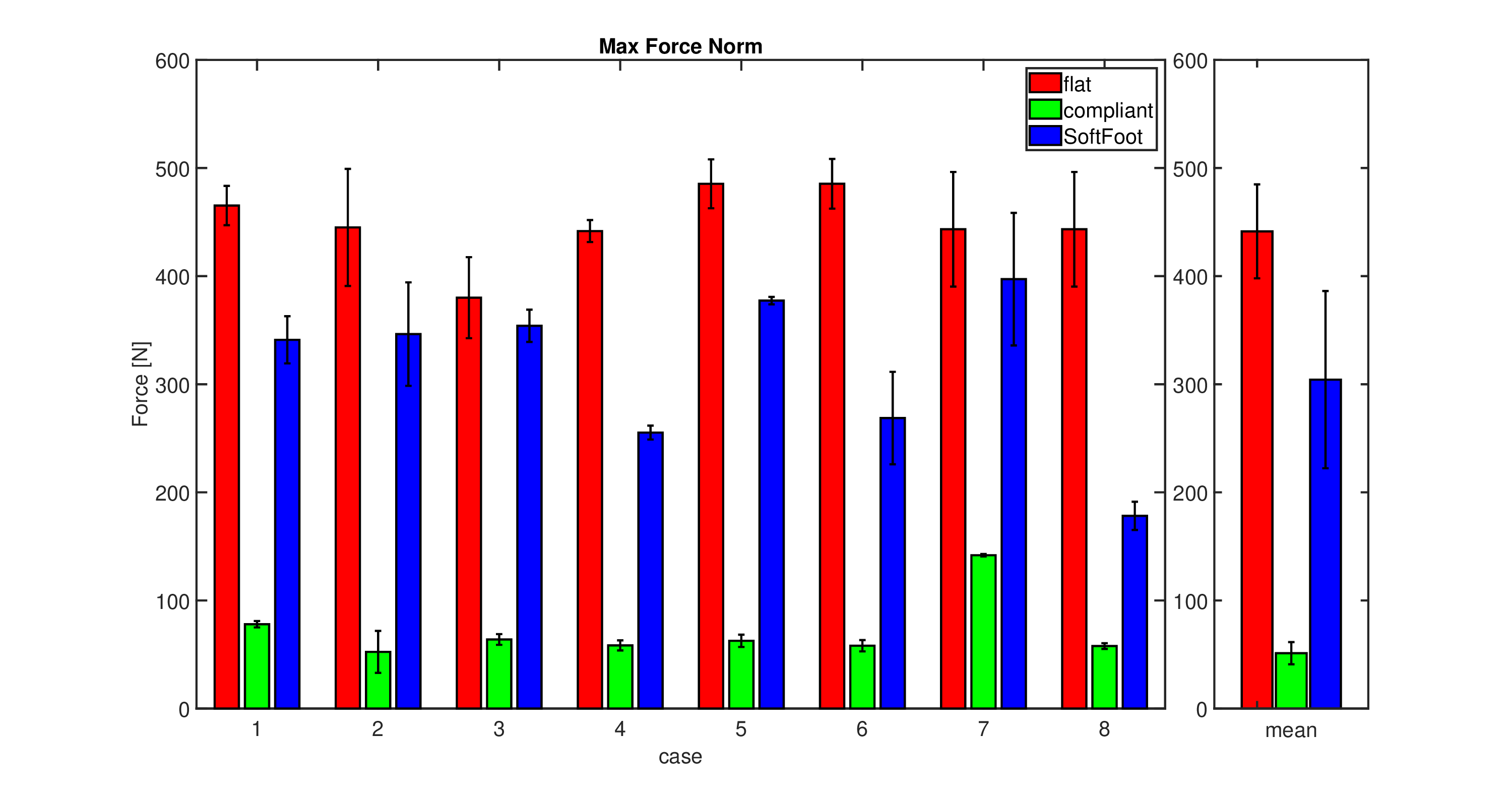}
	\caption{Drop experiment to evaluate maximum impact forces. For the three tested feet, the groups of bars on the left report the mean and standard deviation of the maximum impact forces for each obstacle. The rightmost group of bars reports global mean and standard deviation across all experimental conditions. Global distributions are significantly different using Kruskal-Wallis testing, with $p<0.01$.}
	\label{fig:imp_forces_max}
\end{figure}
\begin{figure}
\centering
	\includegraphics[width =\columnwidth]{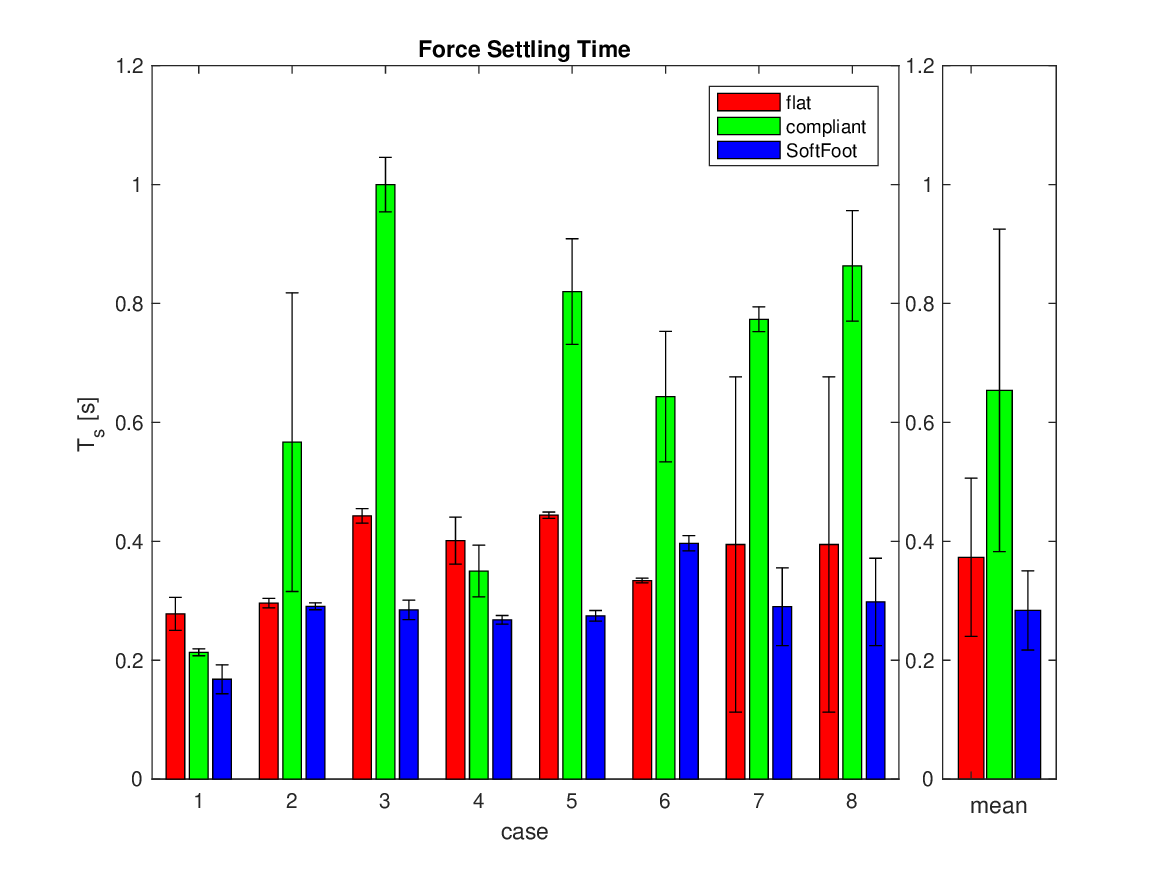}
	\caption{Drop experiment to evaluate impact forces oscillations settling time. For the three tested feet, the groups of bars on the left report the mean and standard deviation of the maximum settling time for each obstacle.}
	\label{fig:imp_forces_ts}
\end{figure}
Based on the outcome of this experimental evaluation, it is reasonable to think that using such a design in a legged robotic system could increase the region on which the robot could exert forces and decrease the tilting momenta, and thus reduce the control effort needed to balance the robot in case of locomotion on uneven terrain.
This is also visible in Fig.~\ref{fig:compensations}, which provides a comparison in terms of feet stability ranges. The results of compensatory ankle rotation are presented as a function of the Center of Pressure Position, for each foot and each artificial obstacle. This metric provides an estimation of the adjustments required to maintain the leg in a vertical position while navigating varied terrains. The SoftFoot outperformed other design solutions in most terrains, particularly those with larger uneven surfaces. The smaller compensatory ankle pitch angle of the SoftFoot translates into a more stable response of the robotic foot to environmental challenges and contributes to increased efficiency in maintaining equilibrium. This helps to minimize the control efforts needed to sustain stability during locomotion. 

Results from the second experiment are presented in Fig.~\ref{fig:imp_forces_max} and ~\ref{fig:imp_forces_ts}. The maximum impact forces for the three feet design and each of the eight artificial terrains are registered and are presented in Fig.~\ref{fig:imp_forces_max}. 
The results show that the SoftFoot consistently exhibited lower maximum impact forces when compared to the rigid foot. As expected, the compliant design shows still smaller impact forces among the three designs and for all eight obstacle cases. However, the analysis of the oscillation settling time for the three tested feet (see Fig.~\ref{fig:imp_forces_ts}) shows a more stable behaviour for the SoftFoot. On average, the SoftFoot outperformed the other two designs tested, suggesting an increased stability and capability to absorb impact for the proposed design.

The encouraging outcomes of this experimental validation highlight the potential of effectively using the SoftFoot in different fields of application, particularly in the realm of locomotion to improve the stability of robotic systems interacting and navigating through challenging environments. Its intrinsic capability to adapt to uneven terrain, while still being able to rigidly support the standing feet and to maintain a good extension of the contact surface, effectively extends the equivalent support polygon. A qualitative evaluation is presented in Fig.~\ref{fig:tattata}, where the SoftFoot is connected to an aluminum bar and guided by an operator through different walking phases when encountering an obstacle. It is possible to observe different sole configurations and the capability of the foot to adapt to several uneven configurations given by the geometry of the obstacle. 
\begin{figure*}[t]
	\centering
	\subfigure[]{\includegraphics[width = 0.45\columnwidth]{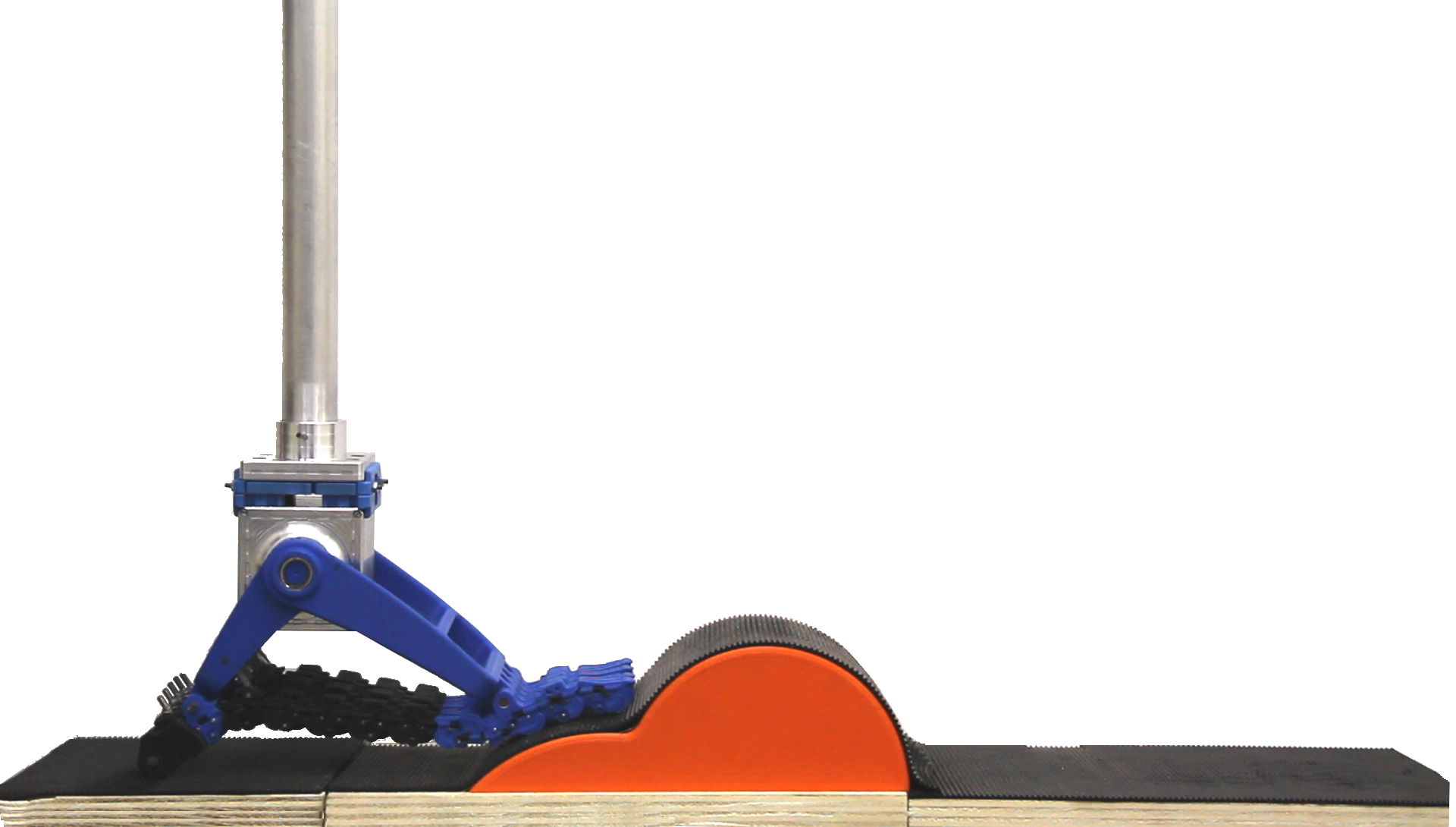}}
	\subfigure[]{\includegraphics[width = 0.45\columnwidth]{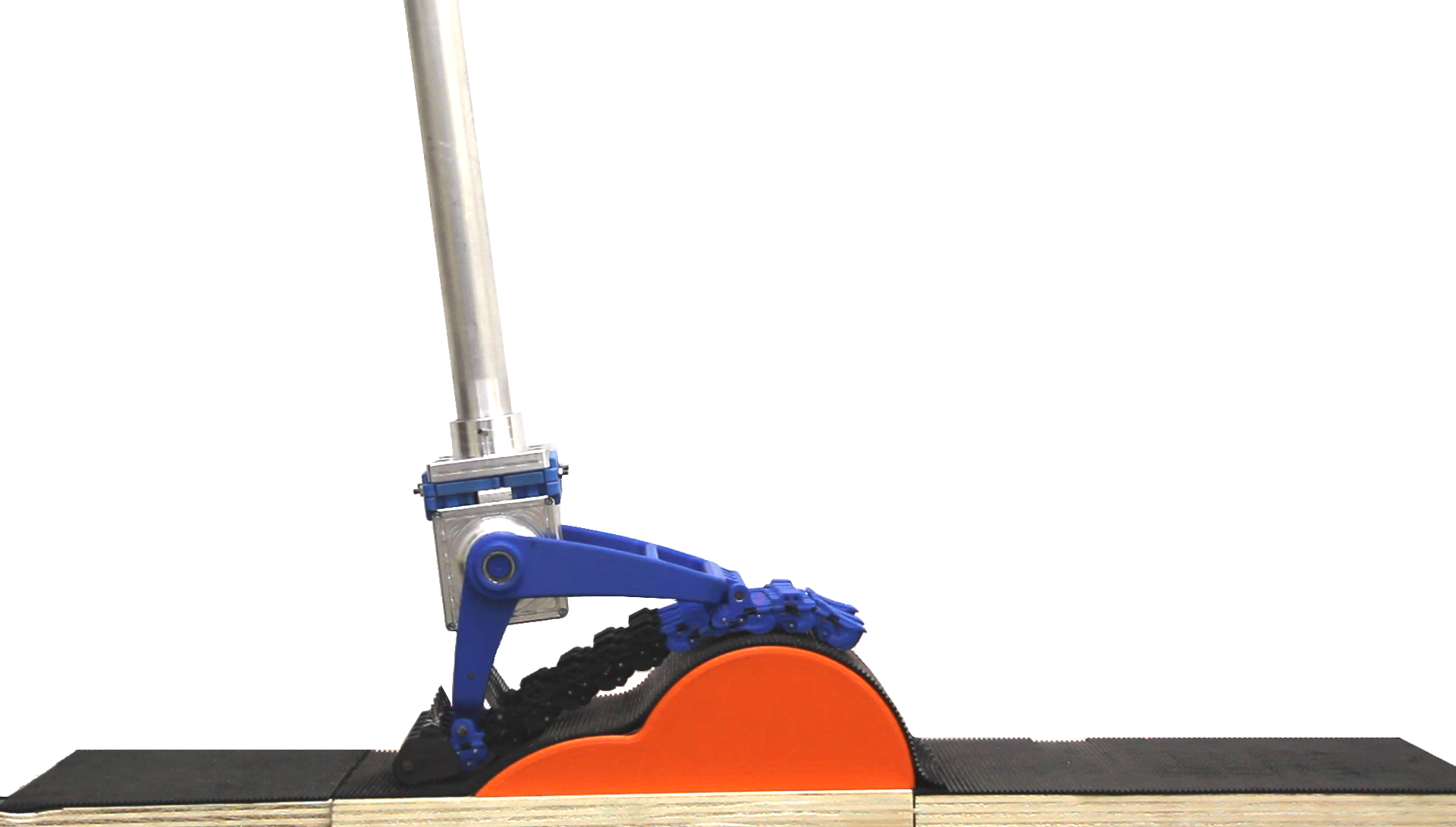}}
	\subfigure[]{\includegraphics[width = 0.45\columnwidth]{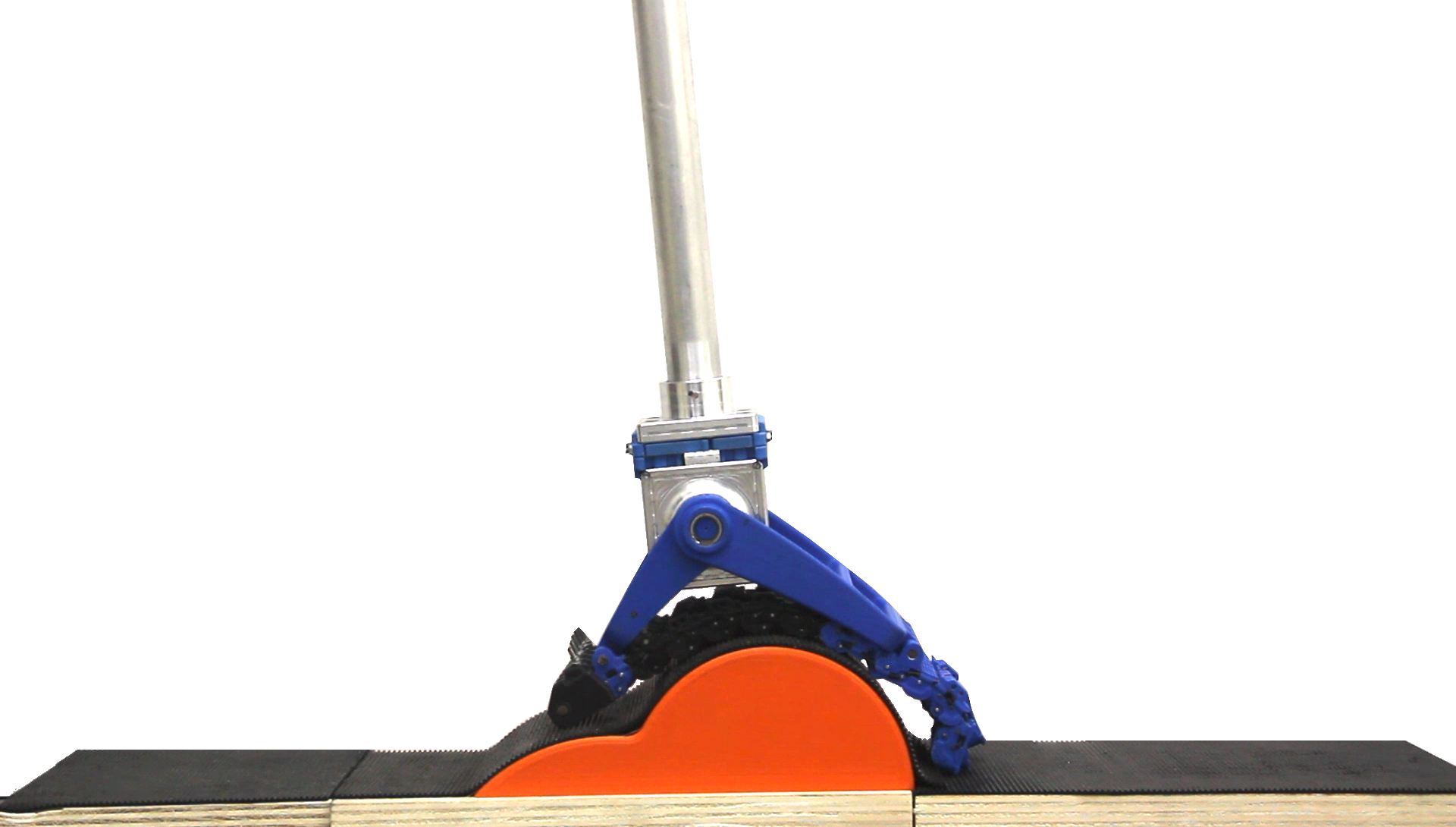}}
	\subfigure[]{\includegraphics[width = 0.45\columnwidth]{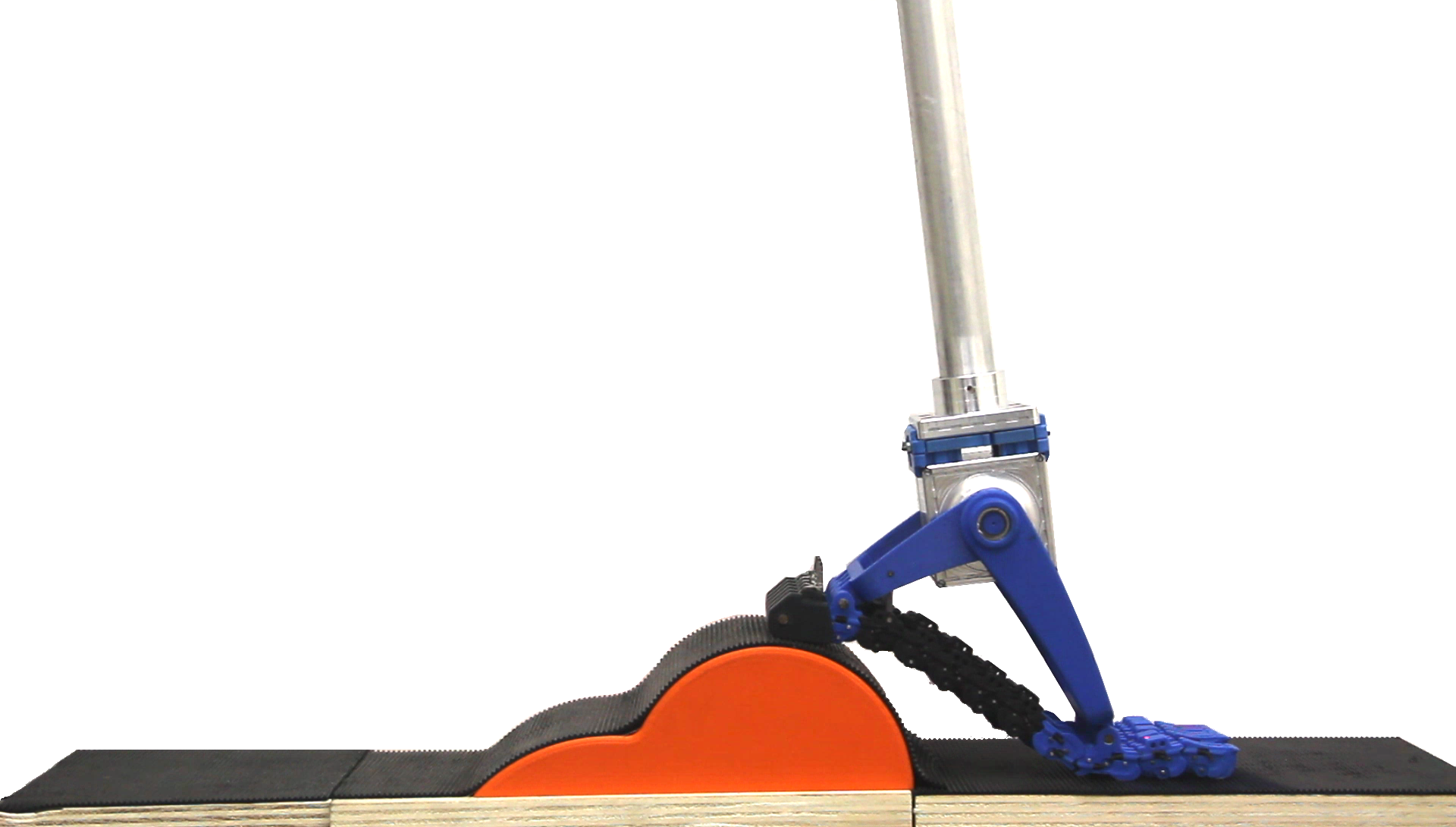}}
	\caption{Qualitative demonstration of different SoftFoot behaviours when walking on an artificial obstacle, simulating an uneven terrain. The SoftFoot is connected to an aluminium bar and guided by an operator through the different stages of the walking.}
	\label{fig:tattata}
\end{figure*}
Despite this paper presents the simulation of walking behavior over rigid obstacles, the compliant nature of the SoftFoot, coupled with its adaptability, could provide safe and robust interactions across various types of terrains, e.g. soft or granular terrains. These testing scenarios will be included in future works. The SoftFoot capability to conform to irregularities in terrain enhances its performance, making it particularly suitable for environments with uncertain or dynamic surfaces. In wet environments, where traditional robotic systems might face operational challenges, the passive design of the SoftFoot enables reliable functionality. 

The ability to maintain stability while walking on obstacles and effectively absorb impact, as evidenced by the experimental results, proves the potential of the SoftFoot for applications that conventionally demand advanced control algorithms, even in unpredictable or uneven scenarios. 

Moreover, the SoftFoot compliant behavior and capability to absorb impact open up possibilities for novel uses, including activities that involve dynamic movements, such as jumping. The reduced impact forces and improved stability, as indicated by the oscillation settling time, suggest that the SoftFoot can be employed in scenarios requiring controlled take-offs and landings. 

\section{REAL-WORLD APPLICATIONS}\label{sec:applications}
Explorative testing of the SoftFoot concept has demonstrated promising results across multiple real-world scenarios, as presented in Fig.~\ref{fig:application}. As initial validation, the SoftFoot was integrated with the humanoid robot HRP-4 \cite{catalano2020hrp}, evaluating its capability of balancing, stepping, and walking on flat ground and different obstacles. Results indicated a substantial improvement in waling performance when using the SoftFoot compared to the robot's original flat feet. This is also visible even without optimising the controller for the SoftFoot. 
Similarly, the SoftFoot concept was adapted for and tested on the quadrupedal robot ANYmal \cite{catalano2021adaptive, bednarek2020cnn}. However, a simplified version of the SoftFoot with reduced compliance in the joint design we adopted for this scenario. Both extensive field  and indoor tests demonstrated significant performance enhancements, particularly in reducing robot slippage, compared to traditional flat or ball feet designs.
Finally, we are investigating the potential of SoftFoot as lower limb prosthesis. As a first step towards this application, the SoftFoot is inserted into a pair of shoes to better replicate the human experience. A preliminary qualitative evaluation have demonstrated that its bio-inspired design effectively mimics the adaptive nature of human feet, even when covered by shoes, providing a compliant and stable base that can adjust to various types of obstacles. Fig.~\ref{fig:application}(c) shows a comparison between a rigid foot (left) with a similar footprint and sole. Future studies will explore this application more extensively, including experiments with subjects with lower limb loss. These promising results highlights the potential to expand the application scope of the SoftFoot into various fields, ranging from robotics to medical technologies.

\begin{figure*}[t]
	\centering
	\subfigure[]{\includegraphics[height = 0.65\columnwidth]{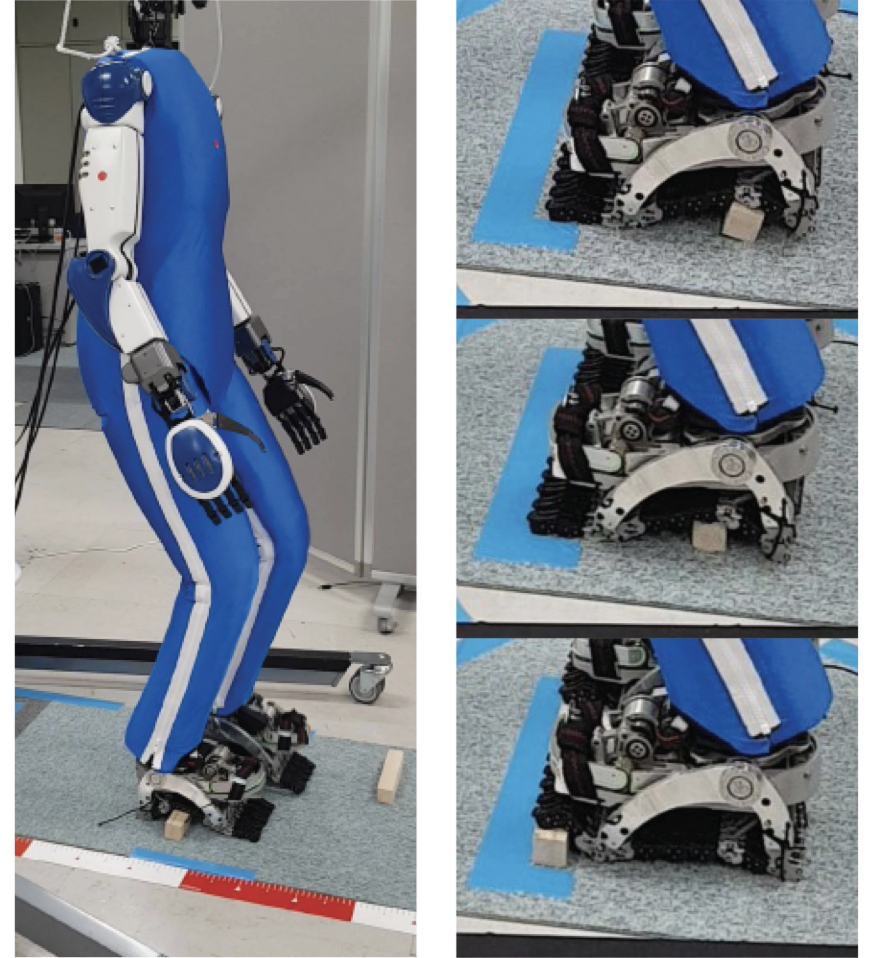}}
	\subfigure[]{\includegraphics[height = 0.65\columnwidth]{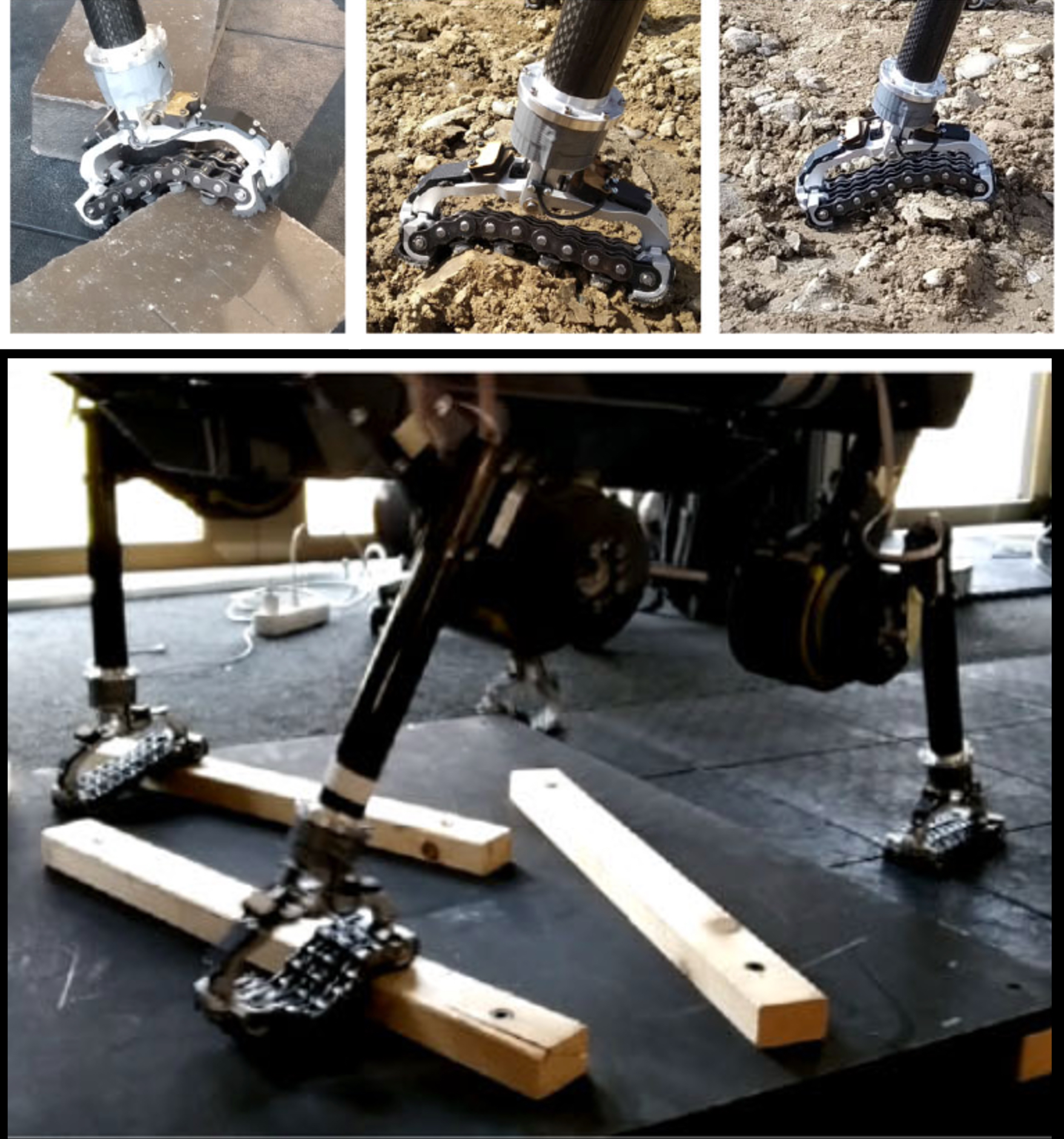}}
	\subfigure[]{\includegraphics[height = 0.65\columnwidth]{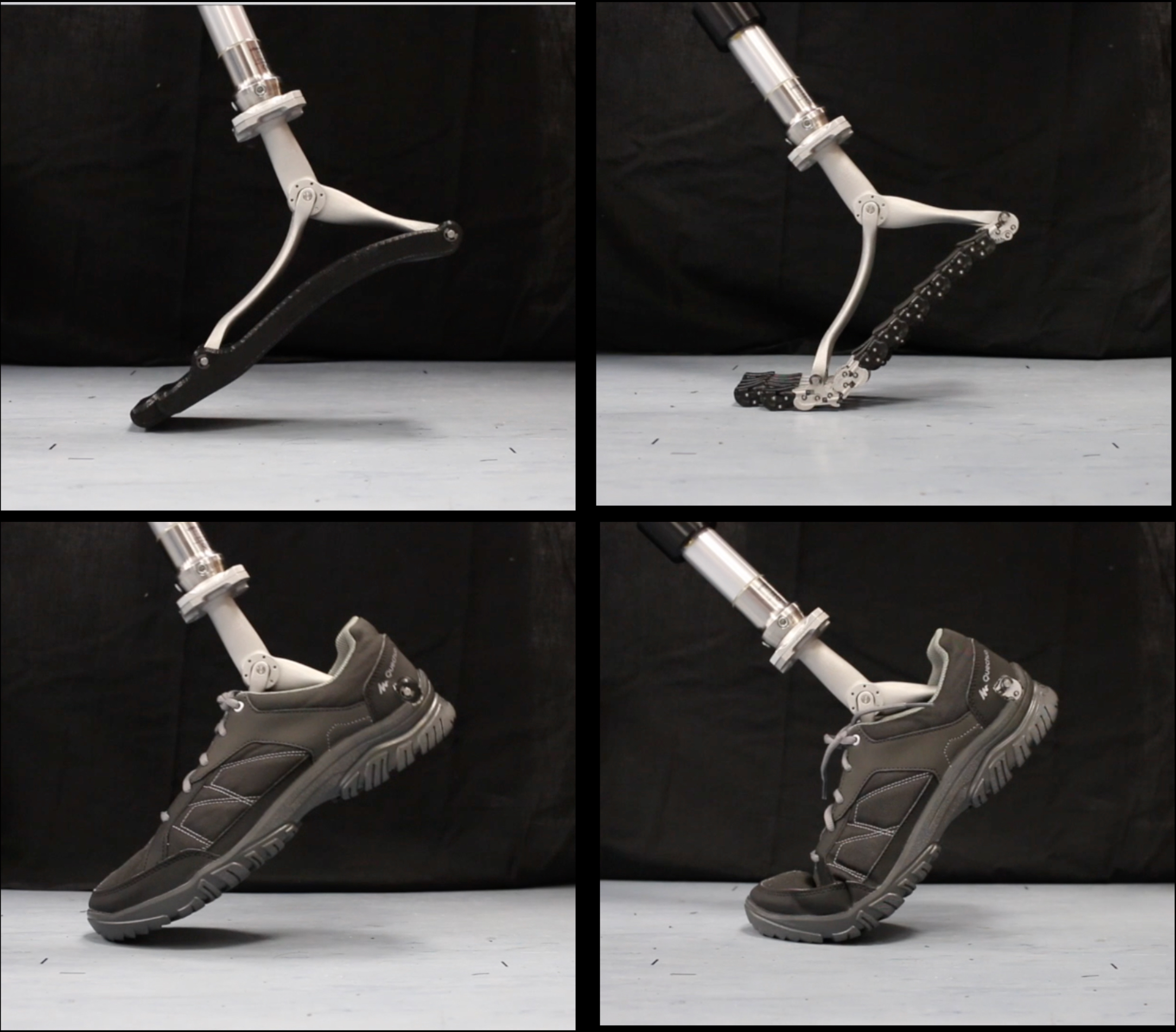}}
	\caption{Examples of possible applications of the SoftFoot: (a) humanoid robotics, where the SoftFoot was integrated with the humanoid robot HRP-4 \cite{catalano2020hrp}, (b) quadrupedal robotics, where a simplified version of the SoftFoot with reduced compliance was adopted \cite{catalano2021adaptive} and (c) potential use as lower limb prosthesis, where the SoftFoot (right) is inserted into a pair of shoes and qualitatively compared with a rigid foot (left) with a similar footprint.}
	\label{fig:application}
\end{figure*}


\section{CONCLUSIONS}\label{sec:concl}
This paper presents a framework for the design of SoftFoot, a passive foot capable to adapt its shape to uneven terrains and different roughness of the ground. The proposed design consists of a combination of flexible and rigid elements, which make the artificial foot robust but at the same time compliant and adaptive.
Its bio-inspired architecture combined with the use of soft robotics technologies allows to obtain good performance in terms of safe interaction and resilience. 
The effectiveness of the SoftFoot was experimentally validated in terms of robot stability, adaptation capabilities, and resistance to impacts. Its performance is compared with a rigid and a compliant foot with the same size and weight. 
The results of this study offer valuable insights for potential design improvements, such as increasing compliance to achieve impact forces similar to those of a compliant foot or expanding the adaptive behavior to other directions besides the sagittal plane. 
Future work will focus on application to dynamic locomotion, including extensive experimental analyses of the effect of SoftFoot on the stability of the swing phase. Moreover, the proposed design will be translated to specific fields of application, to evaluate its performance and potential in real-world environments. This will include its evaluation with humanoid or quadrupedal robots, as well as in novel application fields, such as lower limb prostheses. Further investigations will include improvements on the device implementation, according to the requirements of the selected field of application. Finally, future works will explore alternative design features such as an active version of the system, where it is possible to modulate the adaptive sole configuration through an active controller. 

\section*{ACKNOWLEDGMENT}
The authors would like to thank Andrea Di Basco and Gian Maria Gasparri for their valuable help in the realization of the prototypes and support in the experimental testing.  
\bibliographystyle{ieeetr}
\bibliography{biblio_}

\begin{thebibliography}{10}

\bibitem{putz2006sobotta}
R.~Putz and R.~Pabst, {\em Sobotta-Atlas of Human Anatomy: Head, Neck, Upper
  Limb, Thorax, Abdomen, Pelvis, Lower Limb; Two-volume set.}
\newblock 2006.

\bibitem{venkadesan2020stiffness}
M.~Venkadesan, A.~Yawar, C.~M. Eng, M.~A. Dias, D.~K. Singh, S.~M. Tommasini,
  A.~H. Haims, M.~M. Bandi, and S.~Mandre, ``Stiffness of the human foot and
  evolution of the transverse arch,'' {\em Nature}, vol.~579, no.~7797,
  pp.~97--100, 2020.

\bibitem{torricelli2016human}
D.~Torricelli, J.~Gonzalez, M.~Weckx, R.~Jim{\'e}nez-Fabi{\'a}n,
  B.~Vanderborght, M.~Sartori, S.~Dosen, D.~Farina, D.~Lefeber, and J.~L. Pons,
  ``Human-like compliant locomotion: state of the art of robotic
  implementations,'' {\em Bioinspiration \& biomimetics}, vol.~11, no.~5,
  p.~051002, 2016.

\bibitem{frizza2022humanoids}
I.~Frizza, K.~Ayusawa, A.~Cherubini, H.~Kaminaga, P.~Fraisse, and G.~Venture,
  ``Humanoids’ feet: State-of-the-art \& future directions,'' {\em
  International Journal of Humanoid Robotics}, vol.~19, no.~01, p.~2250001,
  2022.

\bibitem{jaeger2023cybathlon}
L.~Jaeger, R.~d.~S. Baptista, C.~Basla, P.~Capsi-Morales, Y.~K. Kim,
  S.~Nakajima, C.~Piazza, M.~Sommerhalder, L.~Tonin, G.~Valle, {\em et~al.},
  ``How the cybathlon competition has advanced assistive technologies,'' {\em
  Annual Review of Control, Robotics, and Autonomous Systems}, vol.~6,
  pp.~447--476, 2023.

\bibitem{hutter2017anymal}
M.~Hutter, C.~Gehring, A.~Lauber, F.~Gunther, C.~D. Bellicoso, V.~Tsounis,
  P.~Fankhauser, R.~Diethelm, S.~Bachmann, M.~Bl{\"o}sch, {\em et~al.},
  ``Anymal-toward legged robots for harsh environments,'' {\em Advanced
  Robotics}, vol.~31, no.~17, pp.~918--931, 2017.

\bibitem{sprowitz2013towards}
A.~Spr{\"o}witz, A.~Tuleu, M.~Vespignani, M.~Ajallooeian, E.~Badri, and A.~J.
  Ijspeert, ``Towards dynamic trot gait locomotion: Design, control, and
  experiments with cheetah-cub, a compliant quadruped robot,'' {\em The
  International Journal of Robotics Research}, vol.~32, no.~8, pp.~932--950,
  2013.

\bibitem{BostonDyn}
``Boston dynamics website.'' http://www.bostondynamics.com.
\newblock Accessed: 2016-03-01.

\bibitem{park2005mechanical}
I.-W. Park, J.-Y. Kim, J.~Lee, and J.-H. Oh, ``Mechanical design of humanoid
  robot platform khr-3 (kaist humanoid robot 3: Hubo),'' in {\em Humanoid
  Robots, 2005 5th IEEE-RAS International Conference on}, pp.~321--326, IEEE,
  2005.

\bibitem{kaneko2008humanoid}
K.~Kaneko, K.~Harada, F.~Kanehiro, G.~Miyamori, and K.~Akachi, ``Humanoid robot
  hrp-3,'' in {\em Intelligent Robots and Systems, 2008. IROS 2008. IEEE/RSJ
  International Conference on}, pp.~2471--2478, IEEE, 2008.

\bibitem{gouaillier2009mechatronic}
D.~Gouaillier, V.~Hugel, P.~Blazevic, C.~Kilner, J.~O. Monceaux, P.~Lafourcade,
  B.~Marnier, J.~Serre, and B.~Maisonnier, ``Mechatronic design of nao
  humanoid,'' in {\em Robotics and Automation, 2009. ICRA'09. IEEE
  International Conference on}, pp.~769--774, IEEE, 2009.

\bibitem{negrello2016walkman}
F.~Negrello, M.~Garabini, M.~G. Catalano, P.~Kryczka, W.~Choi, D.~Caldwell,
  A.~Bicchi, and N.~Tsagarakis, ``Walk-man humanoid lower body design
  optimization for enhanced physical performance,'' in {\em Robotics and
  Automation, 2016. ICRA'16. IEEE International Conference on}, 2016.

\bibitem{bretl2008testing}
T.~Bretl and S.~Lall, ``Testing static equilibrium for legged robots,'' {\em
  IEEE Transactions on Robotics}, vol.~24, no.~4, pp.~794--807, 2008.

\bibitem{li2008flexible}
J.~Li, Q.~Huang, W.~Zhang, Z.~Yu, and K.~Li, ``Flexible foot design for a
  humanoid robot,'' in {\em Automation and Logistics, 2008. ICAL 2008. IEEE
  International Conference on}, pp.~1414--1419, IEEE, 2008.

\bibitem{najmuddin2012experimental}
A.~Najmuddin, Y.~Fukuoka, and S.~Ochiai, ``Experimental development of
  stiffness adjustable foot sole for use by bipedal robots walking on uneven
  terrain,'' in {\em System Integration (SII), 2012 IEEE/SICE International
  Symposium on}, pp.~248--253, IEEE, 2012.

\bibitem{tsagarakis2011design}
N.~G. Tsagarakis, Z.~Li, J.~Saglia, and D.~G. Caldwell, ``The design of the
  lower body of the compliant humanoid robot ``ccub'','' in {\em Robotics and
  Automation (ICRA), 2011 IEEE International Conference on}, pp.~2035--2040,
  IEEE, 2011.

\bibitem{davis2010design}
S.~Davis and D.~G. Caldwell, ``The design of an anthropomorphic dexterous
  humanoid foot,'' in {\em 2010 IEEE/RSJ International Conference on
  Intelligent Robots and Systems}, pp.~2200--2205, IEEE, 2010.

\bibitem{kang2010realization}
H.-j. Kang, K.~Hashimoto, H.~Kondo, K.~Hattori, K.~Nishikawa, Y.~Hama, H.-o.
  Lim, A.~Takanishi, K.~Suga, and K.~Kato, ``Realization of biped walking on
  uneven terrain by new foot mechanism capable of detecting ground surface,''
  in {\em Robotics and Automation (ICRA), 2010 IEEE International Conference
  on}, pp.~5167--5172, IEEE, 2010.

\bibitem{kuehn2012active}
D.~Kuehn, F.~Beinersdorf, F.~Bernhard, K.~Fondahl, M.~Schilling, M.~Simnofske,
  T.~Stark, and F.~Kirchner, ``Active spine and feet with increased sensing
  capabilities for walking robots,'' in {\em International Symposium on
  Artificial Intelligence, Robotics and Automation in Space (iSAIRAS-12)},
  pp.~4--6, 2012.

\bibitem{narioka2012humanlike}
K.~Narioka, T.~Homma, and K.~Hosoda, ``Humanlike ankle-foot complex for a biped
  robot,'' in {\em 2012 12th IEEE-RAS International Conference on Humanoid
  Robots (Humanoids 2012)}, pp.~15--20, IEEE, 2012.

\bibitem{seo2009modeling}
J.-T. Seo and B.-J. Yi, ``Modeling and analysis of a biomimetic foot
  mechanism,'' in {\em 2009 IEEE/RSJ International Conference on Intelligent
  Robots and Systems}, pp.~1472--1477, IEEE, 2009.

\bibitem{yoon2006novel}
J.~Yoon, H.~Nandha, D.~Lee, and G.-s. Kim, ``A novel 4-dof robotic foot
  mechanism with multi-platforms for humanoid robot (sice-iccas 2006),'' in
  {\em 2006 SICE-ICASE International Joint Conference}, pp.~3500--3504, IEEE,
  2006.

\bibitem{piazza2019century}
C.~Piazza, G.~Grioli, M.~Catalano, and A.~Bicchi, ``A century of robotic
  hands,'' {\em Annual Review of Control, Robotics, and Autonomous Systems},
  vol.~2, pp.~1--32, 2019.

\bibitem{paez2019adaptive}
L.~Paez, K.~Melo, R.~Thandiackal, and A.~J. Ijspeert, ``Adaptive compliant foot
  design for salamander robots,'' in {\em 2019 2nd IEEE International
  Conference on Soft Robotics (RoboSoft)}, pp.~178--185, IEEE, 2019.

\bibitem{asano2016human}
Y.~Asano, S.~Nakashima, T.~Kozuki, S.~Ookubo, I.~Yanokura, Y.~Kakiuchi,
  K.~Okada, and M.~Inaba, ``Human mimetic foot structure with multi-dofs and
  multi-sensors for musculoskeletal humanoid kengoro,'' in {\em 2016 IEEE/RSJ
  International Conference on Intelligent Robots and Systems (IROS)},
  pp.~2419--2424, IEEE, 2016.

\bibitem{kaslin2018towards}
R.~K{\"a}slin, H.~Kolvenbach, L.~Paez, K.~Lika, and M.~Hutter, ``Towards a
  passive adaptive planar foot with ground orientation and contact force
  sensing for legged robots,'' in {\em 2018 IEEE/RSJ International Conference
  on Intelligent Robots and Systems (IROS)}, pp.~2707--2714, IEEE, 2018.

\bibitem{hauser2018compliant}
S.~Hauser, M.~Mutlu, P.~Banzet, and A.~J. Ijspeert, ``Compliant universal
  grippers as adaptive feet in legged robots,'' {\em Advanced Robotics},
  vol.~32, no.~15, pp.~825--836, 2018.

\bibitem{piazza2016toward}
C.~Piazza, C.~Della~Santina, G.~M. Gasparri, M.~G. Catalano, G.~Grioli,
  M.~Garabini, and A.~Bicchi, ``Toward an adaptive foot for natural walking,''
  in {\em 2016 IEEE-RAS 16th International Conference on Humanoid Robots
  (Humanoids)}, pp.~1204--1210, IEEE, 2016.

\bibitem{mura2019exploiting}
D.~Mura, C.~Della~Santina, C.~Piazza, I.~Frizza, C.~Morandi, M.~Garabini,
  G.~Grioli, and M.~G. Catalano, ``Exploiting adaptability in soft feet for
  sensing contact forces,'' {\em IEEE Robotics and Automation Letters}, 2019.

\bibitem{catalano2020hrp}
M.~G. Catalano, I.~Frizza, C.~Morandi, G.~Grioli, K.~Ayusawa, T.~Ito, and
  G.~Venture, ``Hrp-4 walks on soft feet,'' {\em IEEE Robotics and Automation
  Letters}, vol.~6, no.~2, pp.~470--477, 2020.

\bibitem{catalano2021adaptive}
M.~G. Catalano, M.~J. Pollayil, G.~Grioli, G.~Valsecchi, H.~Kolvenbach,
  M.~Hutter, A.~Bicchi, and M.~Garabini, ``Adaptive feet for quadrupedal
  walkers,'' {\em IEEE Transactions on Robotics}, vol.~38, no.~1, pp.~302--316,
  2021.

\bibitem{bednarek2020cnn}
J.~Bednarek, N.~Maalouf, M.~J. Pollayil, M.~Garabini, M.~G. Catalano,
  G.~Grioli, and D.~Belter, ``Cnn-based foothold selection for mechanically
  adaptive soft foot,'' in {\em 2020 IEEE/RSJ International Conference on
  Intelligent Robots and Systems (IROS)}, pp.~10225--10232, IEEE, 2020.

\bibitem{popovic2005ground}
M.~B. Popovic, A.~Goswami, and H.~Herr, ``Ground reference points in legged
  locomotion: Definitions, biological trajectories and control implications,''
  {\em The International Journal of Robotics Research}, vol.~24, no.~12,
  pp.~1013--1032, 2005.

\bibitem{sardain2004forces}
P.~Sardain and G.~Bessonnet, ``Forces acting on a biped robot. center of
  pressure-zero moment point,'' {\em IEEE Transactions on Systems, Man, and
  Cybernetics-Part A: Systems and Humans}, vol.~34, no.~5, pp.~630--637, 2004.

\bibitem{sato2009stability}
T.~Sato, S.~Sakaino, and K.~Ohnishi, ``Stability index for biped robot moving
  on rough terrain,'' {\em IEEJ Transactions on Industry Applications},
  vol.~129, no.~6, 2009.

\bibitem{caron2015zmp}
S.~Caron, Q.-C. Pham, and Y.~Nakamura, ``Zmp support areas for multi-contact
  mobility under frictional constraints,'' {\em arXiv preprint
  arXiv:1510.03232}, 2015.

\bibitem{bicchi1993contact}
A.~Bicchi, J.~K. Salisbury, and D.~L. Brock, ``Contact sensing from force
  measurements,'' {\em The International Journal of Robotics Research},
  vol.~12, no.~3, pp.~249--262, 1993.

\bibitem{kajita2003biped}
S.~Kajita, F.~Kanehiro, K.~Kaneko, K.~Fujiwara, K.~Harada, K.~Yokoi, and
  H.~Hirukawa, ``Biped walking pattern generation by using preview control of
  zero-moment point,'' in {\em Robotics and Automation, 2003. Proceedings.
  ICRA'03. IEEE International Conference on}, vol.~2, pp.~1620--1626, IEEE,
  2003.

\bibitem{hicks1954mechanics}
J.~Hicks, ``The mechanics of the foot: Ii. the plantar aponeurosis and the
  arch,'' {\em Journal of anatomy}, vol.~88, no.~Pt 1, p.~25, 1954.

\bibitem{della2018toward}
C.~Della~Santina, C.~Piazza, G.~Grioli, M.~G. Catalano, and A.~Bicchi, ``Toward
  dexterous manipulation with augmented adaptive synergies: The pisa/iit
  softhand 2,'' {\em IEEE Transactions on Robotics}, vol.~34, no.~5,
  pp.~1141--1156, 2018.

\bibitem{kelley2018numerical}
C.~Kelley, ``Numerical methods for nonlinear equations,'' {\em Acta Numerica},
  vol.~27, pp.~207--287, 2018.

\bibitem{catalano2014adaptive}
M.~G. Catalano, G.~Grioli, E.~Farnioli, A.~Serio, C.~Piazza, and A.~Bicchi,
  ``Adaptive synergies for the design and control of the pisa/iit softhand,''
  {\em The International Journal of Robotics Research}, vol.~33, no.~5,
  pp.~768--782, 2014.

\bibitem{hillberry1976rolling}
B.~Hillberry and A.~Hall~Jr, ``Rolling contact joint,'' Jan.~13 1976.
\newblock US Patent 3,932,045.

\end{thebibliography}

\end{document}